\documentclass[lettersize,journal]{IEEEtran}
\usepackage{amsmath,amsfonts}
\usepackage{algorithmic}
\usepackage{algorithm}
\usepackage{array}
\usepackage[caption=false,font=normalsize,labelfont=sf,textfont=sf]{subfig}
\usepackage{textcomp}
\usepackage{stfloats}
\usepackage{url}
\usepackage{verbatim}
\usepackage{graphicx}
\usepackage{cite}
\DeclareMathOperator*{\argmin}{argmin}
\usepackage{amsthm}
\usepackage{graphicx}
\usepackage{amsthm}
\usepackage{siunitx}
\usepackage{float}
\usepackage{multirow}

\newtheorem{example}{Example}

\DeclareMathOperator*{\argmax}{arg\,max}

\begin{document}

\title{A Scalable Training Strategy for Blind Multi-Distribution Noise Removal}


\markboth{Journal of \LaTeX\ Class Files,~Vol.~14, No.~8, August~2021}%
{Shell \MakeLowercase{\textit{et al.}}: A Sample Article Using IEEEtran.cls for IEEE Journals}
\author{Kevin Zhang, Sakshum Kulshrestha, Christopher Metzler
\thanks{This work was supported in part by AFOSR Young Investigator Program Award no. FA9550-22-1-0208, ONR award no. N000142312752, NSF CAREER Award no.~2339616, and a seed grant from SAAB, Inc.}}

\maketitle

\begin{abstract}
Despite recent advances, developing general-purpose universal denoising and artifact-removal networks remains largely an open problem: Given fixed network weights, one inherently trades-off specialization at one task (e.g.,~removing Poisson noise) for performance at another (e.g.,~removing speckle noise).
In addition, training such a network is challenging due to the curse of dimensionality: As one increases the dimensions of the specification-space (i.e.,~the number of parameters needed to describe the noise distribution) the number of unique specifications one needs to train for grows exponentially. 
Uniformly sampling this space will result in a network that does well at very challenging problem specifications but poorly at easy problem specifications, where even large errors will have a small effect on the overall mean squared error.

In this work we propose training denoising networks using an adaptive-sampling/active-learning strategy. Our work improves upon a recently proposed universal denoiser training strategy by extending these results to higher dimensions and by incorporating a polynomial approximation of the true specification-loss landscape. This approximation allows us to reduce training times by almost two orders of magnitude. We test our method on simulated joint Poisson-Gaussian-Speckle noise and demonstrate that with our proposed training strategy, a single blind, generalist denoiser network can achieve peak signal-to-noise ratios within a uniform bound of specialized denoiser networks across a large range of operating conditions. 
We also capture a small dataset of images with varying amounts of joint Poisson-Gaussian-Speckle noise and demonstrate that a universal denoiser trained using our adaptive-sampling strategy outperforms uniformly trained baselines.
\end{abstract}


\section{Introduction}
Neural networks have become the gold standard for solving a host of imaging inverse problems~\cite{ongie2020deep}. From denoising and deblurring to compressive sensing and phase retrieval, modern deep neural networks significantly outperform classical techniques like BM3D~\cite{dabov2007image} and KSVD~\cite{aharon2006k}. 

The most straightforward and common approach to apply deep learning to inverse problems is to train a neural network to learn a mapping from the space of corrupted images/measurements to the space of clean images. In this framework, one first captures or creates a training set consisting of clean images $x_1,x_2,\dots$ and corrupted images $y_1,y_2,\dots$ according to some known forward model $p(y_i|x_i,\theta)$, where $\theta\in\Theta$ denotes the latent variable(s) specifying the forward model. For example, when training a network to remove additive white Gaussian noise
\begin{align}
    p(y_i|x_i,\theta) = \frac{1}{\sigma\sqrt{2\pi}}\exp{-\frac{\|y_i-x_i\|^2}{2\sigma^2}},
\end{align}
and the latent variable $\theta$ is the standard deviation $\sigma$. With a training set of $L$ pairs $\{x_i,y_i\}_{i=1}^L$ in hand, one can then train a network to learn a mapping from $y$ to $x$.

Typically, we are not interested in recovering signals from a single corruption distribution (e.g., a single fixed noise standard deviation $\sigma$) but rather a range of distributions. For example, we might want to remove additive white Gaussian noise with standard deviations anywhere in the range $[0,50]$ ($\Theta=\{\sigma|\sigma\in [0,50]\}$). The size of this range determines how much the network needs to generalize and there is inherently a trade-off between specialization and generalization. By and large, a network trained to reconstruct images over a large range of corruptions (a larger set $\Theta$) will under-perform a network trained and specialized over a narrow range~\cite{zhang2017beyond}. 

This problem becomes significantly more challenging when dealing with mixed, multi-distribution noise. As one increases the number of parameters (e.g.,~Gaussian standard deviation, Poisson rate, number of speckle realizations, ...) the space of corrupted signals one needs to reconstruct grows exponentially: The specification space becomes the Cartesian product (e.g.,~$\Theta=\Theta_{Gaussian}\times\Theta_{Poisson}\times\Theta_{speckle}$) of the spaces of each of the individual noise distributions.

This expansion does not directly prevent someone (with enough compute resources) from training a ``universal'' denoising algorithm. One can sample from $\Theta$, generate a training batch, optimize the network to minimize some reconstruction loss, and repeat. However, this process depends heavily on the policy/probability-density-function $\pi$ used to sample from $\Theta$. As noted in~\cite{gnanasambandam2020one} and corroborated in Section~\ref{sec:Results}, uniformly sampling from $\Theta$ will produce networks that do well on hard examples but poorly (relative to how well a specialized network performs) on easy examples.

\subsection{Our Contribution}
In this work, we develop an adaptive-sampling/active-learning strategy that allows us to train a single ``universal'' network to remove mixed Poisson-Gaussian-Speckle noise such that the network consistently performs within a uniform bound of specialized bias-free DnCNN baselines~\cite{zhang2017beyond,mohan2019robust}. Our key contribution is a novel, polynomial approximation of the specification-loss landscape. This approximation allows us to tractably apply (using over $50\times$ fewer training examples than it would otherwise require) the adaptive-sampling strategy developed in~\cite{gnanasambandam2020one}, wherein training a denoiser is framed as a constrained optimization problem. 
We validate our technique with both simulated and experimentally captured data.

\section{Related Work}
Overcoming the specialization-generalization trade-off has been the focus of intense research efforts over the last 5 years.

\subsection{Adaptive Denoising}
One approach to improve generalization is to provide the network information about the current problem specifications $\theta$ at test time. 
For example,~\cite{gharbiDenoising2016}~demonstrated one could provide a constant standard-deviation map as an extra channel to a denoising network so that it could adapt to i.i.d.~Gaussian noise.~\cite{zhang2018ffdnet} extends this idea by adding a general standard-deviation map as an extra channel, to deal with spatially-varying Gaussian noise. This idea was recently extended to deal with correlated Gaussian noise~\cite{metzler2021d}. The same framework can be extended to more complex tasks like compressive sensing, deblurring, and descattering as well~\cite{melba:2022:017:wang,Tahir2022}.
These techniques are all non-blind and require an accurate estimate of the specification parameters $\theta$ to be effective.



\subsection{Universal Denoising}
Somewhat surprisingly, the aforementioned machinery may be unnecessary if the goal is to simply remove additive white Gaussian noise over a range of different standard deviations:~\cite{mohan2019robust}~recently demonstrated one can achieve significant invariance to the noise level by simply removing biases from the network architecture.~\cite{Wang2014} also achieves similar invariance to noise level by scaling the input images to the denoiser to match the distribution it was trained on. Alternatively, at a potentially large computational cost, one can apply iterative ``plug and play'' or diffusion models that allow one to denoise a signal contaminated with noise with parameters $\theta'$ using a denoiser/diffusion model trained for minimum mean squared error  additive white Gaussian noise removal~\cite{venkatakrishnan2013plug,Romano2017,kawar2021stochastic}. These plug and play methods are non-blind and require knowledge of the likelihood $p(y|x,\theta')$ at test time.

\subsection{Training Strategies}
Generalization can also be improved by modifying the training set~\cite{Elman1993}. 
In the context of image restoration problems like denoising,~\cite{Gao_2017_ICCV} propose updating the training data sampling distribution each epoch to sample the data that the neural network performed worse on during the prior epoch preferentially, in an ad-hoc way. 
In~\cite{gnanasambandam2020one} the authors developed a principled adaptive training strategy by framing training a denoiser across many problem specifications as a minimax optimization problem. This strategy will be described in detail in Section~\ref{ssec:SampStrategy}.

\subsection{Relationship to Existing Works}
We go beyond~\cite{gnanasambandam2020one} by incorporating a polynomial approximation of the specification-loss landscape. This approximation is the key to scaling the adaptive training methodology to high-dimensional latent parameter spaces. It allows us to efficiently train a blind image denoiser that can operate effectively across a large range of noise conditions.

\section{Problem Formulation}
\subsection{Noise Model} \label{sec:noise_model}
This paper focuses on removing joint Poisson-Gaussian-Speckle noise using a single blind image denoising network. Such noise occurs whenever imaging scenes illuminated by a coherent (e.g.,~laser) source. In this context, photon/shot noise introduces Poisson noise, read noise introduces Gaussian noise, and the constructive and destructive interference caused by the coherent fields scattered off optically rough surfaces 
causes speckle noise~\cite{goodman2007speckle}.


The overall forward model can be described by 
\begin{align}\label{eqn:JointForwardModel}
    y_i=\frac{1}{\alpha}\text{Poisson}(\alpha (r \circ w_i))+n_i,
\end{align}
where the additive noise $n_i$ follows a Gaussian distribution $\mathcal{N}(0,\sigma^2\mathbf{I})$; the multiplicative noise $w_i$ follows a Gamma distribution with concentration parameter $B/\beta$ and rate parameter $B/\beta$, where $B$ is the upper bound on $\beta$; 
and $\alpha$ is a scaling parameter than controls the amount of Poisson noise. The forward model is thus specified by the set of latent variables
$\theta=\{\sigma,\alpha,\beta\}$.

\begin{figure*}[t]
\centering
\includegraphics[width=13.5cm]{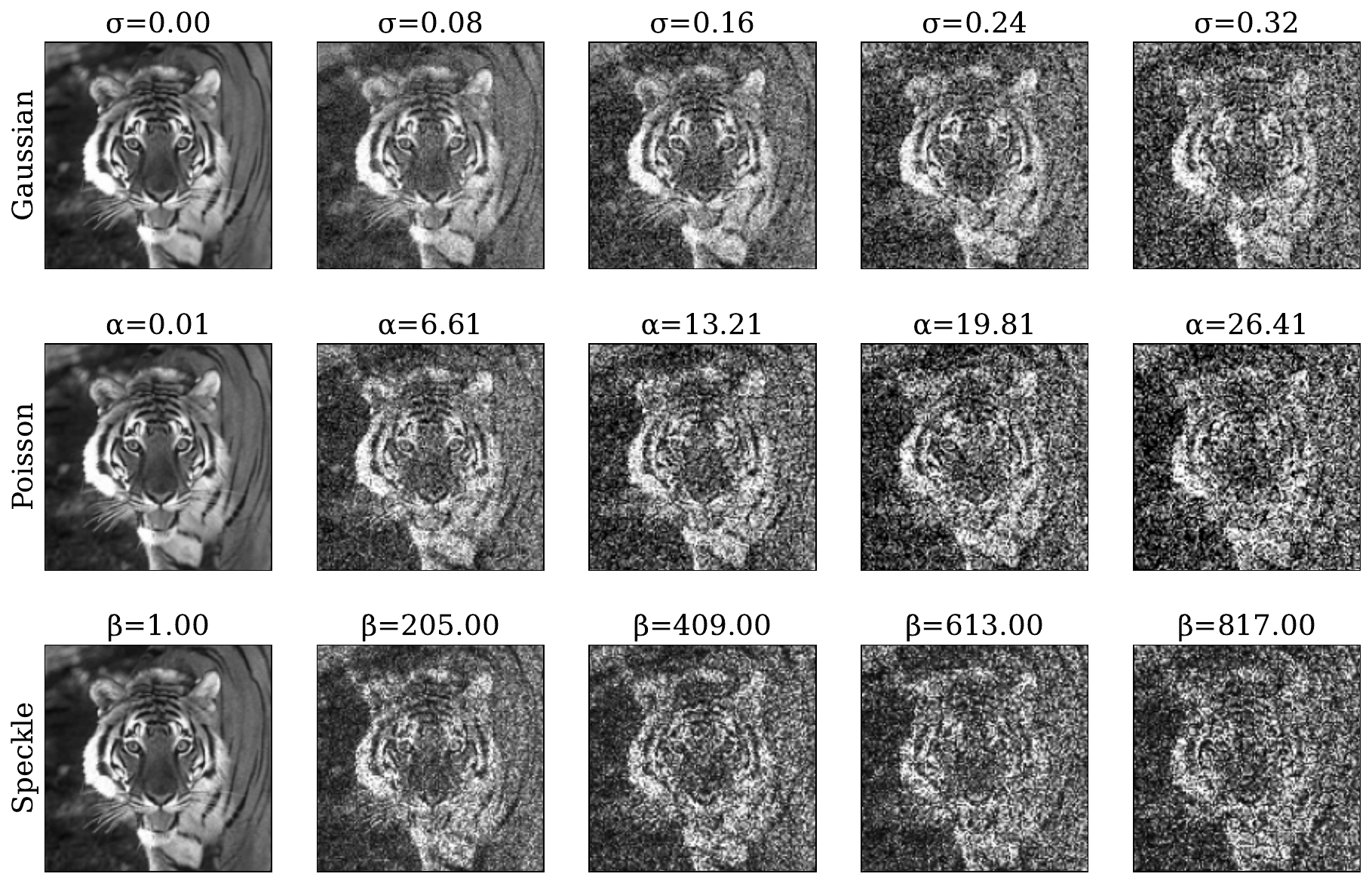}
\caption{{\bf Varying the noise specifications.} The first row shows images corrupted by Gaussian noise, the second row shows images corrupted by Poisson noise, and the last row shows images corrupted by speckle noise. In each of the rows, the other noise parameters are held fixed at $0$, $0.01$, and $1.00$, respectively.}
\label{fig:NoisyExamples}
\end{figure*}

A few example images generated according to this forward model are illustrated in Figure~\ref{fig:NoisyExamples}. Variations in the problem specifications results in drastically different forms of noise.

\subsection{Specification-Loss Landscape}
A \textit{specification} is a set of $n$ parameters that define a task.
In our setting, the specifications are the distribution parameters describing the noise in an image. 
Each of these parameters is bounded in an interval $[l_i, r_i]$, for $1 \leq i \leq n$. The \textit{specification space} $\Theta$ is the Cartesian product of these intervals: $\Theta = [l_1, r_1] \times \dots \times [l_n, r_n]$. 
Suppose we have a function $f$ that can solve a task (e.g.,~denoising) at any specification in $\Theta$, albeit with some error. 
Then the \textit{specification-loss landscape}, for a given $f$ over $\Theta$, is the function $\mathcal{L}_f$ which maps points $\theta$ from $\Theta$ to the corresponding error/loss (e.g.,~mean squared error) that $f$ achieves at that specification.

Now suppose that all functions $f$ under consideration 
come from some family of functions $\mathcal{F}$. 
Let the ideal function from $\mathcal{F}$ that solves a task at a particular specification $\theta$ be $f^{\theta}_{\text{ideal}} = \arg\min_{f \in \mathcal{F}} \mathcal{L}_f(\theta)$.
With this in mind, we define the \textit{ideal specification-loss landscape} as the function that maps points $\theta$ in $\Theta$ to the loss that $f^{\theta}_{\text{ideal}}$ achieves on the task with specification $\theta$, and denote it $\mathcal{L}_{\text{ideal}}$.  

\subsection{The Uniform Gap Problem}
Our goal is to find a single function $f^* \in \mathcal{F}$ that achieves consistent performance across the specification space $\Theta$, compared to the ideal function at each point $\theta \in \Theta$, $f^{\theta}_{\text{ideal}}$. 
More precisely, we want to minimize the maximum gap in performance between $f^*$ and $f^{\theta}_{\text{ideal}}$ across all of $\Theta$. 
E.g., we want a universal denoiser $f^*$  that works {\em almost} as well as specialized denoisers $f^{\theta}_{\text{ideal}}$, at all noise levels $\Theta$. 
Following~\cite{gnanasambandam2020one}, we can frame this objective as the following optimization problem 
\begin{equation} \label{eq:prob}
    f^* = \argmin_{f \in \mathcal{F}} \sup_{\theta \in \Theta} \left \{ \mathcal{L}_f(\theta) - \mathcal{L}_{\text{ideal}}(\theta)\right\},
\end{equation}
which we call the \textit{uniform gap problem}. 

\section{Proposed Method}\label{ssec:SampStrategy}
\subsection{Adaptive Training}
To solve the optimization problem given in \eqref{eq:prob},~\cite{gnanasambandam2020one} proposes rewriting it in its Lagrangian dual formulation and then applying dual ascent, which yields the following iterations: 
\begin{align}
    f^{t+1} &= \argmin_{f\in\mathcal{F}}\left\{\int_{\theta \in \Theta} \mathcal{L}_f(\theta)\lambda^t(\theta) d\theta\right\} \label{eq:0}\\
    \lambda^{t + \frac{1}{2}}(\theta) &= \lambda^t(\theta) + \gamma^t\left(\frac{\mathcal{L}_{f^{t+1}}}{\mathcal{L}_{\text{ideal}}} - 1\right) \label{eq:1}\\
    \lambda^{t+1}(\theta) &= \lambda^{t+\frac{1}{2}}(\theta) / \int_{\theta \in \Theta} \lambda^{t + \frac{1}{2}}(\theta)d\theta, \label{eq:2}
\end{align}
where $\lambda(\theta)$ represents a dual variable at specification $\theta \in \Theta$, and $\gamma$ is the dual ascent step size. 
One detail is we use PSNR constraints instead of MSE constraints, which influences the form of Equation~\ref{eq:1}. Intuitively, the reason is because PSNR is the logarithm of the MSE loss, and a more uniform PSNR gap means that the ratio of the losses is closer to 1, which is where the $\frac{\mathcal{L}_{f^{t+1}}}{\mathcal{L}_{\text{ideal}}} - 1$ term comes from. More details are discussed in the supplement.  
We can interpret \eqref{eq:0} as fitting a model $f$ to the training data, where $\lambda(\theta)$ is the probability of sampling a task at specification $\theta$ to draw training data from. Next,~\eqref{eq:1} updates the sampling distribution $\lambda(\theta)$ based on the difference between the current model $f^{t+1}$'s performance across $\theta \in \Theta$ and the ideal models' performances. 
Lastly \eqref{eq:2} ensures that $\lambda(\theta)$ is a properly normalized probability distribution. 
We provide the derivation of the dual ascent iterations from~\cite{gnanasambandam2020one} in the supplement.

While $\Theta$ has been discussed thus far as a continuum, in practice we sample $\Theta$ at discrete locations and compare the model being fit to the ideal model performance at these discrete locations only, so that the cardinality of $\Theta$, $|\Theta|$, is finite. 
Computing $\mathcal{L}_{\text{ideal}}(\theta)$ for each $\theta \in \Theta$ is extremely computationally demanding if $|\Theta|$ is large; if $f$ is a neural network, it becomes necessary to train $|\Theta|$ neural networks. 
Furthermore, while $\mathcal{L}_{\text{ideal}}(\theta)$ can be computed offline independent of the dual ascent iterations, during the dual ascent iterations, each update of $\lambda$ requires the evaluation of $\mathcal{L}_{f^{t+1}}$ for each $\theta \in\Theta$, which is also time intensive if $|\Theta|$ is large.

The key insight underlying our work is that one can approximate $\mathcal{L}_{\text{ideal}}$ and $\mathcal{L}_{f}$ in order to drastically accelerate the training process.

\begin{figure}
\centering
\includegraphics[width=.5\textwidth]{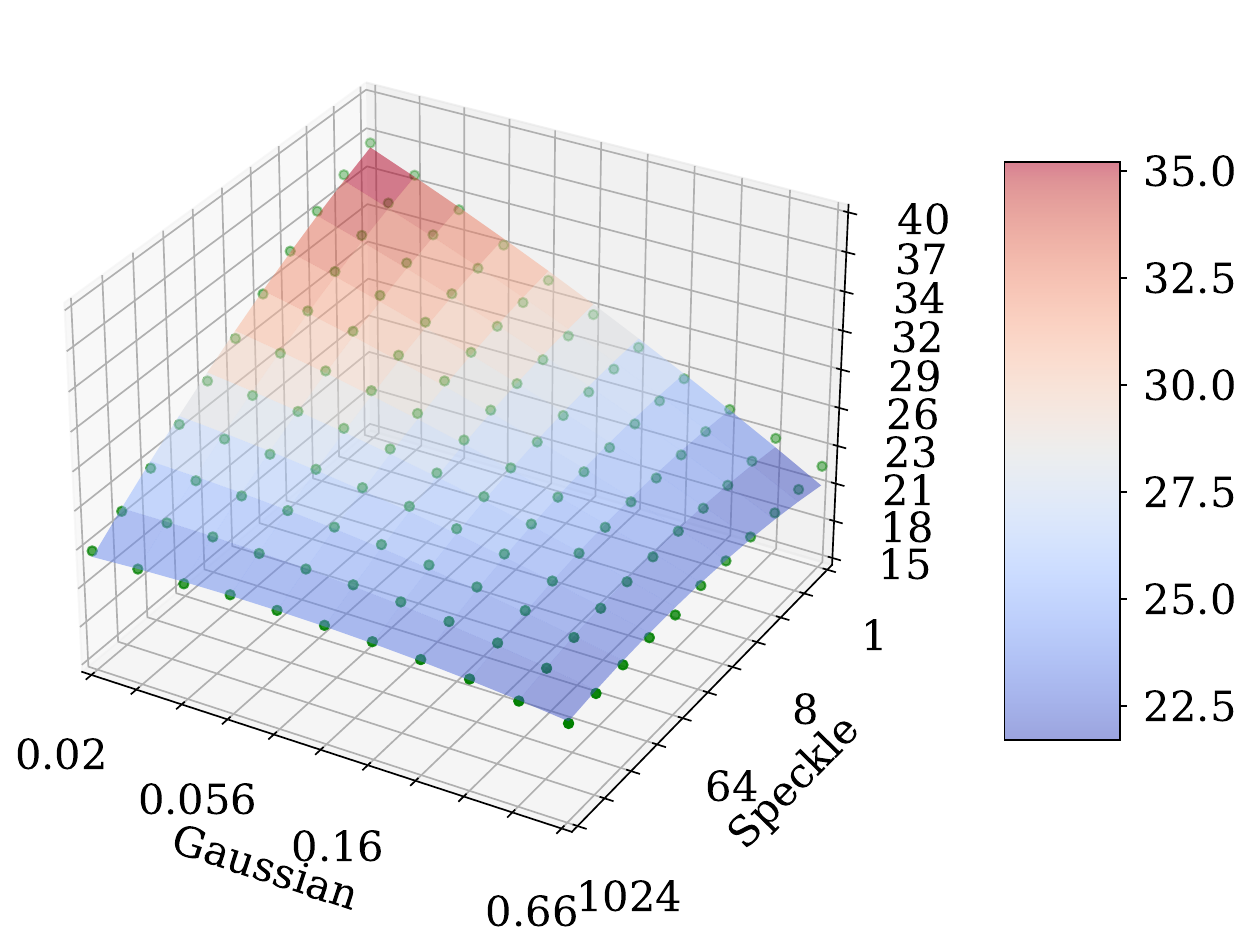}
\caption{\textbf{Loss Landscape Visualizations.} PSNR, which we use as our metric for error, versus denoising task specifications. The specification-loss landscapes (which represent the PSNRs a specialized denoiser can achieve at each specification) are smooth and amenable to approximation.
}
\label{fig:LossLandscapeApprox}
\end{figure}

\subsection{Specification-loss Landscape Approximations}
Let $\mathcal{Q}$ be a class of functions (e.g.,~quadratics) which we will use to approximate the specification-loss landscape.
Instead of computing $\mathcal{L}_{\text{ideal}}(\theta)$ for each $\theta \in \Theta$, we propose instead computing $\mathcal{L}_{\text{ideal}}(\theta)$ at a set of locations $\theta \in \Theta_{\text{sparse}}$, where $|\Theta_{\text{sparse}}| \ll |\Theta|$, and then using these values to form an approximation $\tilde{\mathcal{L}}_{\text{ideal}}$ of $\mathcal{L}_{\text{ideal}}(\theta)$, as 
\begin{equation}
   \tilde{\mathcal{L}}_{\text{ideal}} = \argmin_{\tilde{\mathcal{L}} \in \mathcal{Q}} \sum_{\theta \in \Theta_{\text{sparse}}} ||\tilde{\mathcal{L}}(\theta) - \mathcal{L}_{f_{\text{ideal}}}(\theta)||_2^2.
\end{equation}
We can similarly approximate $\mathcal{L}_{f^{t+1}}$ with a polynomial $\tilde{\mathcal{L}}_{f^{t+1}}$. 
Then we can solve \eqref{eq:prob} using dual ascent as before, replacing $\mathcal{L}_{\text{ideal}}$ and $\mathcal{L}_{f^{t+1}}$ with $\tilde{\mathcal{L}}_{\text{ideal}}$ and $\tilde{\mathcal{L}}_{f^{t+1}}$ where appropriate, resulting in a modification to \eqref{eq:1}:
\begin{equation}
    \lambda^{t + \frac{1}{2}} = \lambda^t + \gamma^t\left(\frac{\tilde{\mathcal{L}}_{f^{t+1}}}{\tilde{\mathcal{L}}_{\text{ideal}}} - 1\right).
\end{equation}


To justify our use of this approximation, we show that the specification loss-landscapes of the linear subspace projection ``denoiser'' and the soft-thresholding denoiser are linear and polynomial with respect to their specifications, respectively, and are thus both easy to approximate using polynomials. 

First we define our noise model. Let $y=\alpha \text{Poisson}(\frac{1}{\alpha}x_o)+n$ with $n\sim N(0,\sigma^2\mathbf{I})$. First note that if $\frac{1}{\alpha}x_o$ is large the distribution of $\alpha\text{Poisson}(x/\alpha)$ can be approximated with $N(x,\alpha\text{diag}(x))$. Accordingly, $y\approx x+\nu$ where $\nu\sim N(0,\sigma^2\mathbf{I} + \alpha \text{diag}(x))$.

\begin{example}\label{ex:subspace} 
Let $C$ denote a $k$-dimensional subspace of $\mathbb{R}^n$ ($k< n$), and the denoiser be the projection of $y$ onto subspace $C$ denoted by $P_C(y)=\mathbf{P} y$. 
Then, assuming $\frac{1}{\alpha}x_o$ is large, for every $x_o \in C$
\[
\mathbb{E} \| P_C(y) - x_o\|^2_2 \approx k \sigma^2+\alpha tr(\mathbf{P}\text{diag}(x_o)\mathbf{P}^t),
\]
where $tr(\cdot)$ denotes the trace. The loss landscape, $\mathcal{L}(\sigma^2,\alpha)=\mathbb{E} \| P_C(y) - x_o\|^2_2$, is linear with respect to $\sigma^2$ and $\alpha$.

\begin{proof}
 Since the projection onto a subspace is a linear operator and since $P_C(x_o) = x_o$ we have
\[
\mathbb{E} \| P_C(y) - x_o \|_2^2\approx \mathbb{E} \| x_o + P_C( \nu) - x_o \|_2^2= \mathbb{E} \|{P}_C(\nu)\|_2^2.
\]

Let $r=\mathbf{P}\nu$. Note that $r\sim N(0,\mathbf{\Sigma})$ with $\mathbf{\Sigma}=\sigma^2\mathbf{P}\mathbf{P}^t+\alpha \mathbf{P}\text{diag}(x_o)\mathbf{P}^t$. Accordingly,

\begin{align}
\mathbb{E} \|{P}_C(\nu)\|_2^2&=\mathbb{E}\|r\|^2=tr(\mathbf{\Sigma}),\nonumber\\
&=\sigma^2tr(\mathbf{P}\mathbf{P}^t)+\alpha tr(\mathbf{P}\text{diag}(x_o)\mathbf{P}^t),\nonumber\\
&=k\sigma^2+\alpha tr(\mathbf{P}\text{diag}(x_o)\mathbf{P}^t),\nonumber
\end{align}
where the last equality follows from the fact that $\mathbf{P}\mathbf{P}^t=\mathbf{P}$ and the trace of a $k$-dimensional projection matrix is $k$.

\end{proof}

\end{example}

\begin{example}\label{ex:soft_thresh} 
Let $\Gamma_k$ denote the set of $k$-sparse vectors, $k \ll n$. Let the family of soft-thresholding denoisers be defined as $\eta_{\tau}(y)=(|y|-\tau)_+\text{sign}(y)$, for $\tau > 0$. Assume that the ground truth signal is bounded, or $||x_o||_{\infty} \leq c$, for some $c \in \mathbb{R}^+$. Then, assuming $\frac{1}{\alpha}x_o$ is large, for every $x_o \in \Gamma_k$
\begin{align*}
\mathbb{E} &\| \eta_{\tau}(y) - x_o\|^2_2 \leq k (\sigma^2 + c\alpha + \tau^2)\\
&+ 2(n-k)\left((\sigma^2 + \tau^2)\Phi\left(-\frac{\tau}{\sigma}\right) -\tau\sigma\phi\left(\frac{-\tau}{\sigma}\right)\right),    
\end{align*}
where $\phi(\cdot)$ denotes the Gaussian density function and $\Phi(\cdot)$ denotes the Gaussian distribution function. The loss landscape, $\mathcal{L}(\sigma^2,\alpha)=\mathbb{E} \| \eta_{\tau}(y) - x_o\|^2_2$, can be upper bounded closely by a polynomial with respect to $\sigma^2$ and $\alpha$, because $\phi$ and $\Phi$ are smooth functions with convergent Taylor series.
\begin{proof}
We have \begin{align*}
    \mathbb{E}||\eta_{\tau}(x_o+\nu )-x_o||_2^2 &= \sum_{i=1}^k\mathbb{E}(\eta_{\tau}(x_{o,i}+\nu_i)-x_{o,i})^2\\ 
    &+ (n-k)\mathbb{E}(\eta(\nu_n;\tau))^2
\end{align*}
$\mathbb{E}(\eta_{\tau}(x_{o,i}+\nu_i)-x_{o,i})^2$ is an increasing function of $x_{o,i}$, so we can bound its value from above by taking the limit as \begin{align*}
        \mathbb{E}(\eta_{\tau}(x_{o,i}+\nu_i)-x_{o,i})^2 &\leq \lim_{x_{o,i} \to \infty} \mathbb{E}(\eta_{\tau}(x_{o,i}+\nu_i)-x_{o,i})^2 \\
        &= \sigma^2 + c\alpha + \tau^2.
\end{align*} Switching the limit and expectation is justified by the dominated convergence theorem. To calculate the second term, first note that since the ground truth is $0$ here there is only Gaussian noise and no Poisson noise. Then, with $\beta = \frac{-\tau}{\sigma}$, we have 
\begin{align*}
    \mathbb{E}(\eta_{\tau}(\nu_n))^2 &= 2P(\nu_n \leq -\tau)(\mathbb{E}(\nu_i^2\mid\nu_n\leq -\tau) \\
        &+ 2\mathbb{E}(\nu_i\mid\nu_n\leq -\tau)\tau + \tau^2) \\
        &= 2\Phi\left(\beta\right)\left(\sigma^2\left(1-\beta \frac{\phi(\beta)}{\Phi(\beta)}\right) - 2\sigma\frac{\phi(\beta)}{\Phi(\beta)}\tau + \tau^2\right) \\
        &= 2\left((\sigma^2 + \tau^2)\Phi\left(-\frac{\tau}{\sigma}\right) -\tau\sigma\phi\left(\frac{-\tau}{\sigma}\right)\right)
\end{align*}
where we rewrite the expectations in terms of expressions for the first and second moments of the one-sided truncated Gaussian distribution. Combining the results of these two calculations gives us our desired conclusion.
\end{proof}
\end{example}

The specification-loss landscapes of high-performance image denoisers is highly regular as well.
We show the achievable PSNR (peak signal-to-noise ratio) versus noise parameter plots for denoising Gaussian-speckle noise with the DnCNN denoiser~\cite{zhang2017beyond} in Figure~\ref{fig:LossLandscapeApprox}. 
Analogues figures for Poisson-Gaussian and Poisson-speckle noise can be found in the supplement. 
These landscapes are highly regular and can accurately approximated with quadratic polynomials. (Details on how we validated these quadratic approximations using cross-validation can be found in the supplement.)

Because we're more interested in ensuring a uniform PSNR gap than a more uniform MSE gap, we approximate the ideal PSNRs rather than the ideal mean squared errors. 
Then, following~\cite{gnanasambandam2020one}, we convert the ideal PSNRs to mean squared errors with the mapping $\mathcal{L}(\theta) = 10^{-\text{PSNR}(\theta)/10}$ for use in the dual ascent iterations.

\subsection{Exponential Savings in Training-Time}


The key distinction between our adaptive training procedure and the adaptive procedure from~\cite{gnanasambandam2020one} is that~\cite{gnanasambandam2020one} relies upon a dense sampling of the specification-loss landscape whereas our method requires only a sparse sampling of the specification-loss landscape. Because ``sampling'' points on the specification-loss landscape requires training a CNN, sampling fewer points can result in substantial savings in time and cost.


As one increases the number of specifications, $n$, needed to describe this landscape ($n=1$ for Gaussian noise, $n=2$ for Poisson-Gaussian noise, $n=3$ for Poisson-Gaussian-Speckle noise, ...), the number points needed to densely sample the landscape grows exponentially. Fortunately, the number of samples needed to fit a quadratic to this landscape only grows quadratically with $n$: 
The number of possible nonzero coefficients, i.e.,~unknowns, of a quadratic of $n$ variables is ${n+2 \choose 2} = \frac{(n+1)(n+2)}{2}$ and thus one can uniquely specify this function from $\frac{(n+1)(n+2)}{2} + 1$ non-degenerate samples.
Accordingly, our method has the potential to offer training-time savings that are exponential with respect to the problem specification dimensions. 
In the next section, we demonstrate our method reduces training-time over $50\times$ for $n=3$.

\section{Synthetic Results}\label{sec:Results}

In this section we compare the performance of universal denoising algorithms that are trained  (1) by uniformly sampling the noise specifications during training (``uniform''); (2) using the adaptive training strategy from~\cite{gnanasambandam2020one} (``dense''), which adaptively trains based on a densely sampled estimate of the loss-specification landscape, and (3) using our proposed adaptive training strategy (''sparse''), which adaptively trains based on a sparsely sampled approximation of the loss-specification landscape. 
We do not compare against the simple noise level sampling rules such as the 80-20 rule mentioned in~\cite{gnanasambandam2020one} which oversamples lower noise levels because it is more difficult to define what region of the space of possible noise levels is considered high versus low noise when there are multiple noise types. 

\begin{figure*}
\centering
\includegraphics[width=.3\textwidth]{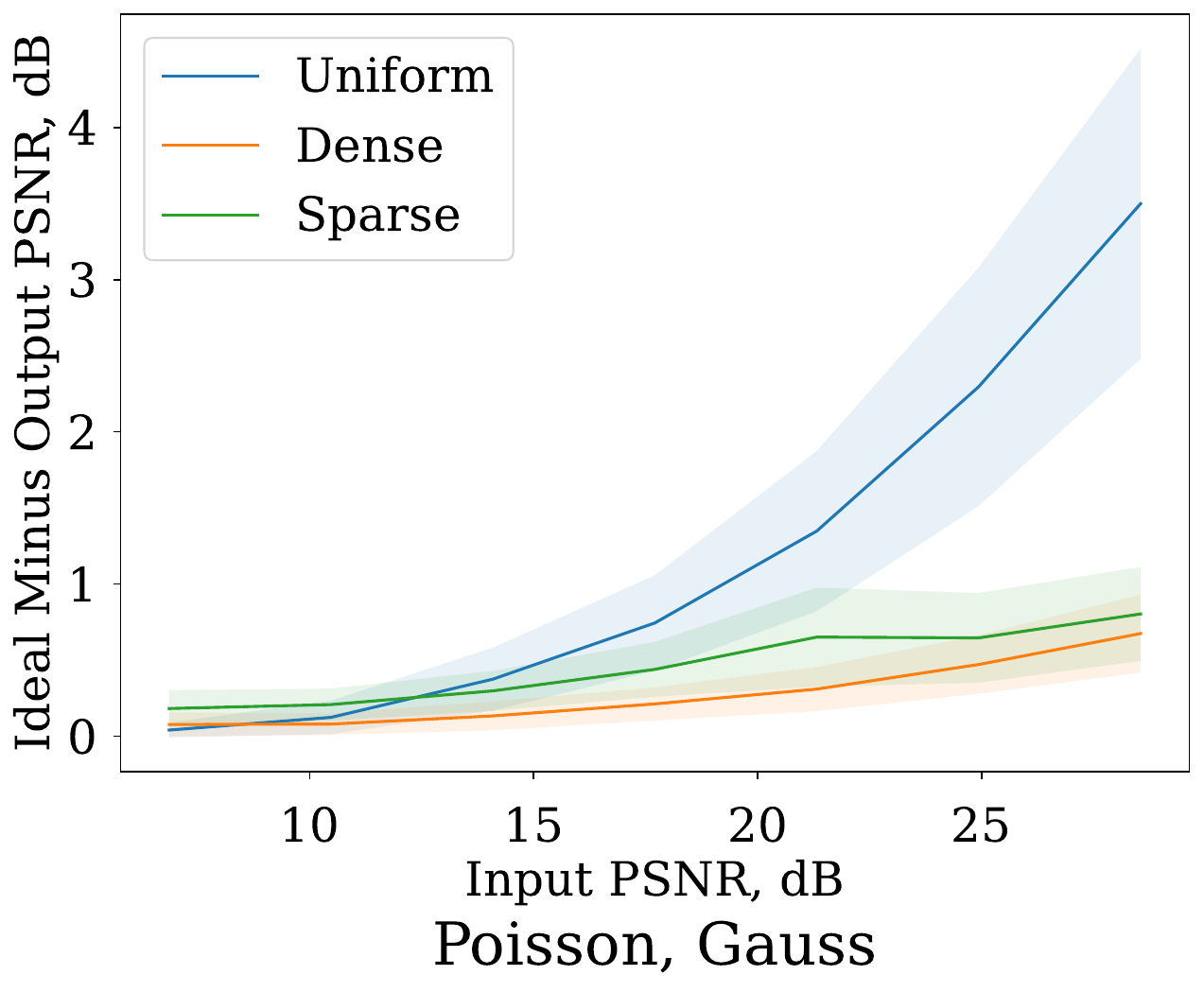}
\includegraphics[width=.3\textwidth]{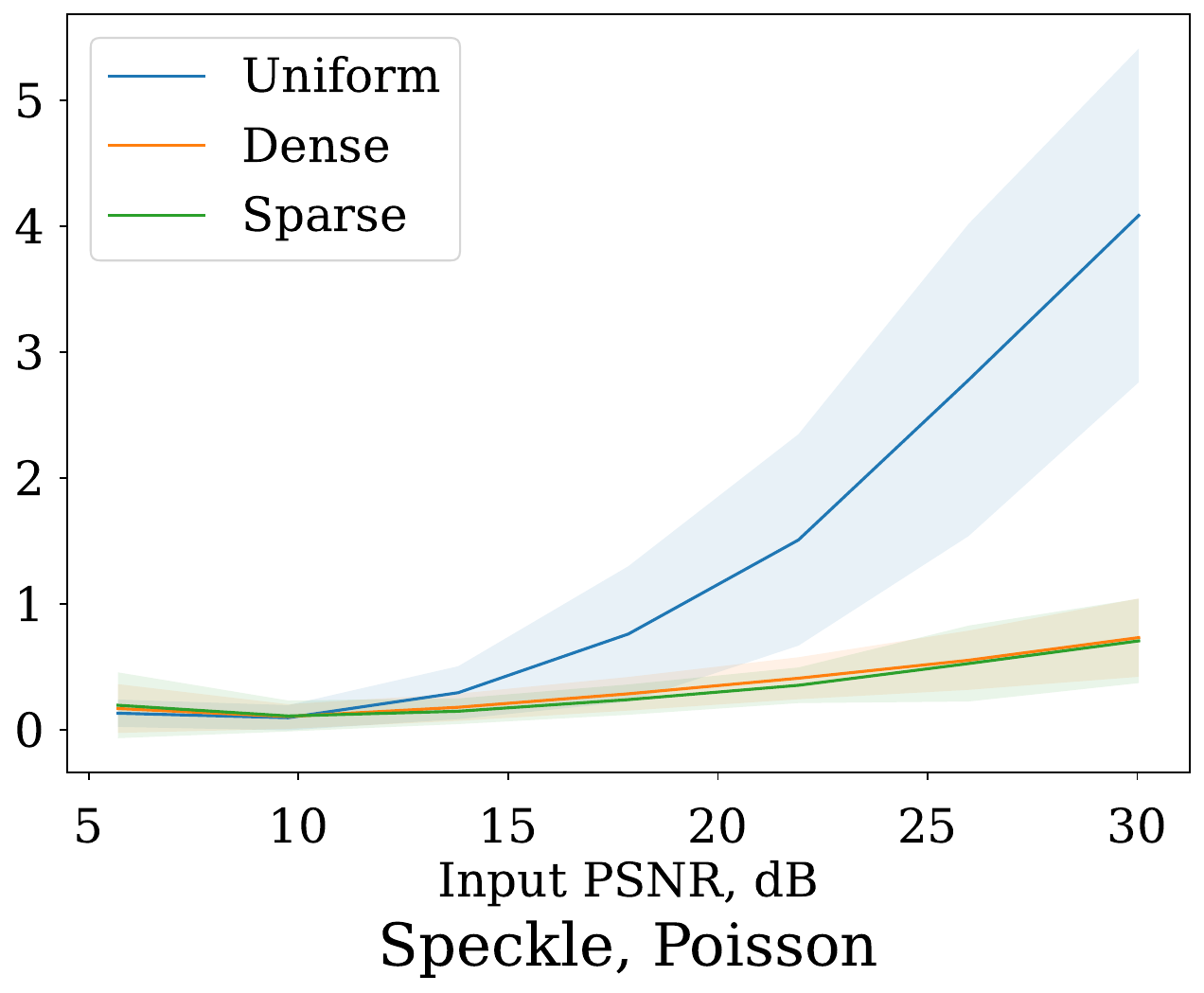}
\includegraphics[width=.3\textwidth]{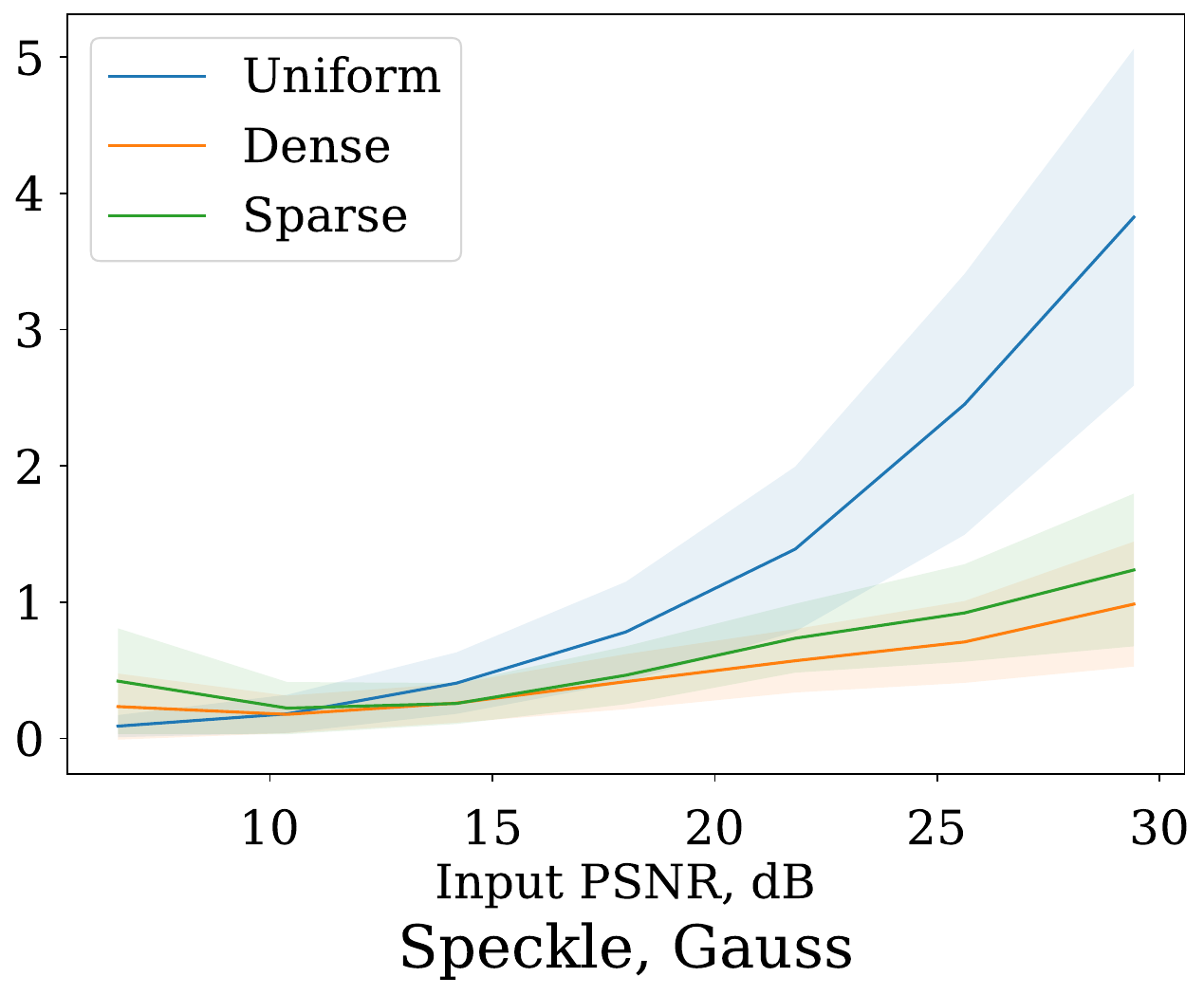}
\caption{ \textbf{Adaptive vs Uniform Training, 2D Specification Space.} Adaptive training with sparse sampling and the polynomial approximation works effectively in the 2D problem space and produces a network whose performance is consistently close to the ideal. By contrast, a network trained by uniformly sampling from the space performs far worse than the specialized networks in certain contexts. The error bars represent one standard-deviation. Lower is better.}

\label{fig:AdaptivevSUniform2D}
\end{figure*}

\begin{figure}
\centering
\includegraphics[width=.3\textwidth]{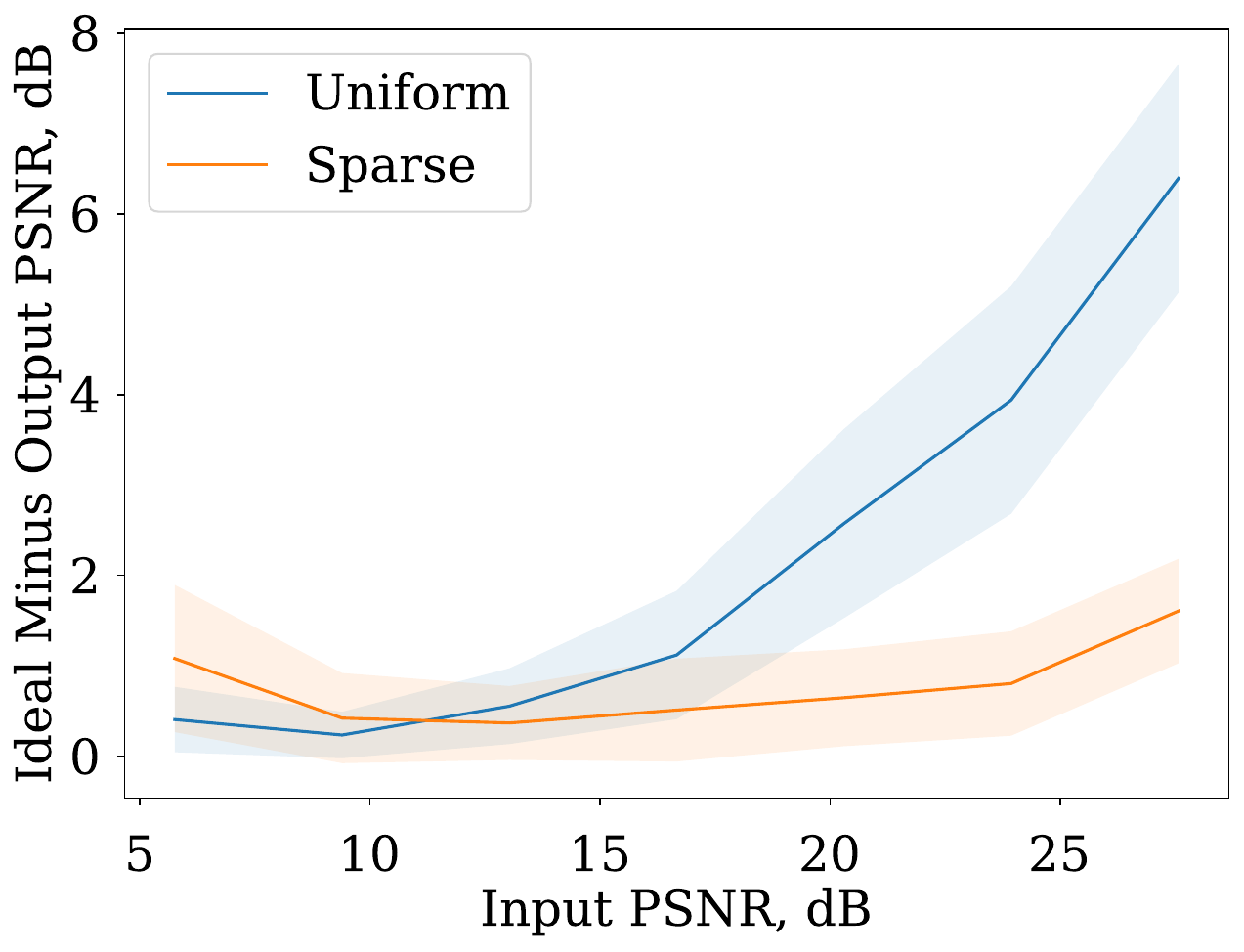}
\caption{ \textbf{Adaptive vs Uniform Training, 3D Specification Space.} Adaptive sampling with the polynomial approximation works effectively in the 3D problem space and produces a network whose performance is consistently close to the ideal. By contrast, a network trained by uniformly sampling from the space performs far worse than the specialized networks in certain contexts. The error bars represent one standard-deviation. Lower is better.}

\label{fig:AdaptiveSparseLossDelta3D}
\end{figure}

\begin{figure*}
\centering

\includegraphics[width=.24\textwidth]{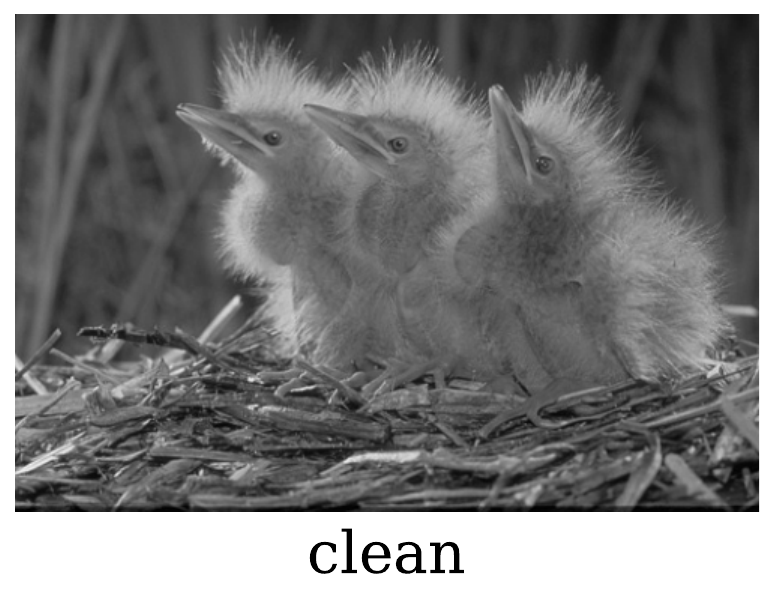}

\includegraphics[width=.26\textwidth]{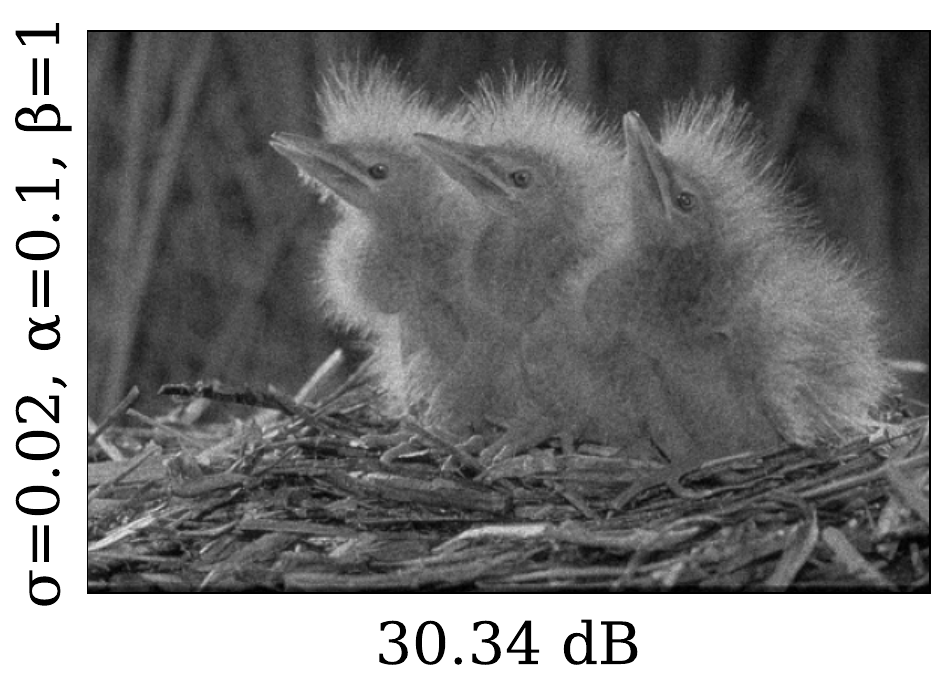}
\includegraphics[width=.24\textwidth]{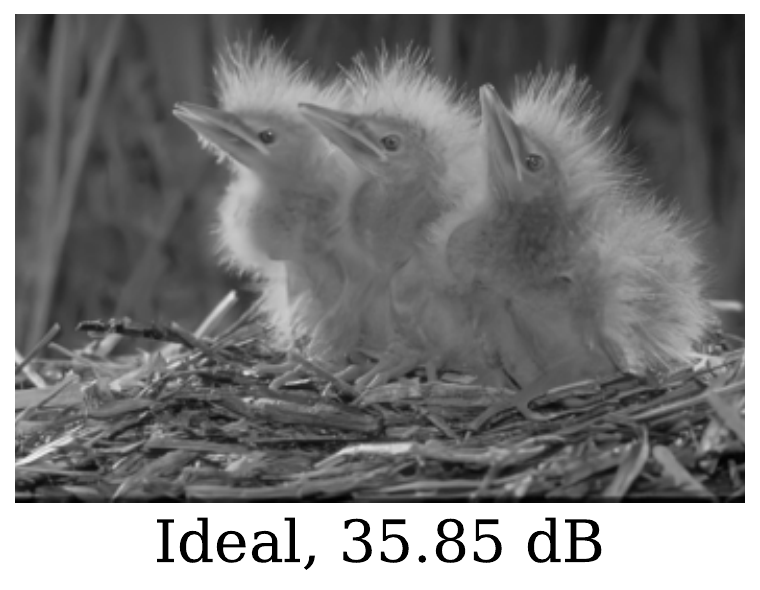} 
\includegraphics[width=.24\textwidth]{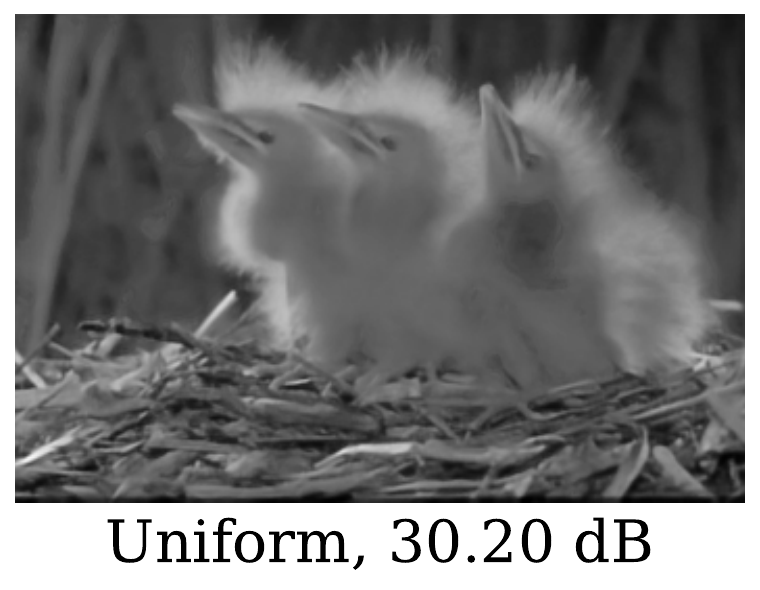}
\includegraphics[width=.24\textwidth]{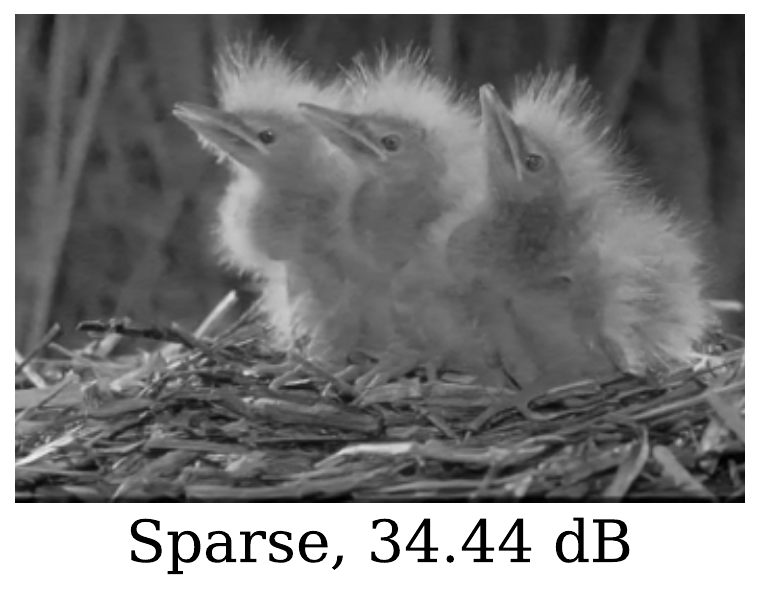}

\includegraphics[width=.26\textwidth]{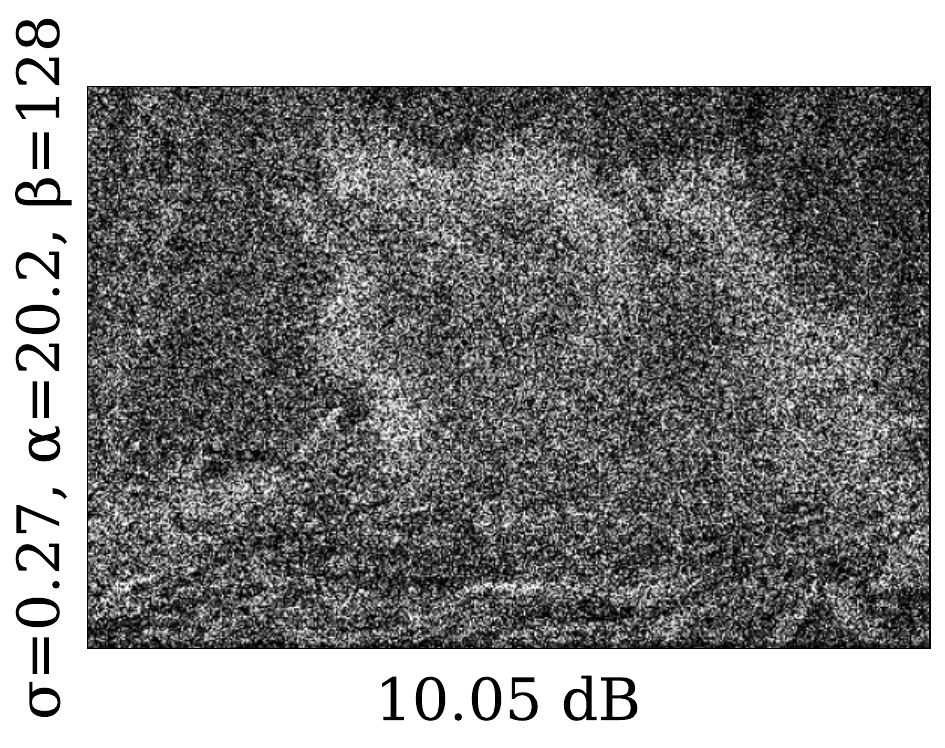}
\includegraphics[width=.24\textwidth]{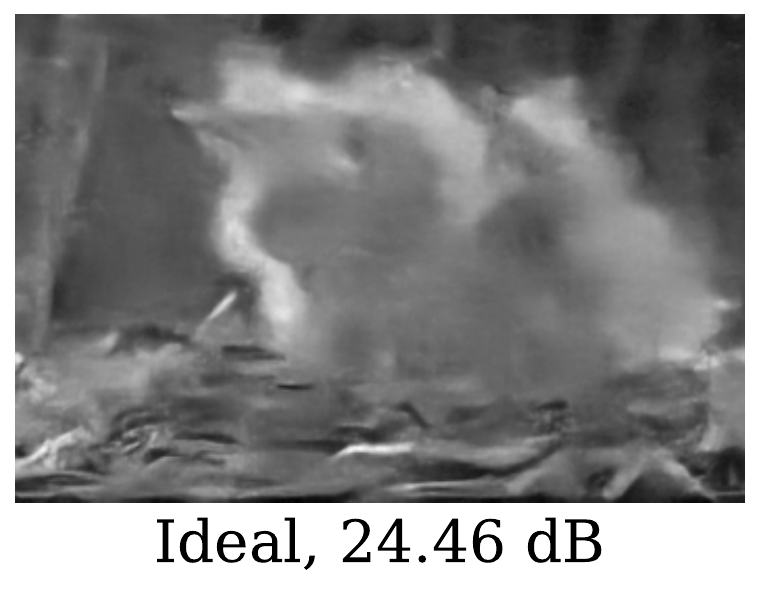} 
\includegraphics[width=.24\textwidth]{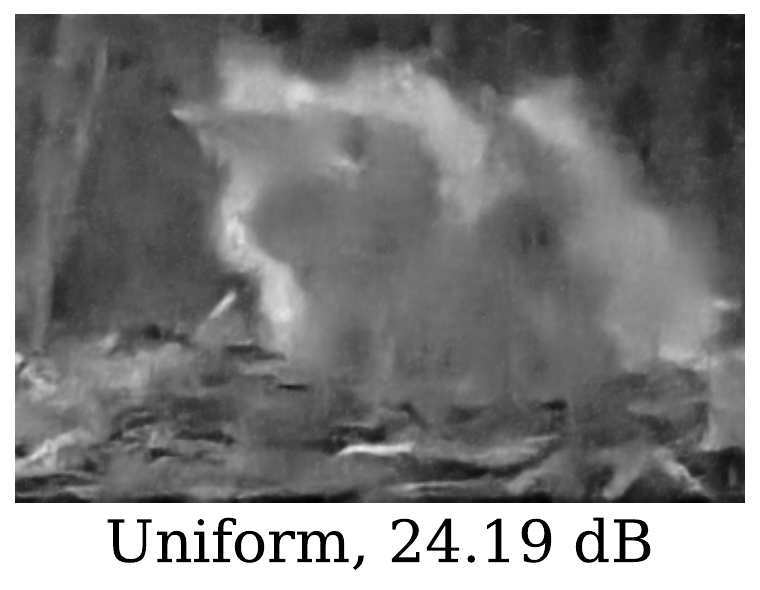}
\includegraphics[width=.24\textwidth]{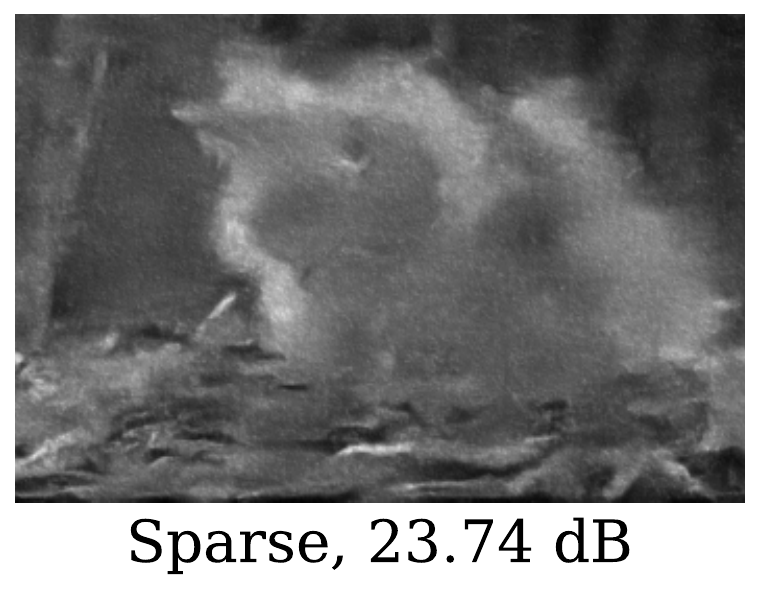}
\caption{ \textbf{Qualitative comparisons, simulated data.} Comparison between the performance of the ideal, uniform-trained, and adaptive distribution, sparse sampling-trained denoisers on a sample image corrupted with a low amount of noise and corrupted with a high amount of noise. Our adaptive distribution sparse approximation based blind training strategy performs only marginally worse than an ideal, non-blind baseline when applied to ``easy'' problem specifications, and significantly better than the uniform baseline, while also being only marginally worse than an ideal baseline and uniform baseline under ``hard'' problem specifications.}
\label{fig:qualitative}
\end{figure*}

\subsection{Setup}

\paragraph{Implementation Details.} \label{sec:impl}
We use a 20-layer DnCNN architecture~\cite{zhang2017beyond} for all our denoisers. 
Following~\cite{mohan2019robust}, we remove all biases from the network layers.
We train all of our networks for 50 epochs, with 3000 mini-batches per epoch and 128 image patches per batch, for a total of 384,000 image patches total. 
We use the Adam optimizer~\cite{DBLP:journals/corr/KingmaB14} to optimize the weights with a learning rate of \num{1e-4}, with an L2 loss. 
During adaptive training, we update the noise level sampling distribution every 10 epochs. 

\paragraph{Data} \label{sec:data}
To train our denoising models, we curate a high-quality image dataset that combines multiple high resolution image datasets: the Berkeley Segmentation Dataset~\cite{MartinFTM01}, the Waterloo Exploration Database~\cite{ma2017waterloo}, the DIV2K dataset~\cite{Agustsson_2017_CVPR_Workshops}, and the Flick2K dataset~\cite{Lim_2017_CVPR_Workshops}, which is the same as the dataset used in~\cite{dpir}, but different from the one used in~\cite{gnanasambandam2020one}. The choice of dataset is not critical to our method, so we use the dataset recommended by the later work,~\cite{dpir}.
To test our denoising models, we use the validation dataset from the DIV2K dataset.
We use a patch size of 40 pixels by 40 pixels, and patches are randomly cropped from the training images with flipping and rotation augmentations, to generate a total of 384000 patches.
All images are grayscale and scaled to the range $[0, 1]$. 
We use the BSD68 dataset~\cite{bsd68} as our testing dataset.

\paragraph{Noise Parameters.} We consider four types of mixed noise distributions: Poisson-Gaussian, Speckle-Poisson, Speckle-Gaussian, and Speckle-Poisson-Gaussian noise. 
In each mixed noise type, $\sigma \in  [0.02, 0.66]$, $\alpha \in [0.1, 41]$, and $\beta \in [1, 1024]$, and we discretize each range into 10 bins. 
Note that $B=1024$ for speckle noise, following the parameterization in Section \ref{sec:noise_model}.
These ranges were chosen as they correspond to input PSNRs of roughly 5 to 30 dB. 
When training the networks with uniform sampling, a new noise specification is first sampled, and then an instance of the noise is sampled from that distribution. The number of training instances is the same as for sparse training, but the proportion of noise distributions seen is different (given by the noise distribution parameter sampling distribution).

\paragraph{Sampling the Specification-Loss Landscape.}
Both our adaptive universal denoiser training strategy (``sparse'') and Gnanasambandam and Chan's adaptive universal universal denoiser training strategy (``dense'') require sampling the loss-specification landscapes, $\mathcal{L}_{\text{ideal}}(\theta)$, before training.
That is, each requires training a collection of denoisers specialized for specific noise parameters $\theta\in\Theta$.

For the loss-specifications landscapes with two dimensions (Poisson-Gaussian, Speckle-Poisson, and Speckle-Gaussian noise), we densely sampling the landscape (i.e.,~train networks) on a $10\times10$ grid for the ``dense'' training. For the ``sparse'' training (our method) we sample the landscape at 10 random specifications as well as at the specification support's endpoints, for a total of 14 samples for Poisson-Gaussian, Speckle-Poisson, and Speckle-Gaussian noise.

For the loss-specifications landscape with three dimensions (Speckle-Poisson-Gaussian), sampling 
densely would take nearly a year of GPU hours, so we restrict ourselves to sparsely sampling at only 18 specifications: 10 random locations plus the 8 corners of the specification cube. 



\paragraph{Training Setup.} For each Poisson-Gaussian, Speckle-Poisson, and Speckle-Gaussian noises separately, we (1) use approximations $\tilde{\mathcal{L}}_{\text{ideal}}$ and $\tilde{\mathcal{L}}_{f}$ to adaptively train a denoiser $f^*_{\text{sparse}}$, (2) use densely sampled landscapes $\mathcal{L}_{\text{ideal}}$ and $\mathcal{L}_{f}$ to adaptively train a denoiser $f^*_{\text{dense}}$, and (3) sample specifications uniformly to train a denoiser $f^*_{\text{uniform}}$.  
For Speckle-Poisson-Gaussian noise, the set of possible specifications is too large to train an ideal denoiser for each specification (i.e.,~compute $\mathcal{L}_{\text{ideal}}$) and so we only provide results for $f^*_{\text{sparse}}$ and $f^*_{\text{uniform}}$.




\begin{table}[t]
\begin{center}
\begin{tabular}{ |c|c||c|c|c|c| }
 \hline
   Poisson & Gaussian & Ideal & Uniform & Dense & Sparse \\ 
 \hline
	0.1 & 0.02 & 35.3 &	31.4 & 34.5 & 34.6\\ 
   0.1 & 0.66 & 22.0&	22.0&	21.9&	21.9\\ 
   41 & 0.02 &22.9	&22.9	&22.9&	22.9\\ 
   41 & 0.66 & 21.4&	21.3&	21.3&	21.3\\ 
      2.0 & 0.11 &27.1&	26.6&	27.0&	27.0\\ 
 \hline
\end{tabular}
\vspace{3pt}
\caption{Quantitative comparison of methods on Poisson-Gaussian noise sampled at various levels, using PSNR (dB). } \label{table:poiss_gauss_perf}

\begin{tabular}{ |c|c||c|c|c|c| } 
 \hline
   Speckle & Gaussian & Ideal & Uniform & Dense & Sparse \\ 
 \hline
	1.0 & 0.02 & 36.6	&32.0	&35.4&	35.1\\ 
   1.0 & 0.66 & 22.0&	21.9&	21.7&	21.6\\ 
   1024 & 0.02 &23.1&	23.1&	22.6&	22.4\\ 
   1024 & 0.66 & 21.5&	21.4&	20.8&	20.7\\ 
   32 & 0.11 &27.4	&27.0	&27.2&	27.2\\ 
 \hline
\end{tabular}
\vspace{3pt}
\caption{Quantitative comparison of methods on Speckle-Gaussian noise sampled at various levels, using PSNR (dB).}\label{table:speckle_gauss_perf}

\begin{tabular}{ |c|c||c|c|c|c| } 
 \hline
   Speckle & Poisson & Ideal & Uniform & Dense & Sparse \\ 
 \hline
	1.0&0.1 &36.1&	31.5&	35.3&	35.3\\ 
 1.0 & 41  & 23.0&	22.9&	22.9&	22.9\\ 
  1024& 0.1 &23.0&	23.1&	23.0&	22.9\\ 
  1024& 41  & 22.0&	21.8&	21.6&	21.6\\ 
   32 & 2.0  &27.8&	27.3&	27.6&	27.7\\ 
 \hline
\end{tabular}
\vspace{3pt}
\caption{Quantitative comparison of methods on Speckle-Poisson noise sampled at various levels, using PSNR (dB).}\label{table:speckle_poiss_perf}

\begin{tabular}{ |c|c|c||c|c|c| } 
 \hline
  Speckle & Poisson & Gaussian & Ideal & Uniform  & Sparse \\ 
 \hline
1 &	0.1 &	0.02 &	34.7&	29.3	&	33.4\\ 
1 &	0.1 &	0.66 &	22.0&	21.9&		21.5\\ 
1 &	41 &	0.02 &	23.0&	22.9&		22.7\\ 
1 &	41 &	0.66 &	21.4&	21.3&		20.8\\ 
1024 &	0.1 &	0.02 &	23.2&	23.2&		22.6\\ 
1024 &	0.1 &	0.66 &	21.5&	21.4&		20.7\\ 
1024 &	41 &	0.02 &	22.0&	21.9&		21.2\\ 
1024 &	41 &	0.66 &	21.0&	20.8&		20.0\\ 
64 &	2.3 &	0.54 &25.8&	25.5&		25.5\\ 
 \hline
\end{tabular}
\vspace{3pt}
\caption{Quantitative comparison of methods on Speckle-Poisson-Gaussian noise sampled at various levels, using PSNR (dB).}\label{table:speckle_poiss_gauss_perf}
\end{center}
\end{table}

\subsection{Quantitative Results} 
Tables~\ref{table:poiss_gauss_perf},~\ref{table:speckle_gauss_perf},~\ref{table:speckle_poiss_perf}, and~\ref{table:speckle_poiss_gauss_perf} compare the performance of networks trained with our adaptive training method, which uses sparse samples to approximate of the specification-loss landscape; ``ideal'' non-blind baselines, which are trained for specific noise parameters; networks trained using the adaptive training procedure from~\cite{gnanasambandam2020one}, which requires training specialized ideal baselines at all noise specifications (densely) beforehand; and networks trained by uniformly sampling the specification-space. Both adaptive strategies approach the performance of the specialized networks and dramatically outperform the uniformly trained networks at certain problem specifications.

These results are further illustrated in Figures~\ref{fig:AdaptivevSUniform2D} and~\ref{fig:AdaptiveSparseLossDelta3D}, which report how much the various universal denoisers underperform specialized denoisers across various noise parameters. As hoped, both adaptive training strategies (``dense'' and ``sparse'') produce denoisers which consistently perform nearly as well as the specialized denoisers.\footnote{To form these figures we train specialized networks at an additional 100 specifications. These networks were not used for training the universal denoisers.} 
In contrast, the uniformly trained denoisers underperform the specialized denoisers in some contexts. 
Additional quantitative results can be found in the Supplement.

%


\subsection{Qualitative Results} 
Figure \ref{fig:qualitative} illustrates the denoisers trained with the different strategies (ideal, uniform, adaptive) on an example corrupted with a low amount of noise and an example corrupted with a high amount of noise.
Notice that in the low-noise regime the uniform trained denoiser oversmooths the image so achieves worse performance than the adaptive trained denoiser, whereas in the high noise regime the uniform trained denoiser outperforms the adaptive trained denoiser. 
Additional qualitative results can be found in the supplement.

\subsection{Time and Cost Savings} 

Recall our adaptive training strategy requires pretraining significantly fewer specialized denoisers than the method developed in~\cite{gnanasambandam2020one}: 14 vs 100 networks with 2D noise specifications and 18 vs 1000 networks with 3D specifications.
We trained the DnCNN denoisers on Nvidia GTX 1080Ti GPUs, which took 6--8 hours to train each network. 
Overall training times associated with each of the universal denoising algorithms are presented in Table~\ref{table:training_times}. In the 3D case, our technique saves over 9 months in GPU compute hours.

\begin{table}[h] 
\begin{center}
\begin{tabular}{ |c|c|c|c| } 
 \hline
 & Uniform & Dense & Sparse \\
 \hline
 Poisson-Gauss & 6.5 & 891.6  & 95.2 \\
 Speckle-Gauss & 7.8 & 747.8 & 78.8 \\
 Speckle-Poisson & 8.0 & 655.9  & 70.8\\
 Speckle-Poisson-Gauss & 7.5 & $>$7000 (predicted) &138.6\\
 \hline
\end{tabular}
\vspace{3pt}
\caption{Comparing the overall training times (in GPU hours) between the various universal denoiser training strategies. Sparsely sampling and approximating the specification-loss results in orders of magnitude savings in training time and cost.
}\label{table:training_times}
\end{center}
\end{table}

\section{Experimental Validation}
To validate the performance of our method, we experimentally captured a small dataset consisting of 5 scenes with varying amounts of Poisson, Gaussian, and speckled-speckle noise.

\subsection{Optical Setup}

\begin{figure}
\centering
\includegraphics[width=.45\textwidth]{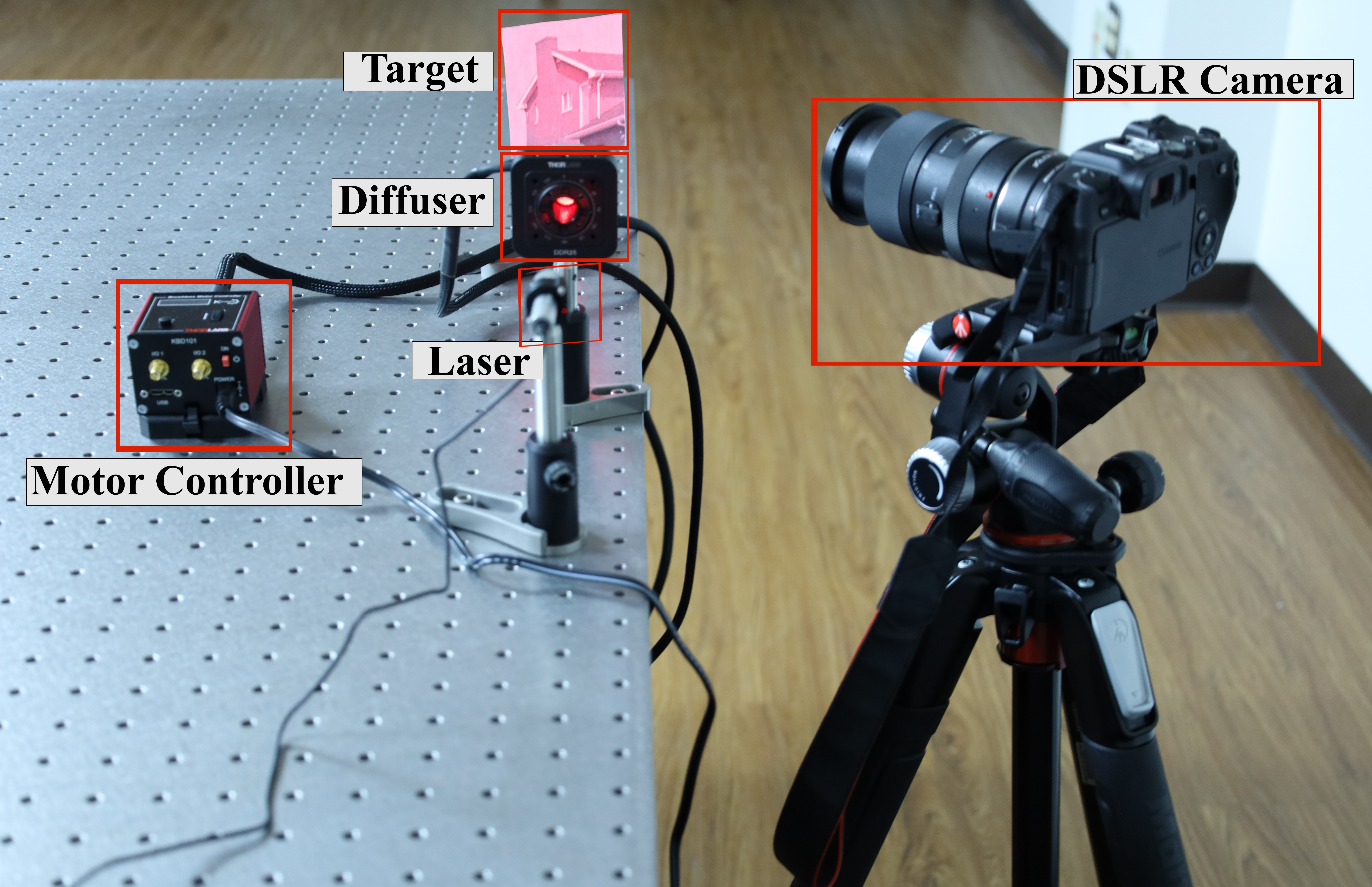}
\caption{ \textbf{Optical System.} We capture images of paper with images on them, illuminated by a laser, using a DSLR camera. The laser beam passes through a diffuser which results in a speckle noise pattern on the image. We rotate the diffuser with a rotation mount connected to a motor controller to gather different independent realizations of speckle.}
\label{fig:optical_system}
\end{figure}

Our optical setup consists of a bench-top setup and a DSLR camera mounted on a tripod, pictured in ~\ref{fig:optical_system}. A red laser is shined through a rotating diffuser to illuminate an image printed on paper. The diffuser is controlled by an external motor controller, which allows us to capture a distinct speckle realization for each image. The resulting signal is captured by the DSLR camera. The pattern on the rotating diffuser causes a speckle noise pattern in the signal. Note, however, that because the paper being imaged is rough with respect to the wavelength of red light, an additional speckle pattern is introduced to the signal---we're capturing speckled-speckle. This optical setup allows us to capture images with varying levels of Gaussian, Poisson, and (speckled) speckle noise. The Gaussian noise can be varied by simply increasing or decreasing the ISO on the camera. Similarly, decreasing exposure time will produce more Poisson noise. Lastly, speckle noise can be varied by changing the number of recorded realizations that are averaged together; as the number of realizations increases, the corruption caused by speckle noise decreases.
For each noisy capture, we keep the aperture size fixed at F/11. 


\subsection{Data Collection}

Using the previously described optical setup, we collected noisy and ground truth images for five different targets. 
To compute our ground truth image, we capture 16 images with a shutter speed of 0.4 seconds, ISO 100, and F/4 and average them together.
To gather the noisy images, for each of the 5 scenes we consider, we capture 16 images at 4 different shutterspeed/ISO pairs, at a fixed aperture of F/11: 1/15 seconds/ISO 160, 1/60 seconds/ISO 640, 1/250 seconds/ISO 2500, and 1/1000 seconds/ISO 10000, for a total of 320 pictures captured. 
We choose these pairs such that the product of the two settings is approximately the same across settings and thus dynamic range is approximately the same across samples. 
Additionally, we capture dark frames for each exposure setting and subtract them from the corresponding images gathered. 
The noisy images were scaled to match the exposure settings of the ground truth image.

\subsection{Performance}

\begin{figure}
\centering
\includegraphics[width=.3\textwidth]{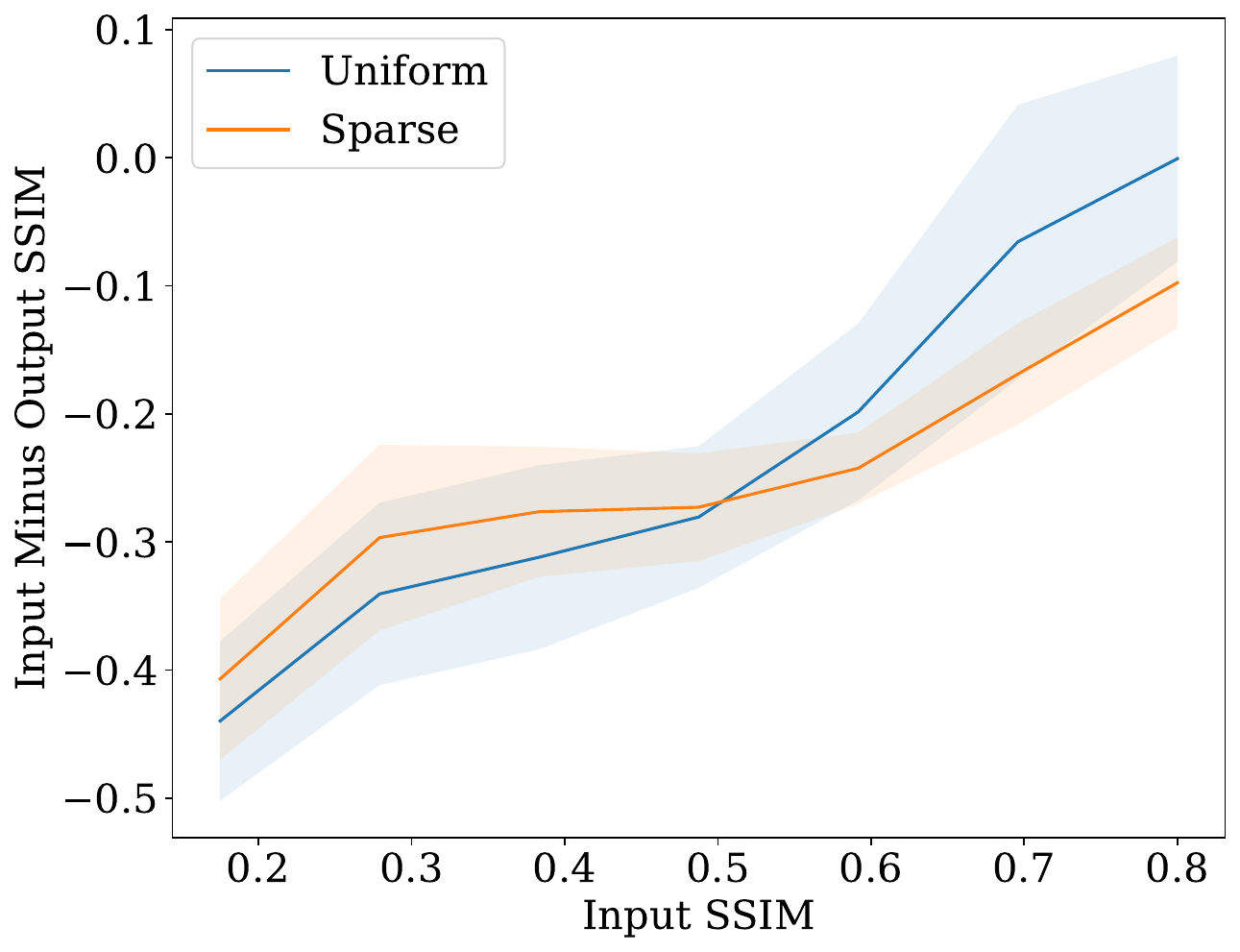}
\caption{ \textbf{Adaptive vs Uniform Training, Experimental Data.} Adaptive training with the polynomial approximation from sparse samples works effectively with real data. Our adaptive training strategy consistently outperforms a uniformly trained baseline at lower noise levels while not being significantly worse at higher noise levels (lower input SSIM). Lower is better.}
\vspace{-10pt}
\label{fig:SparsevsUniversalRealData}
\end{figure}

\paragraph{Quantitative Results}
To compare our adaptive sparse training method to the uniform baseline on the real experimental data, we compare the SSIM of the noisy input to the networks with the SSIM of the output, both with respect to the computed ground truth~\cite{journals/tip/WangBSS04}. 
In this comparison, we use the networks that were trained on the images corrupted with Speckle-Poisson-Gaussian noise. 
We choose to focus SSIM in this case because it has better perceptual qualities than PSNR, but also include a similar plot using PSNR instead in the supplemental material.
We plot the input SSIM minus the output SSIM vs the input SSIM in Figure~\ref{fig:SparsevsUniversalRealData} (lower is better). 
Even though our adaptive distribution trained networks perform worse than the uniform trained networks in the higher noise regime, the adaptive distribution networks perform better in the low noise regime.
Note that for all samples the adaptive trained network improves the image quality, whereas for some lower noise images the uniform trained network actually lowers the image quality.
This is because the uniform trained network is oversmoothing the images due to the higher noise images having a large impact on its training loss. 

\begin{figure*}
\centering




\includegraphics[height=4.0cm]{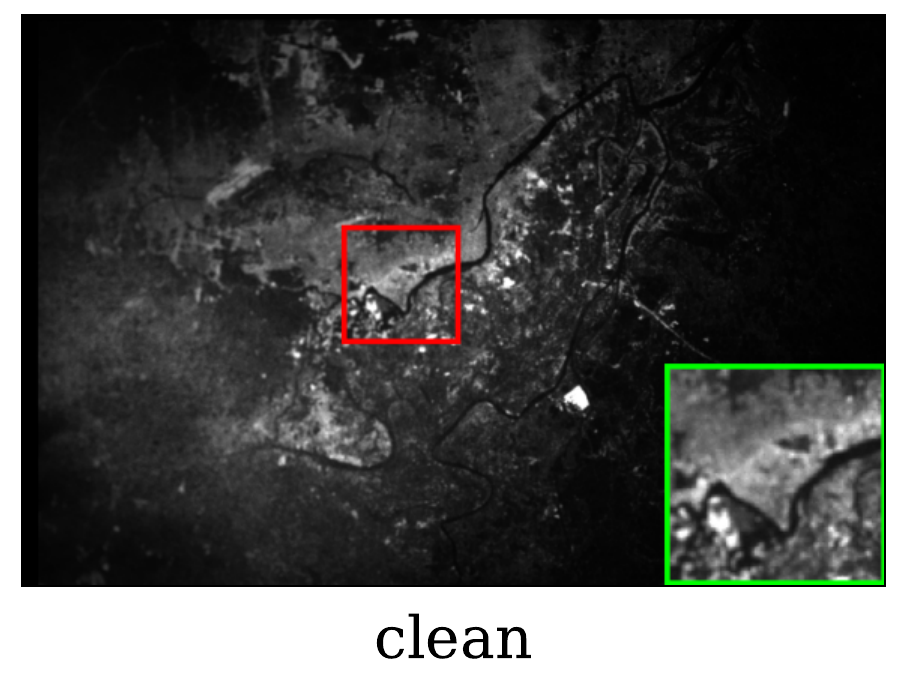}

\includegraphics[height=4.0cm]{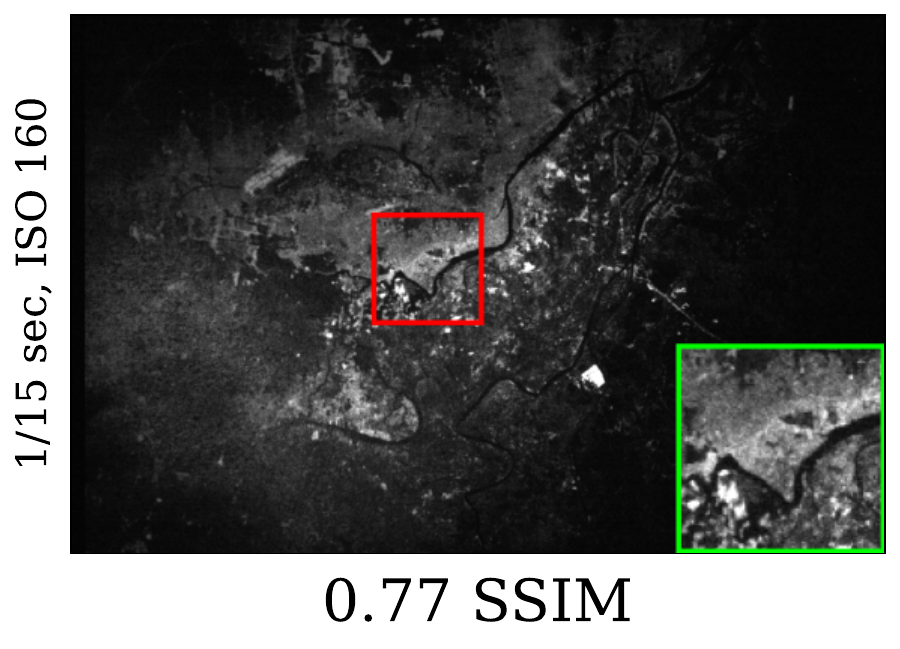}
\includegraphics[height=4.0cm]{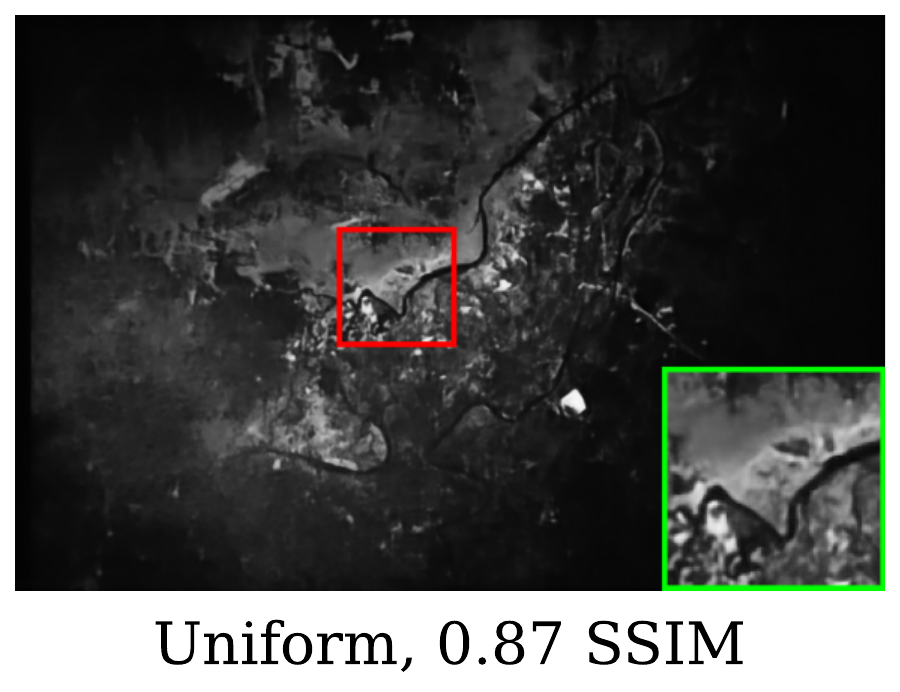} 
\includegraphics[height=4.0cm]{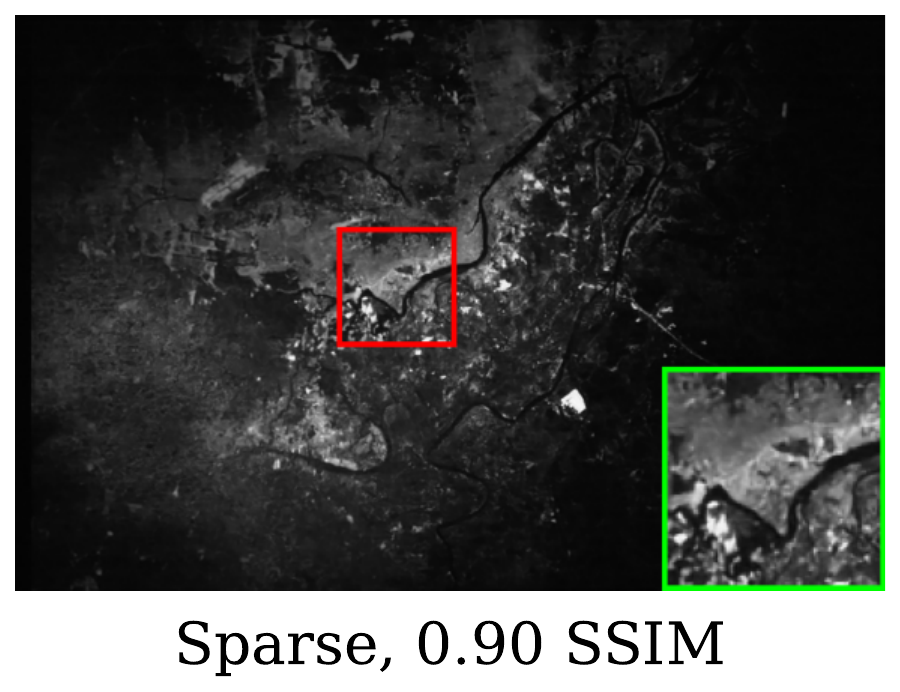}

\includegraphics[height=4.0cm]{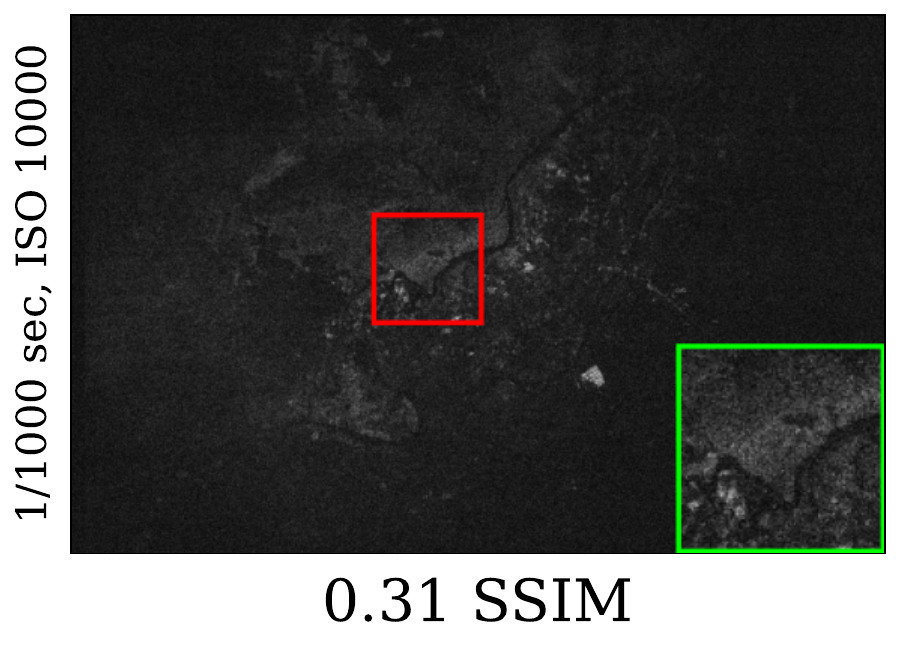}
\includegraphics[height=4.0cm]{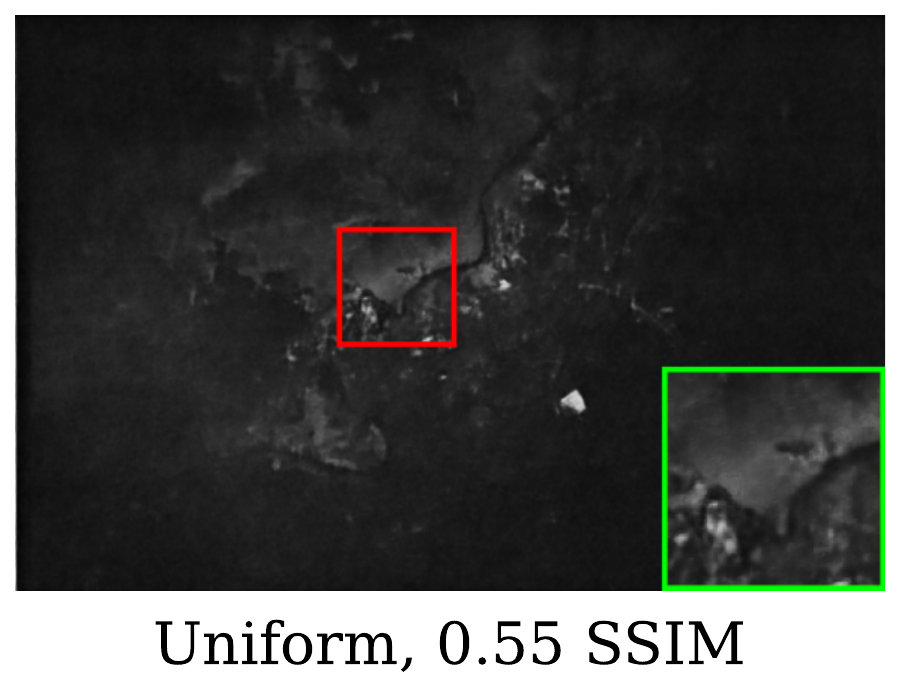} 
\includegraphics[height=4.0cm]{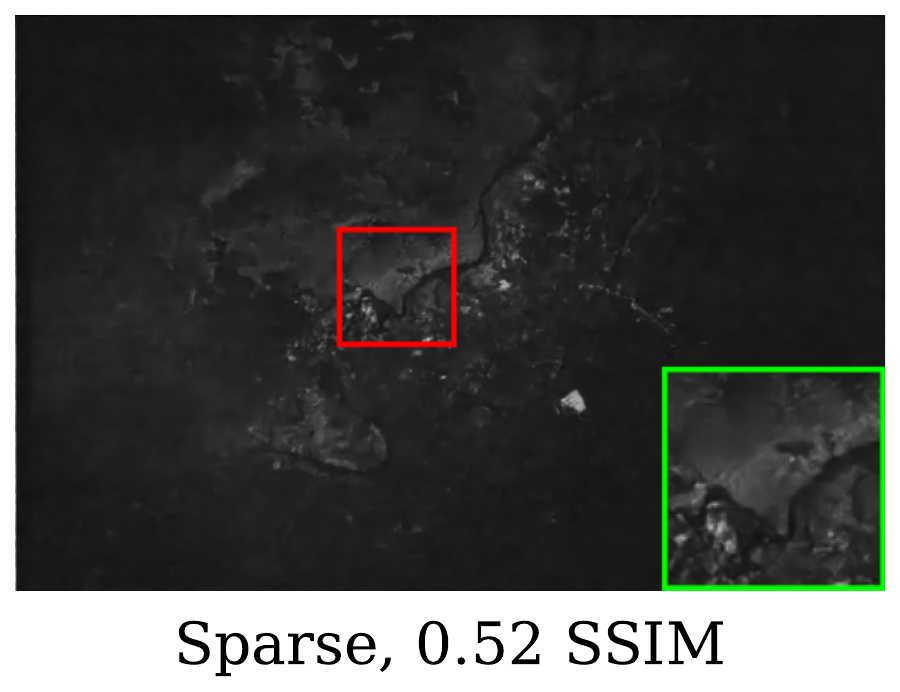}

\caption{ \textbf{Qualitative comparisons, experimental data.} Comparison between the performance of the ideal, uniform-trained, and
adaptive distribution, sparse sampling-trained denoisers on a sample image corrupted with a low amount of noise and
corrupted with a high amount of noise, as determined by the parameters of our experimental setup.
From the closeups it is apparent that the uniform trained denoiser has a tendency to oversmooth its inputs compared to the adaptive distribution trained denoiser. This leads to better performance for the uniform trained at higher noise levels but worse performance at lower noise levels. }
\label{fig:qualitativereal}
\end{figure*}

\paragraph{Qualitative Results} 
A qualitative evaluation on real world data collected in our lab suggests that our method effectively extends passed simulation. Close inspection of the samples in ~\ref{fig:qualitativereal} show that our method is more consistent at preserving information after denoising at low noise levels. Notably, the uniform denoiser smooths the image, losing high frequency details. Even on highly corrupted samples, the our sparse sampling approach is able to achieve performance close to that of the uniformly trained denoiser. Again, our method retains details that the uniform denoiser smooths out of the image. The drop in performance for our method at high noise levels compared to the uniform sampling strategy is further evidence that the uniform strategy is liable to over-correct for high noise samples.

Additional qualitative results on the experimental data can be found in the supplement.

\section{
Further Analysis}
We also analyze some possible extensions of our method such as modifying training to address non-uniform distribution of noise specifications, reducing training time by using finetuning, and visualize noise level sampling distributions and PSNR landscapes of networks during training to empirically justify our choice of approximation. 
\subsection{
Non-Uniform Distribution of Noise Specifications}
In practice, some noise specifications/distributions are more likely to occur than others. For instance,~\cite{brooksUnprocessingImagesLearned2019} demonstrates that camera auto-exposure settings often operate such that there is a positive correlation between shot and read noise parameters. 

If one knows the expected distribution of the noise specifications, i.e.,~how likely different noise distributions are to be encountered in practice, our universal denoiser strategy can be easily modified to weight the training loss according to each specification's likelihood. Essentially, in order to encourage preferential sampling (and thus improved performance) for certain noise configurations, we can increase the target PSNR in the ideal loss landscape by a constant amount over the configurations we care most about (or are most likely to encounter). This has the effect of prioritizing sampling noise levels from those configurations. To validate this idea, for the case of Poisson-Gaussian noise, we consider three cases where we have prior knowledge about the distribution of noise configurations: 1) Poisson and Gaussian noise are correlated, 2) we are more likely to encounter lower noise levels, and 3) we are more likely to encounter higher noise levels. Specifically, assuming the parameter for Poisson noise is $\alpha$ and for Gaussian noise it is $\sigma$, for 1) we add 3 dB to the 10 $(\alpha, \sigma)$ pairs which are linearly between the extreme points $(0.1, 0.02)$ and $(41, 0.66)$ (preferentially sample correlated noise), for 2) we add 3dB to the 25 pairs where $0.1 \leq \alpha \leq 20.5$ and $0.06 \leq \sigma \leq 0.33$ (preferentially sample weak noise), and for 3) we add 3dB to the 25 pairs where $20.5 \leq \alpha \leq 41$ and $0.33 \leq \sigma \leq 0.66$ (preferentially sample strong noise). The results are reported in Table~\ref{table:noise_cfgs}. As expected, our method performs better inside the regions where we biased the sampling, but worse outside the regions, compared to uniform sampling.

\subsection{
Finetuning
}
\begin{figure}
    \centering
    \includegraphics[width=0.3\textwidth]{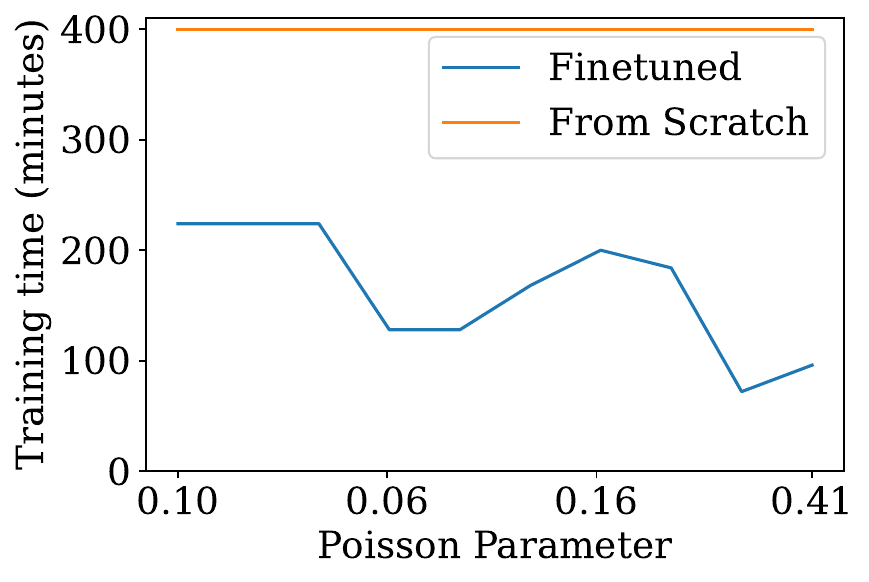}
    \caption{\textbf{Training time of ideal denoisers} Here we compare the time it takes to train a denoiser on one noise configuration from scratch versus finetuned from a general denoiser. In general, we see time savings of 2--4$\times$.}
    \label{fig:ft_time}
\end{figure}

Fine-tuning is a complementary acceleration strategy which allows one to save a constant factor on the time it takes to calculate $\mathcal{L}_{\text{ideal}}$ by finetuning a general denoiser at each of the noise configurations considered instead of training from scratch. We demonstrate this by considering the 1D version of our problem, where we vary the Poisson noise parameter from $0.1$ to $41$. The training time for finetuning was determined by stopping when the validation loss matched that of training from scratch for the full amount of time. As can be seen in Figure~\ref{fig:ft_time}, finetuning saves 2--4$\times$ on time compared to training from scratch---it is equally applicable to our method and our relative acceleration factor stays the same.

\begin{table}
\begin{center}
 \begin{tabular}{|c || c |c ||c |c|} 
 \hline
 Setting & \multicolumn{2}{c||}{Uniform} & \multicolumn{2}{c|}{Adaptive}\\
  & Inside & Outside & Inside & Outside \\  
 \hline
 Correlated & 3.23 & 0.22 & 2.97 & 0.30  \\ 
 Low & 3.13 & 0.26 & 3.04 & 0.50 \\
 High & 3.41 & 0.16 & 3.25 & 0.43 \\
 \hline
 \end{tabular}
 \end{center}
 \caption{\textbf{Tweaking the loss landscape:} We tweak the loss landscapes as in the settings described in the R1QA response and compare our adaptive method with tweaked loss landscapes to uniform training. The numbers reported are the difference in PSNR between the ideal PSNR and the achieved PSNR; lower is better. Our method performers better inside the regions where we biased the sampling, but worse outside the regions, compared to uniform sampling.}
 \label{table:noise_cfgs}
\end{table}

\subsection{
Noise Level Sampling Distribution}
\begin{figure*}
    \centering
    \includegraphics[width=0.95\textwidth]{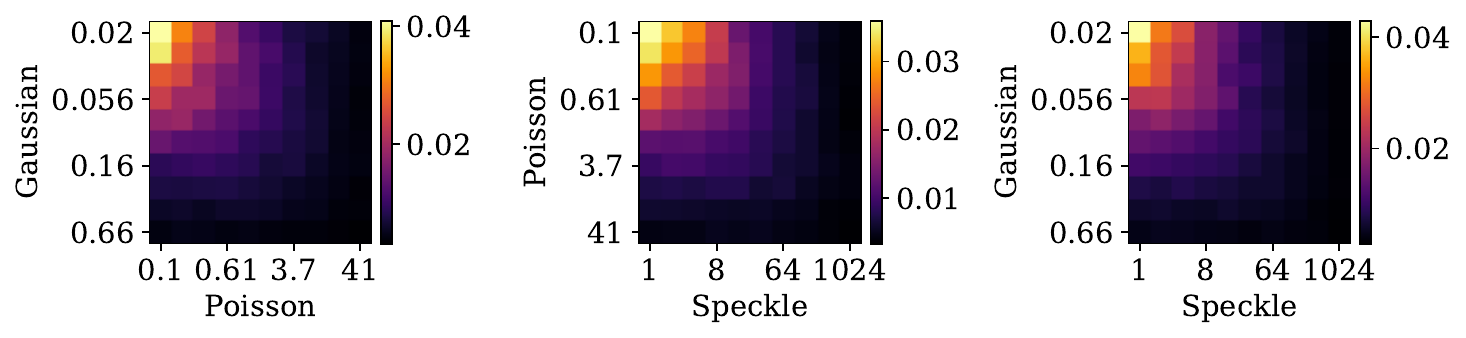}
    \caption{\textbf{Sampling distributions of noise configurations fit by our method} We show the sampling distributions fit by our method for the Poisson-Gaussian, Poisson-Speckle, and Speckle-Gaussian settings. Note that in these sampling distributions, the lower noise configurations have higher probability mass and are preferentially sampled.
    }
    \label{fig:ad}
\end{figure*}

For the three settings, Speckle-Poisson, Speckle-Gaussian, and Poisson-Gaussian noise, we show the sampling distribution of the noise configurations optimized by our method in Figure~\ref{fig:ad}. Note that our training strategy learns to preferentially sample configurations with lower noise, similar to the 1D case in~\cite{gnanasambandam2020one}.

\subsection{
PSNR Landscapes of Networks During Training}
\begin{figure*}[!t]
    \centering
    \includegraphics[width=0.3\textwidth]{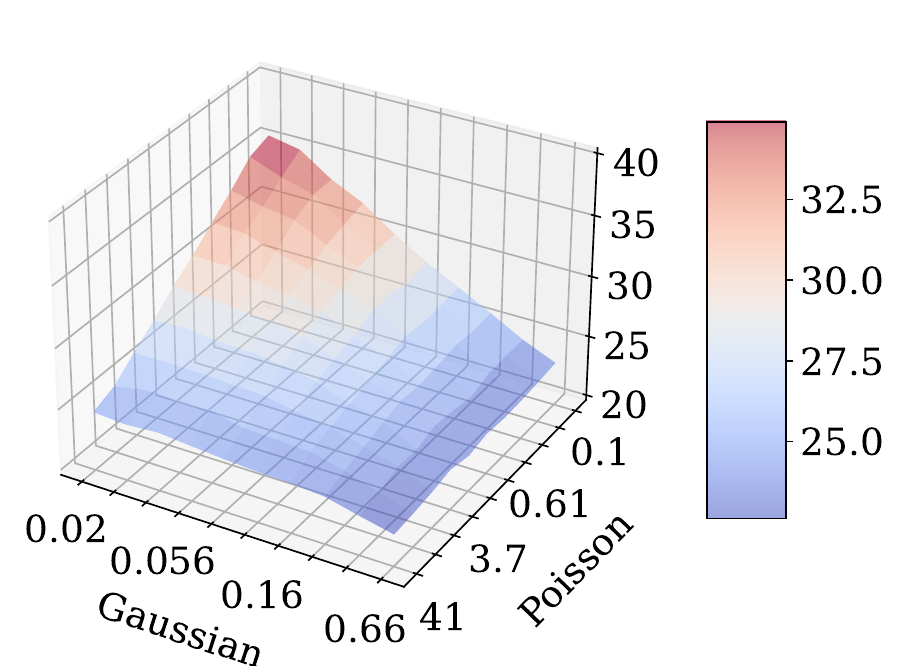}
    \includegraphics[width=0.3\textwidth]{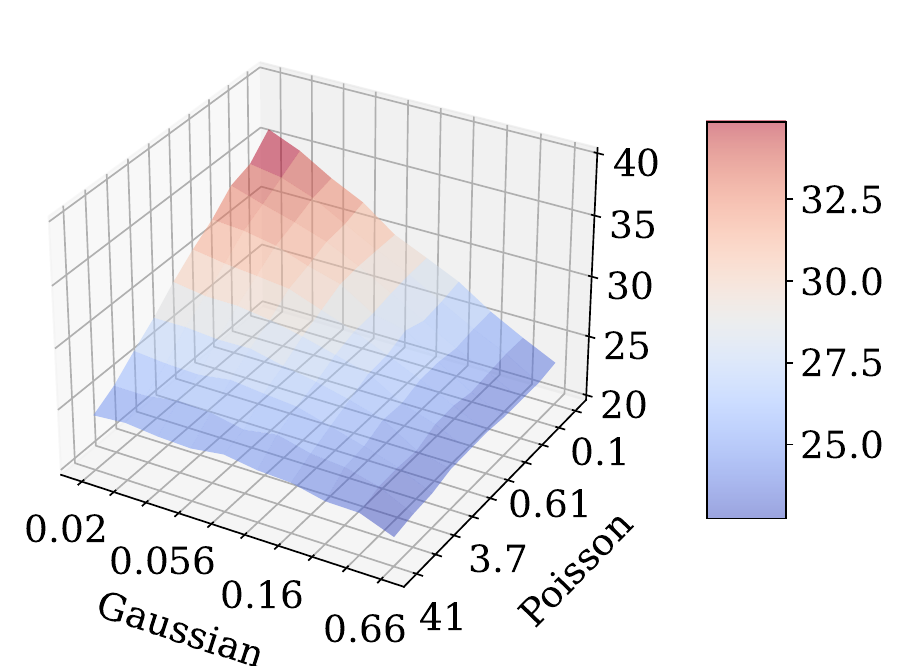}
    \includegraphics[width=0.3\textwidth]{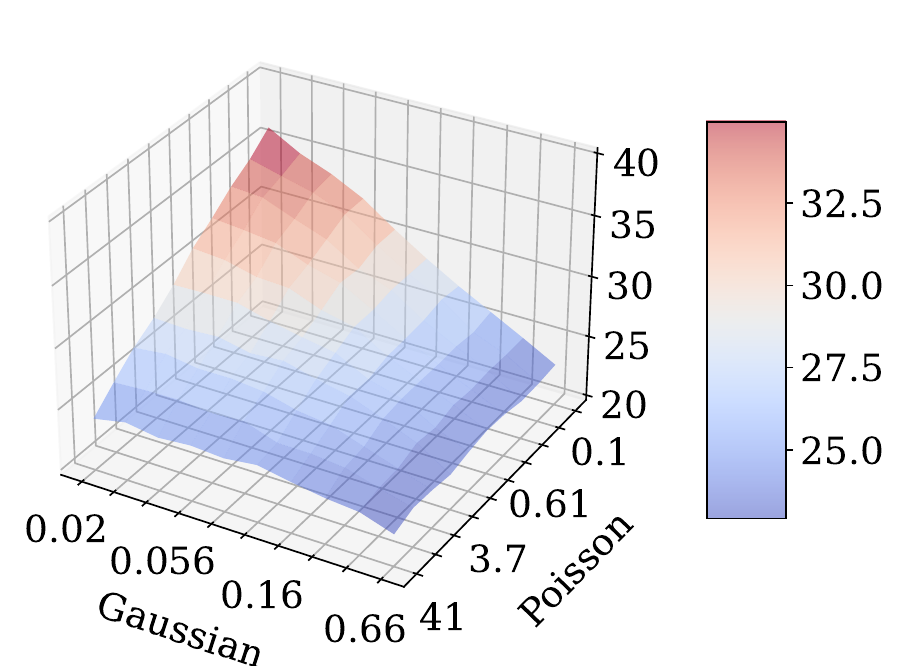}
    \caption{\textbf{PSNR landscape of $\mathcal{L}_{f^{t+1}}$} We show the PSNR landscape of $\mathcal{L}_{f^{t+1}}$ at iterations 10, 30, and 50, three iterations where we update the noise level sampling distribution. They are all smooth.}
    \label{fig:ll_evo}
\end{figure*}

The loss landscapes of $\mathcal{L}_{f^{t+1}}$ are shown in Figure~\ref{fig:ll_evo} from which it is apparent that the ideal PSNR landscape of $\mathcal{L}_{f^{t+1}}$ is smooth can be well-approximated by a polynomial.

\section{Conclusions}
In this work, we demonstrate that we can leverage a polynomial approximation of the specification-loss landscape to train a denoiser to achieve performance which is uniformly bounded away from the ideal across a variety of problem specifications. 
Our results extend to high-dimensional problem specifications (e.g.,~Poisson-Gaussian-Speckle noise) and in this regime our approach offers $50\times$ reductions in training costs compared to alternative adaptive sampling strategies.
We experimentally demonstrate our method extends to real-world noise as well.
More broadly, the polynomial approximations of the specification-loss landscape developed in this work may provide a useful tool for efficiently training networks to perform a range of imaging and computer visions tasks.

\bibliographystyle{IEEEtran}
\bibliography{iclr2023_conference}

\begin{IEEEbiography}
[{\includegraphics[width=1in,height=1.25in,clip,keepaspectratio]{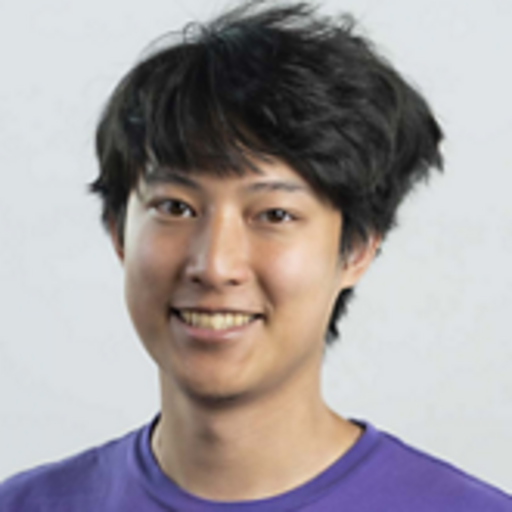}}]{Kevin Zhang} received the B.A. in Computer
Science and Pure Mathematics from the University
of California, Berkeley. He is currently a
Computer Science Ph.D. student at the University
of Maryland, College Park, based primarily
in the Intelligent Sensing Lab, advised by Professor
Christopher Metzler and Jia-bin Huang.
His research focuses on 3D reconstruction from
various sensing modalities. 
\end{IEEEbiography}

\begin{IEEEbiography}
[{\includegraphics[width=1in,height=1.25in,clip,keepaspectratio]{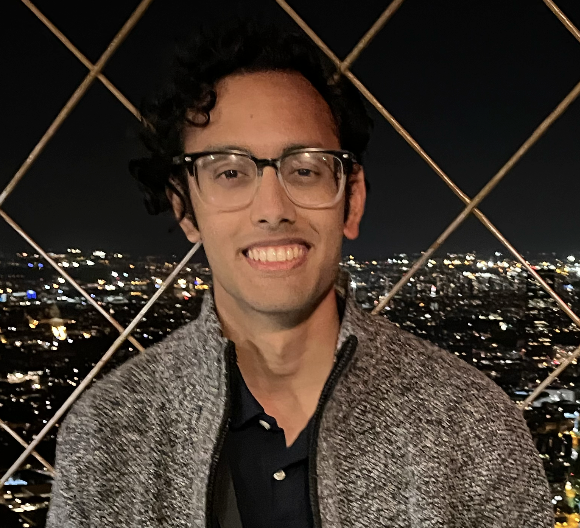}}]{Sakshum Kulshrestha} attended the University of Maryland, where he pursued a B.S. and M.S. in computer science. During his masters, he was based in the Intelligent Sensing lab, where he studied computational imaging and computer vision problems under Dr. Christopher Metzler. Currently, Sakshum is at Waymo, where he works on diffusion models and generative approaches to data synthesis. 
\end{IEEEbiography}

\begin{IEEEbiography}
[{\includegraphics[width=1in,height=1.25in,clip,keepaspectratio]{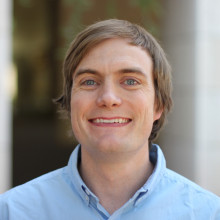}}]{Christopher A. Metzler} is an Assistant Professor
in the Department of Computer Science at
the University of Maryland College Park, where
he leads the UMD Intelligent Sensing Laboratory.
He is a member of the University of Maryland
Institute for Advanced Computer Studies
(UMIACS) and has a courtesy appointment in
the Electrical and Computer Engineering Department.
His research develops new systems
and algorithms for solving problems in computational
imaging and sensing, machine learning,
and wireless communications. His work has received multiple best paper
awards; he recently received NSF CAREER, AFOSR Young Investigator
Program, and ARO Early Career Program awards; and he was an Intelligence
Community Postdoctoral Research Fellow, an NSF Graduate
Research Fellow, a DoD NDSEG Fellow, and a NASA Texas Space
Grant Consortium Fellow.
\end{IEEEbiography}

\newpage



\section{A Scalable Training Strategy for Blind Multi-Distribution Noise Removal - Supplemental Material}


\markboth{Journal of \LaTeX\ Class Files,~Vol.~14, No.~8, August~2021}%
{Shell \MakeLowercase{\textit{et al.}}: A Sample Article Using IEEEtran.cls for IEEE Journals}


\maketitle



\section{Polynomial Approximation} \label{sec:app}

\begin{figure*}
\centering
\includegraphics[height=.28\textwidth]{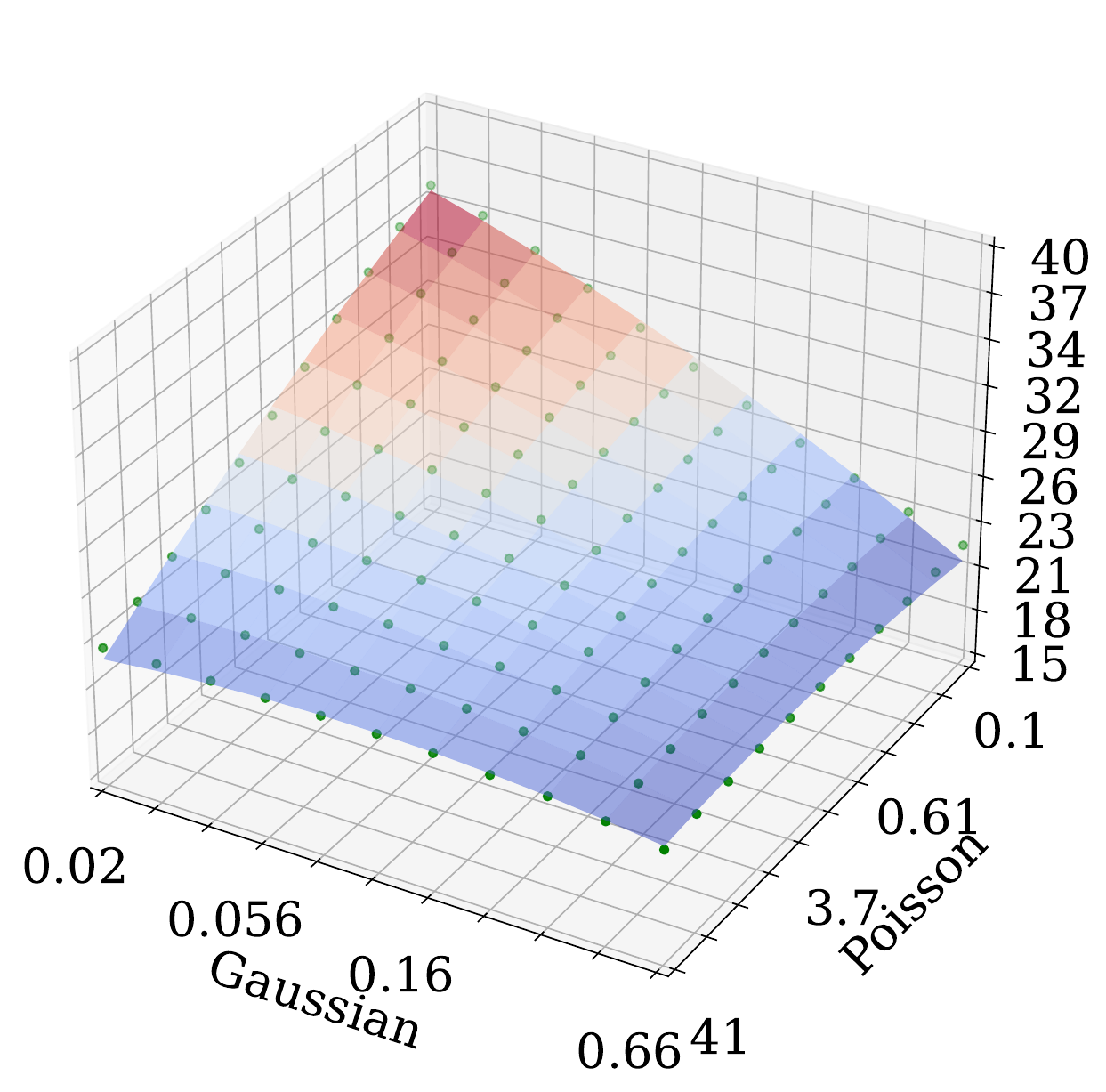}
\includegraphics[height=.28\textwidth]{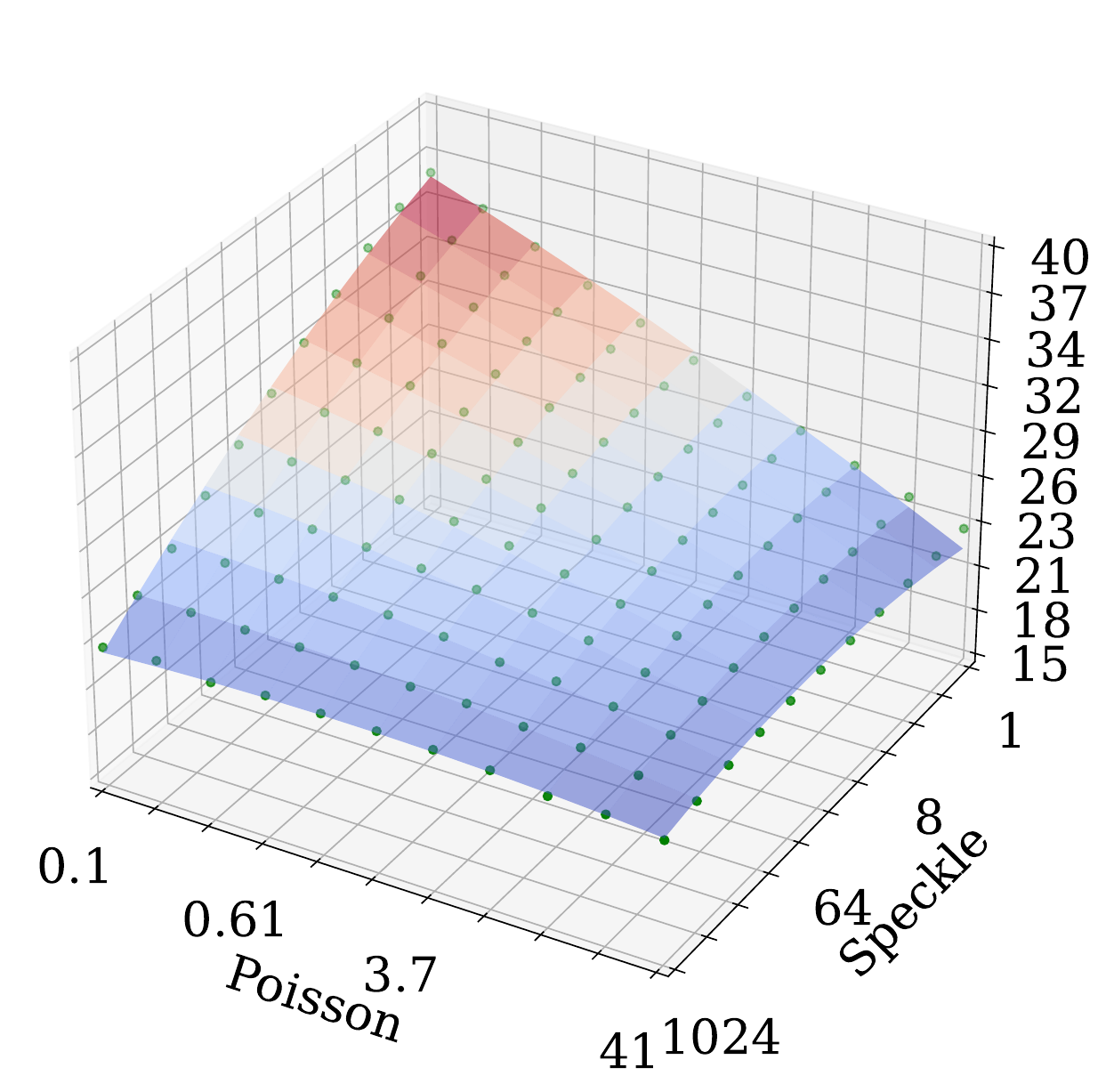}
\includegraphics[height=.28\textwidth]{figures/dense-speckle-gauss-approx.pdf}
\caption{\textbf{Specification-loss Landscape.} PSNR, which we use as our proxy for error, versus denoising task specifications. The surfaces are highly smooth with respect to task specification.}
\label{fig:LossLandscapeApproxSP}
\end{figure*}

We used cross-validation to empirically determine what degree polynomial we should use to fit the specification-loss landscape.
Though densely sampling the specification-loss landscape becomes intractable if its dimension is $3$ or greater, we can still densely sample 2 dimensional specification-loss landscapes then subsample to simulate sparse sampling. 
Through this method, we compared a linear, quadratic, and cubic approximation to the specification-loss landscape and determined that a quadratic polynomial is the most suitable for approximating the specification-loss landscape in this setting. 
More specifically, this means for a given point $s \in S$ we approximate the corresponding loss with the function 
$$P(s) = s^TAs + b^Ts + c,$$
where $A \in R^{n \times n}$ is symmetric, and $b, c \in \mathbb{R}^n$. 
We fit this quadratic using linear least squares with a ridge penalty, where the ridge penalty parameter is also determined using cross-validation. We swept over the values $\{0.1, 0.01, 0.001, 0.0001, 0.00001\}$ and settled on $0.00001$. 

\section{Additional Quantitative Results} \label{sec:extra}
We plot the specification-loss landscapes for all the 2D noise distributions we consider in \ref{fig:LossLandscapeApproxSP} and performance comparisons for adaptive dense and sparse sampling versus uniform sampling in Figures \ref{fig:AdaptiveDenseLossDelta}, \ref{fig:AdaptiveSparseLossDelta}.

\section{Derivation of Dual Ascent iterations} \label{sec:deriv}
First, we restate the main optimization problem, the \textit{uniform gap problem}, that we address in this paper
\begin{equation} \label{eq:prob1}
    f^* = \argmin_{f \in \mathcal{F}} \sup_{\theta \in \Theta} \left \{ \mathcal{L}_f(\theta) - \mathcal{L}_{\text{ideal}}(\theta)\right\}.
\end{equation}
Next, we restate the derivation of the dual ascent algorithm to solve the optimization problem of \eqref{eq:prob1} from \cite{gnanasambandam2020one} in our notation here. 
First, we rewrite the optimization problem from Equation \eqref{eq:prob1} as 
\begin{align*}
    &\min_{f, t}\quad\quad\quad t \\
    &\text{subject to}\quad \mathcal{L}_{f}(\theta) - \mathcal{L}_{\text{ideal}}(\theta) \leq t, \forall \theta \in \Theta.
\end{align*}
Then the Lagrangian is defined as 
\begin{align*}
    L(f, t, \lambda) = t + \int_{\theta \in \Theta} \left\{\mathcal{L}_{f}(\theta) -  \mathcal{L}_{\text{ideal}}(\theta) - t\right\}\lambda(\theta)d\theta
\end{align*}
To get the dual function, we minimize over $f$ and $t$:
\begin{align*}
    g(\lambda) &= \inf_{f, t}L(f, t, \lambda) \\
    &= \begin{cases}
    \inf_f \int (\mathcal{L}_{f}(\theta) -  \mathcal{L}_{\text{ideal}}(\theta))\lambda(\theta)d\theta, \text{ if } \int \lambda(\theta)d\theta = 1 \\
    -\infty, \quad\quad\quad\quad\quad\quad\quad\quad\quad\quad\quad\enspace \text{ otherwise. }
    \end{cases}
\end{align*}
Then the dual problem is defined as 
\begin{align}
    \lambda^* = &\argmax_{\lambda \geq 0} \inf_f \left\{\int(\mathcal{L}_{f}(\theta) -  \mathcal{L}_{\text{ideal}}(\theta))\lambda(\theta)d\theta \right\} \label{eq:dual}\\
    &\text{subject to } \int \lambda(\sigma)d\sigma = 1. \nonumber
\end{align}
Then we can write down the dual ascent iterations as 
\begin{align}
        f^{t+1} &= \argmin_{f\in\mathcal{F}}\left\{\int_{\theta \in \Theta} \mathcal{L}_f(\theta)\lambda^t(\theta) d\theta\right\} \label{eq:iter_0}\\
    \lambda^{t + \frac{1}{2}} &= \lambda^t + \gamma^t\left(\mathcal{L}_{f^{t+1}} - \mathcal{L}_{\text{ideal}}\right) \label{eq:iter_1}\\
    \lambda^{t+1} &= \lambda^{t+\frac{1}{2}} / \int_{\theta \in \Theta} \lambda^{t + \frac{1}{2}}(\theta)d\theta, \label{eq:iter_2}
\end{align}
Here, \eqref{eq:iter_0} solves the inner optimization of \eqref{eq:dual}, fixing $\lambda$, \eqref{eq:iter_1} is a gradient ascent step for $\lambda$, and \eqref{eq:iter_2} ensures that the normalization constraint on $\lambda$ is satisfied. 
Note that because we use PSNR constraints instead of MSE constraints, we use the step 
\begin{align}
    \lambda^{t + \frac{1}{2}} &= \lambda^t + \gamma^t\left(\frac{\mathcal{L}_{f^{t+1}}}{\mathcal{L}_{\text{ideal}}} - 1\right) \label{eq:psnr_step}
\end{align}
in place of \eqref{eq:iter_1}.
Intuitively, the reason is because PSNR is the logarithm of the MSE loss, and a more uniform PSNR gap means that the ratio of the losses is closer to 1, which is where the $\frac{\mathcal{L}_{f^{t+1}}}{\mathcal{L}_{\text{ideal}}} - 1$ term comes from. 

\section{Additional Quantitative Results} \label{sec:additional_results}

\begin{figure*}
\centering
\includegraphics[height=.27\textwidth]{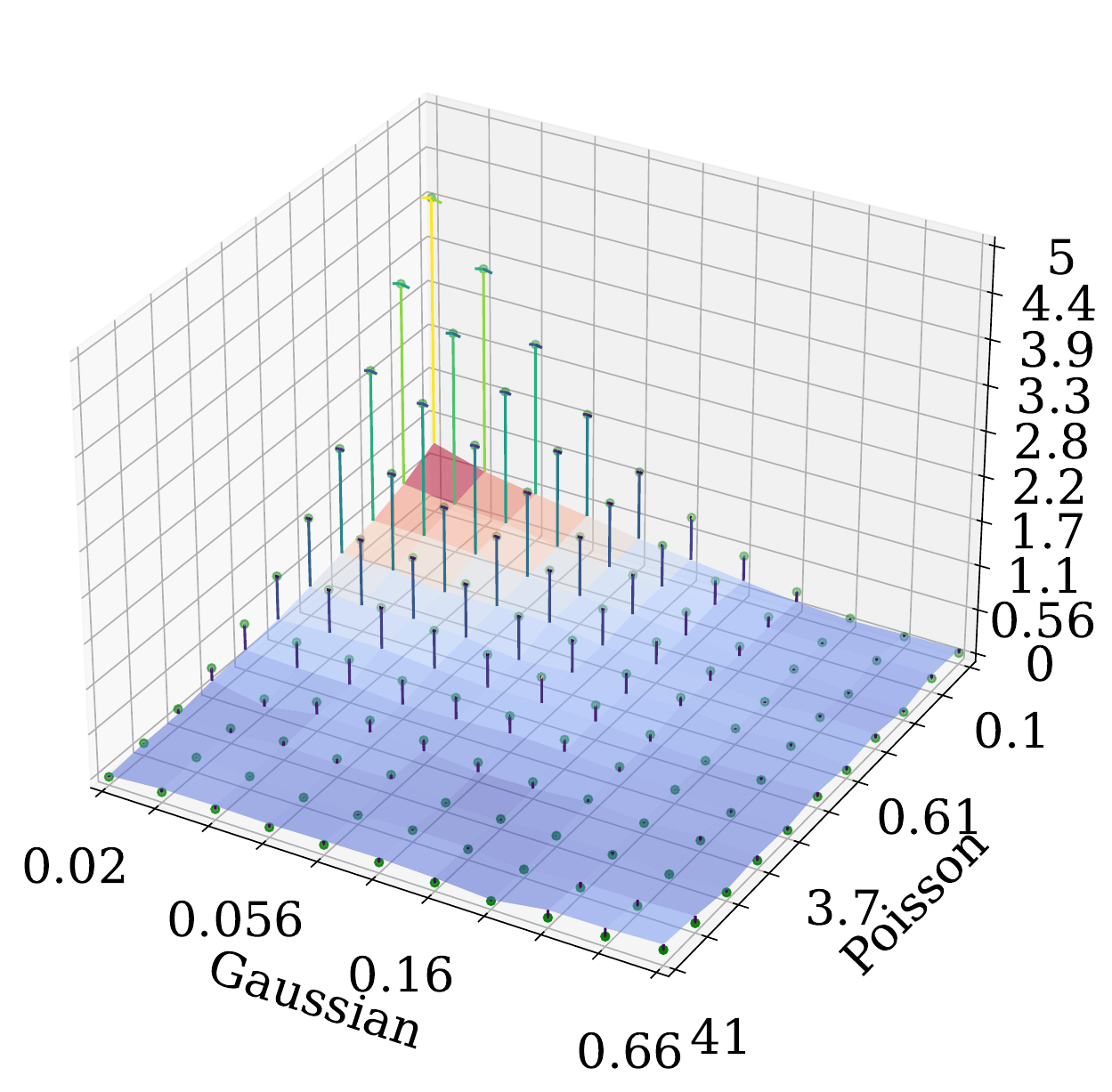}
\includegraphics[height=.27\textwidth]{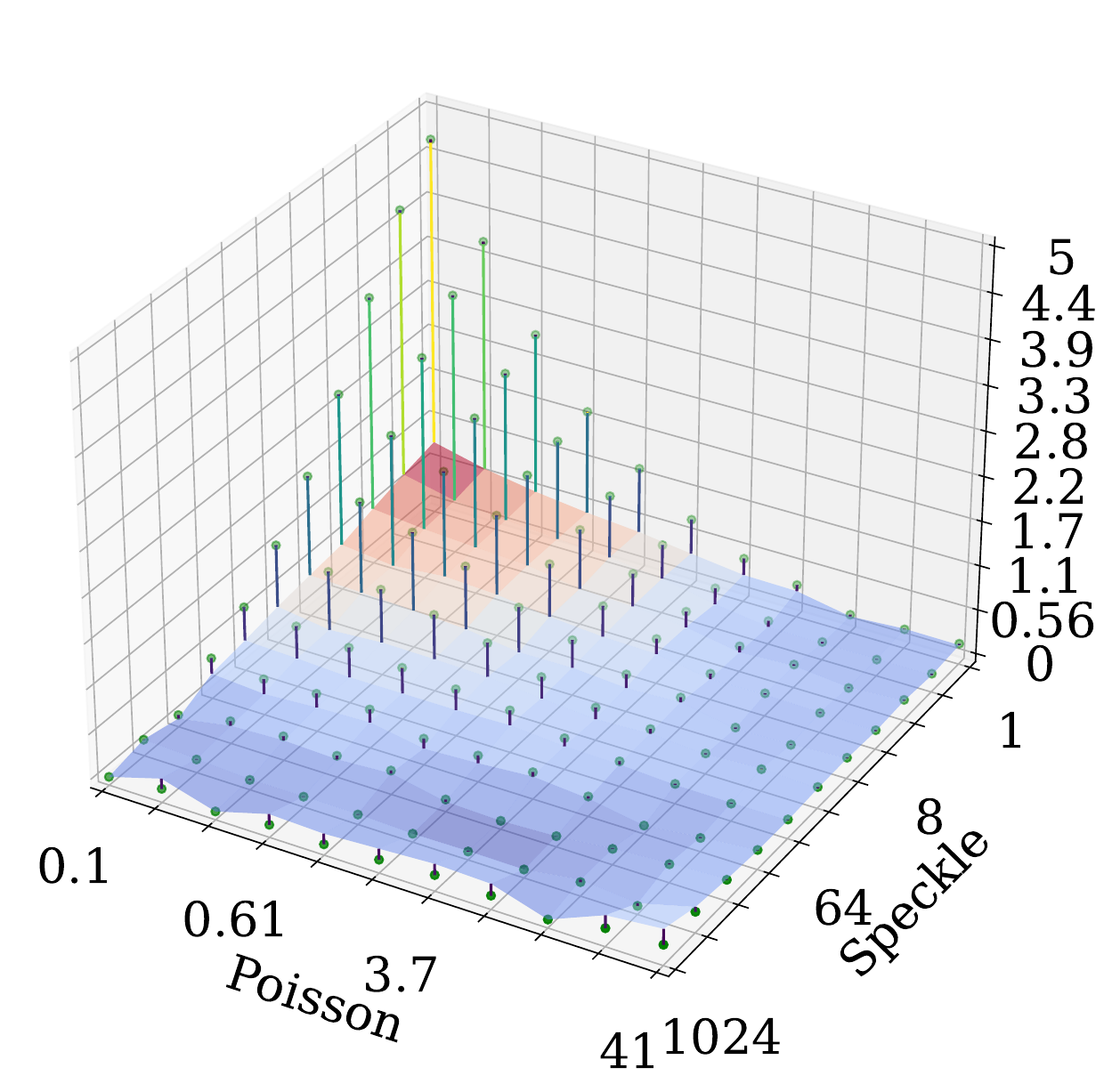}
\includegraphics[height=.27\textwidth]{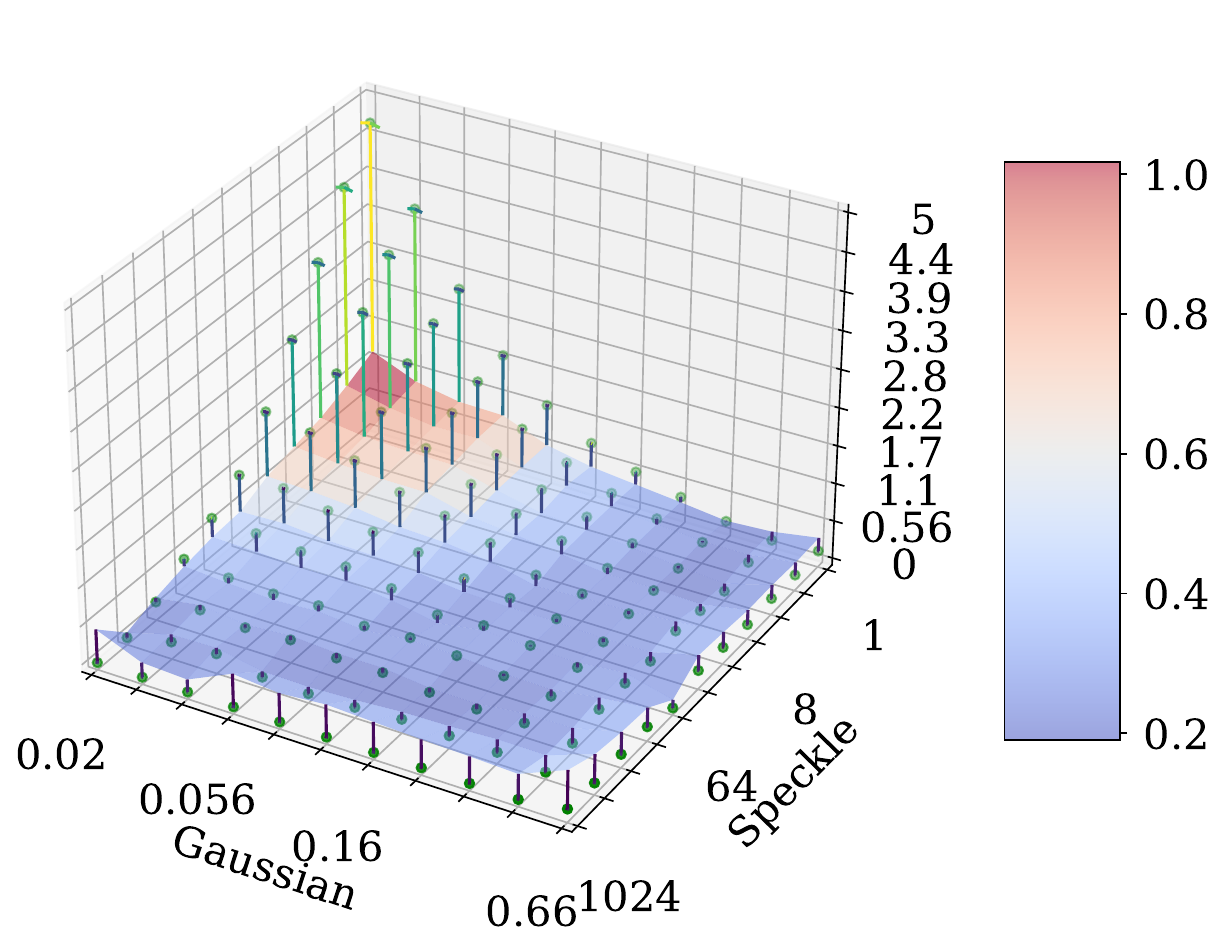}
\caption{ \textbf{Performance comparison between a network trained with adaptive training, using dense sampling of the specification-loss landscape, and a network trained with uniform sampling of the noise levels of the training data.} The surface in the above plots represents the difference in performance between a network trained with the adaptive training strategy with dense sampling of the specification-loss landscape and the ideal networks, and the points represent the differences in performance between a network trained with uniform sampling of the noise levels of the training data and the ideal. Adaptively sampling the noise levels of the training data using a dense sampling of the specification-loss landscape results in a network which uniformly under-performs specialized networks (surface is flat) whereas uniform sampling results in networks that do terribly under certain conditions (points are very high in some regions).}
\label{fig:AdaptiveDenseLossDelta}
\end{figure*}

\begin{figure*}
\centering
\includegraphics[height=.27\textwidth]{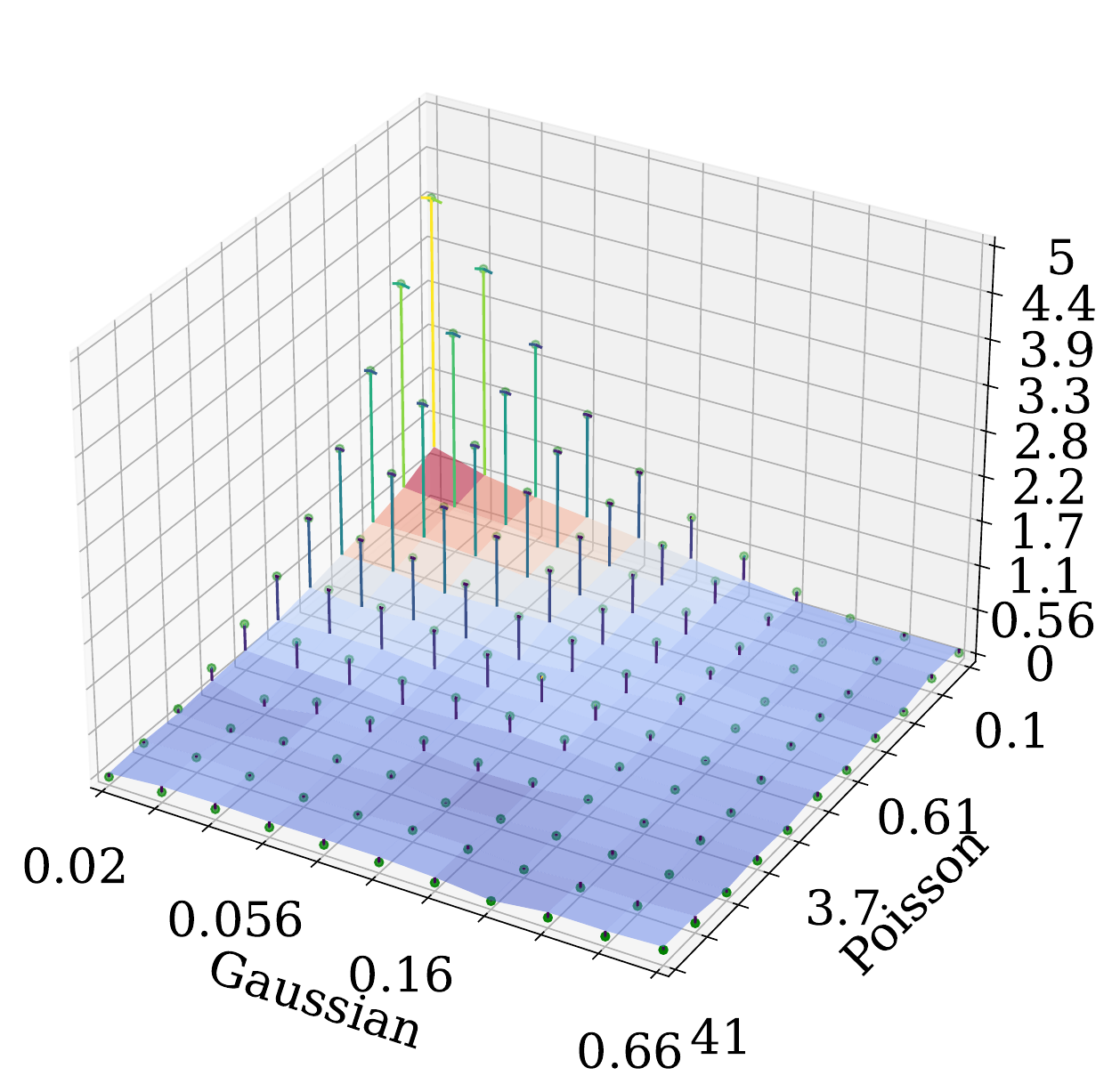}
\includegraphics[height=.27\textwidth]{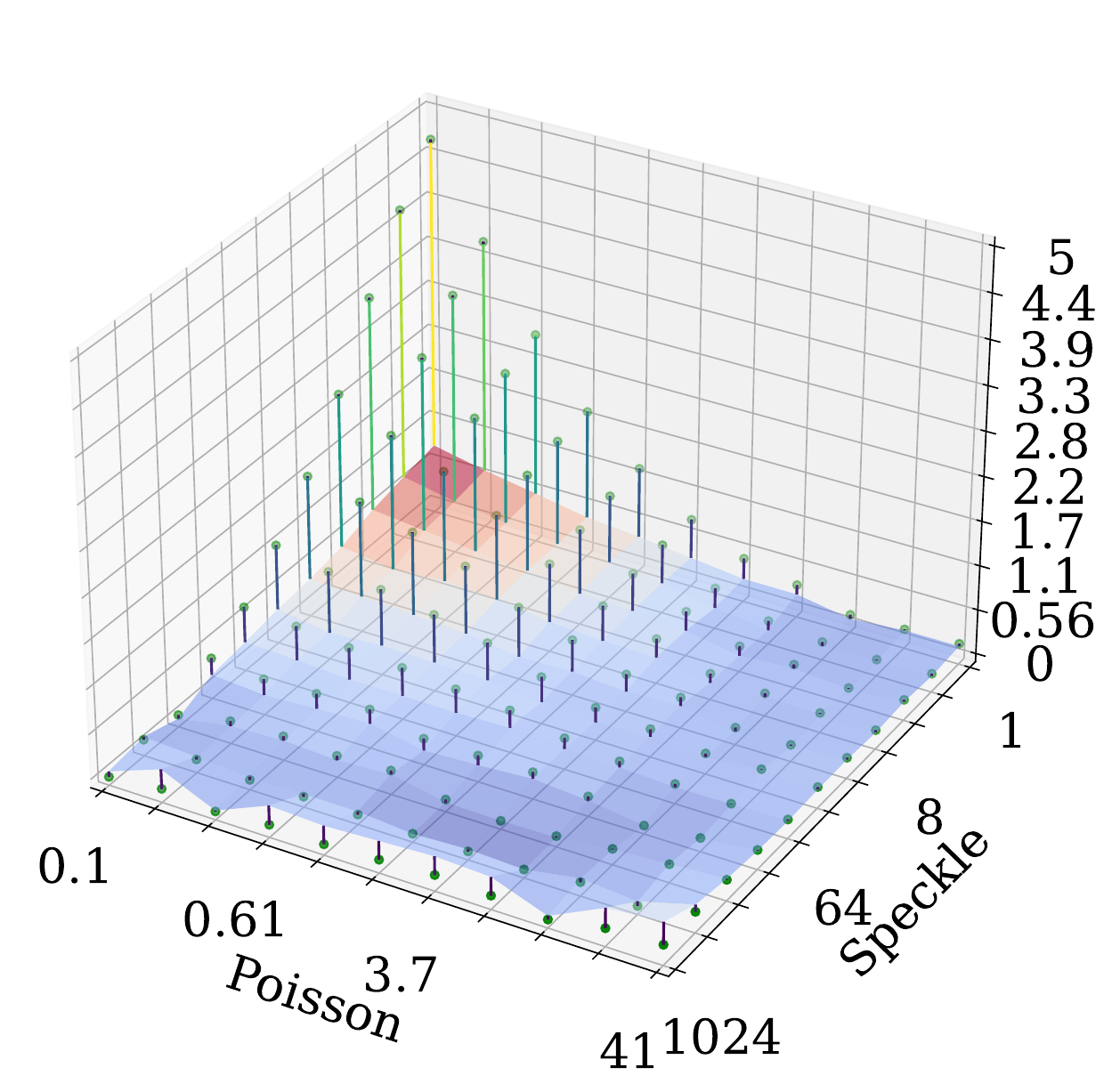}
\includegraphics[height=.3\textwidth]{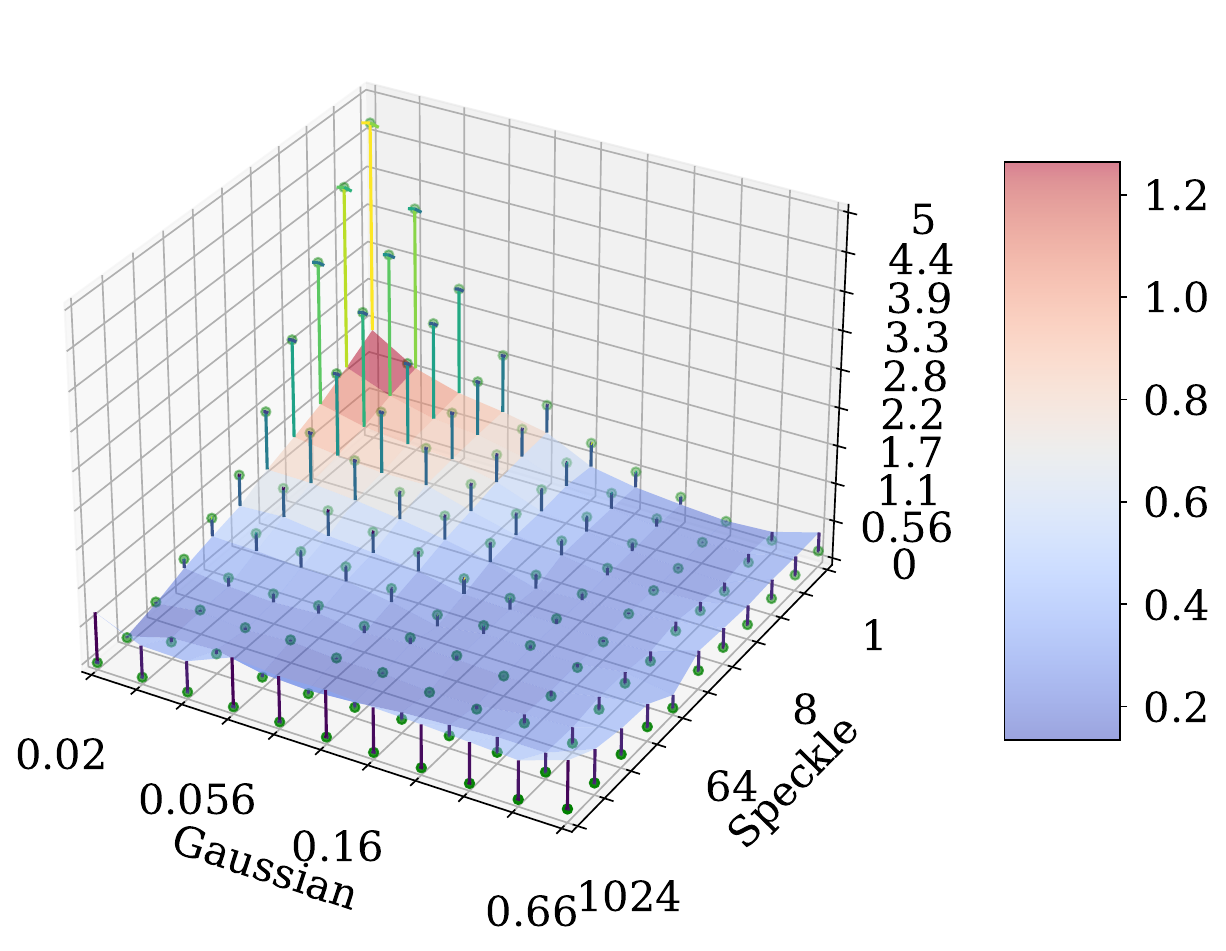}
\caption{\textbf{Performance comparison between a network trained with adaptive training, using
sparse sampling of the specification-loss landscape, and a network trained with uniform sampling
of the noise levels of the training data.} The surface in the above plots represents the difference in performance between the a network trained with the adaptive strategy with sparse sampling and polynomial interpolation of the specification-loss landscape and the ideal networks, and the points represent the differences in performance between a network trained with uniform sampling of the noise levels of the training data and the ideal networks. Like the networks trained with adaptive noise level sampling and dense sampling, we still achieve performance that is uniformly worse than the ideal without severe failure modes. }
\label{fig:AdaptiveSparseLossDelta}

\end{figure*}

\begin{figure}
\centering
\includegraphics[width=.5\textwidth]{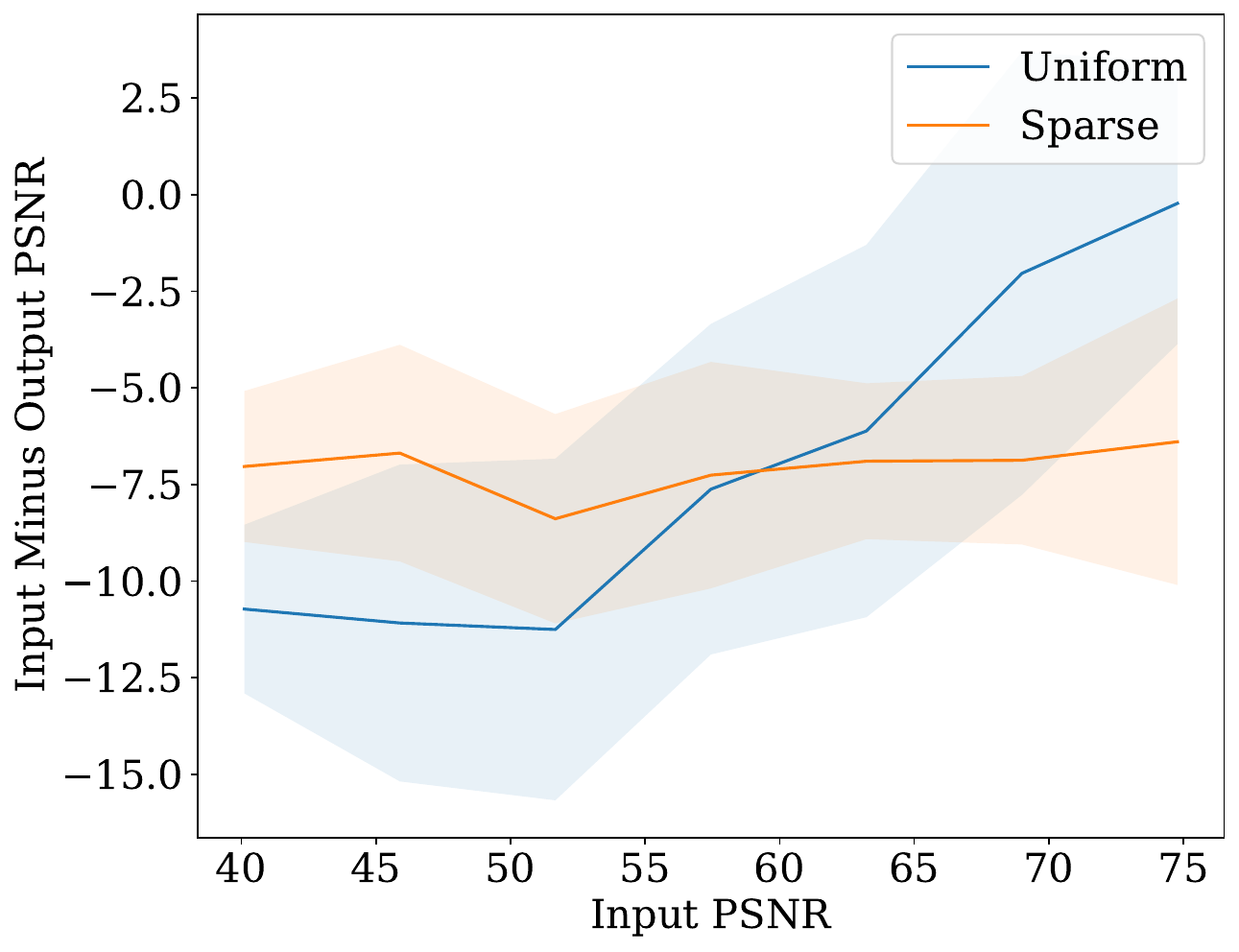}
\caption{ \textbf{Adaptive vs Uniform Training, Experimental Data.} Adaptive training with the polynomial approximation from sparse samples works effectively with real data. Our adaptive training strategy consistently outperforms a uniformly trained baseline at lower noise levels while not being significantly worse at higher noise levels (lower input PSNR).}

\label{fig:SparsevsUniversalRealData_PSNR}
\end{figure}

Figure \ref{fig:AdaptiveDenseLossDelta} shows that Chan et. al's sampling strategy for training denoisers can be applied directly to mixed noise distributions to achieve performance much more consistently close to the ideal when compared a uniform specification sampling strategy.

However, Figure \ref{fig:AdaptiveSparseLossDelta} shows that instead of constructing the entire loss landscape, we can sparsely sample the specification-loss landscape and interpolate an approximation to the true specification-loss landscape to adaptively training a denoiser. The resulting performance is more closely uniformly bounded from the ideal performance compared to the denoiser trained with uniform specification sampling.

We plot the PSNR of the output of our trained networks versus the input noisy image PSNR in Figure~\ref{fig:SparsevsUniversalRealData_PSNR}.

Note that although the input PSNR seems high, the images are quite noisy. 
In general, PSNR has less utility comparing images across different distributions than images from the same or similar distributions, so we would like to emphasize here looking at the relative PSNR between the network noisy inputs and denoised outputs.

\section{Additional Qualitative Results} \label{sec:additional_qualitative}

\begin{figure*}[t]
\centering

\includegraphics[width=.24\textwidth]{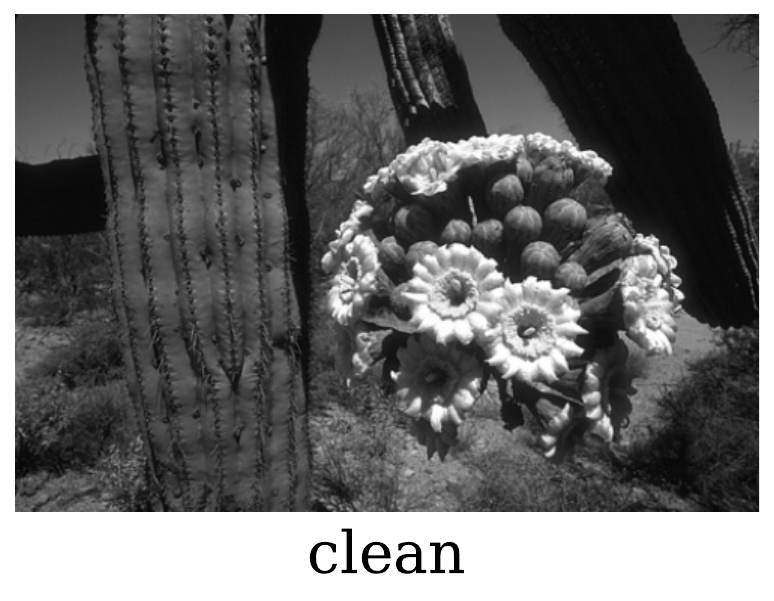}

\includegraphics[width=.26\textwidth]{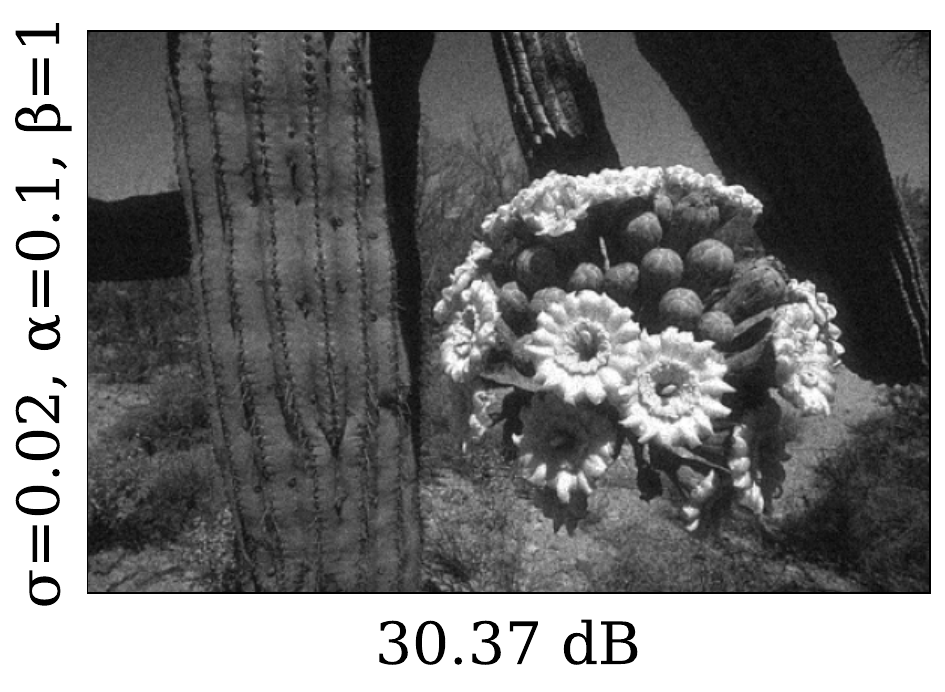}
\includegraphics[width=.24\textwidth]{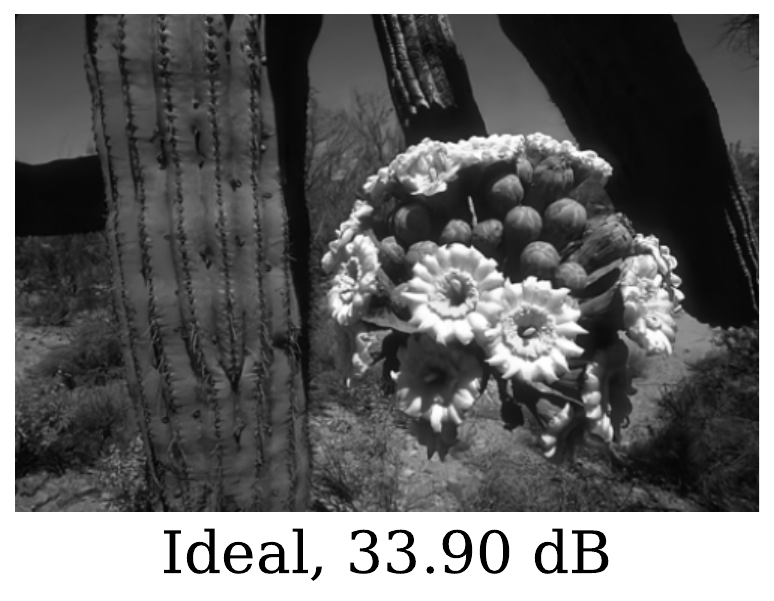} 
\includegraphics[width=.24\textwidth]{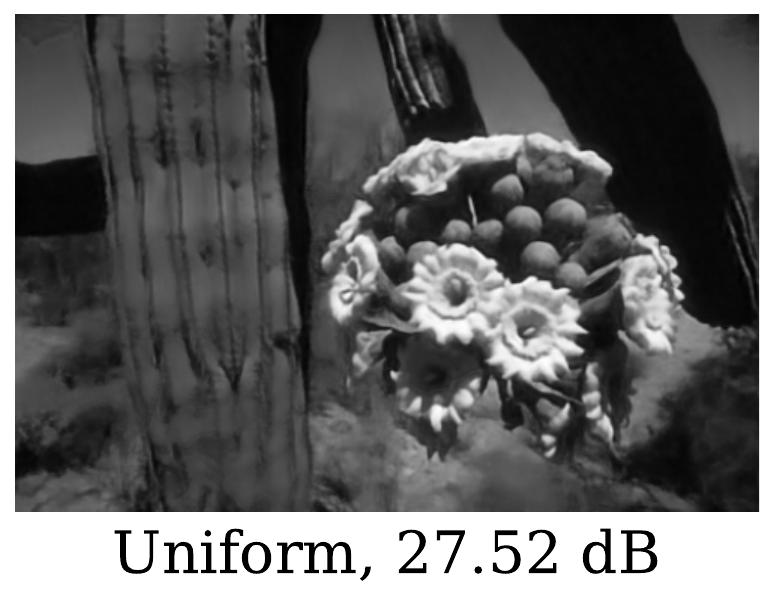}
\includegraphics[width=.24\textwidth]{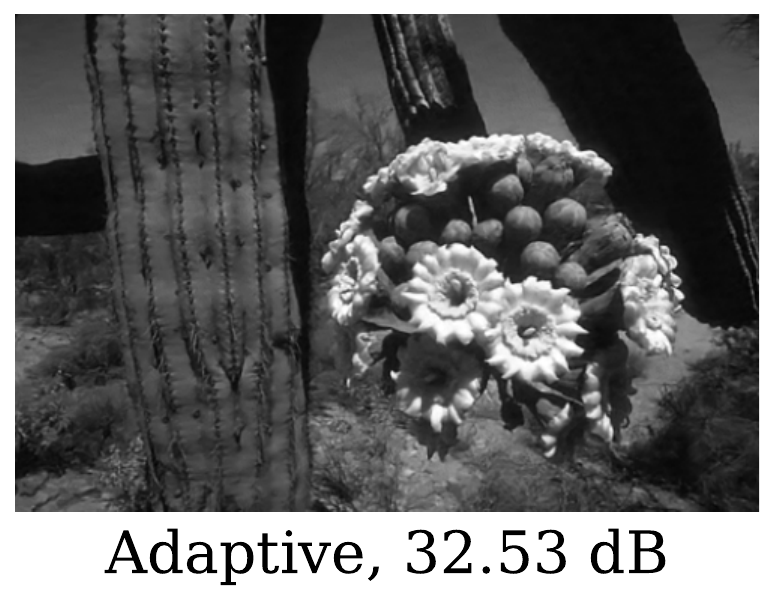}

\includegraphics[width=.26\textwidth]{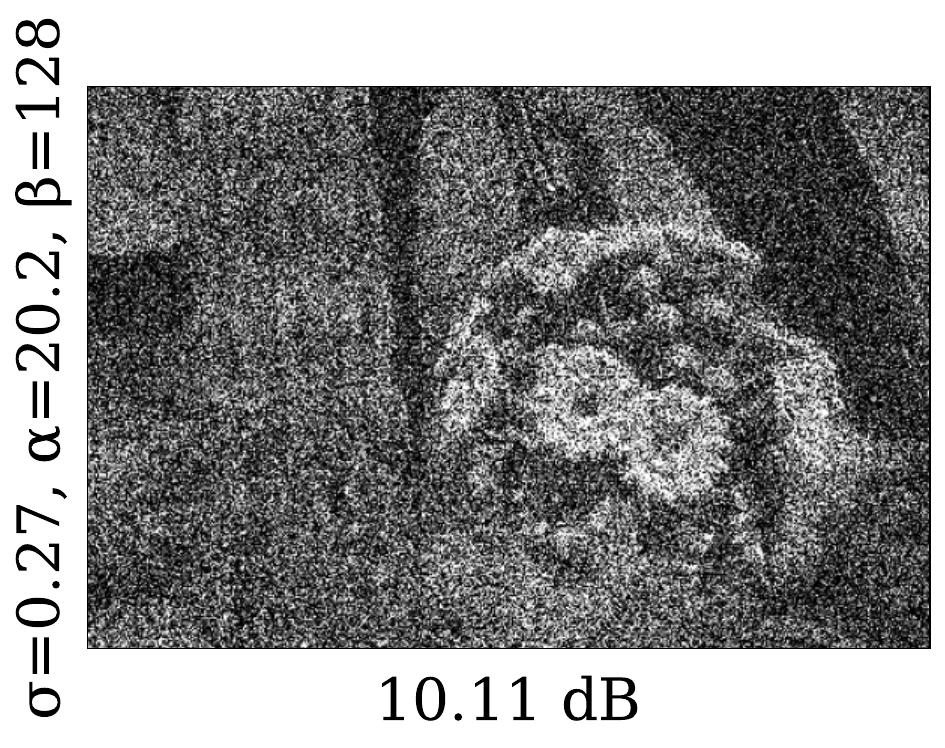}
\includegraphics[width=.24\textwidth]{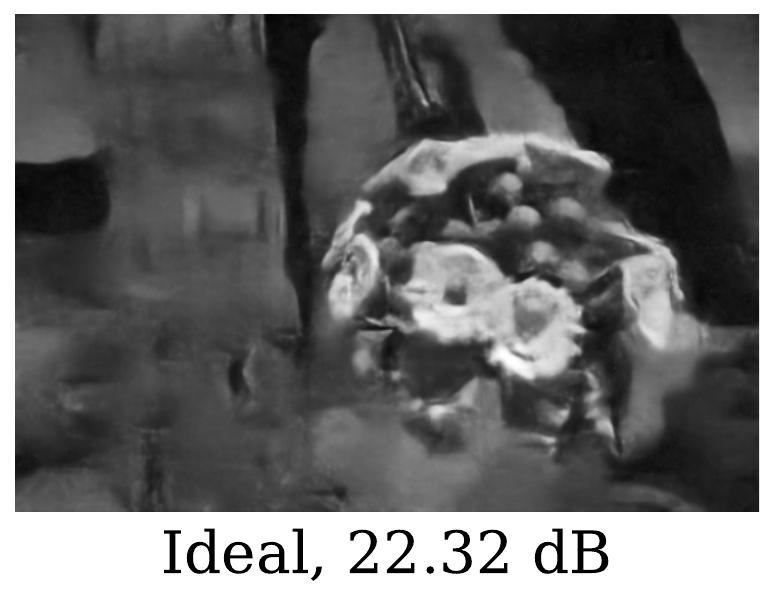} 
\includegraphics[width=.24\textwidth]{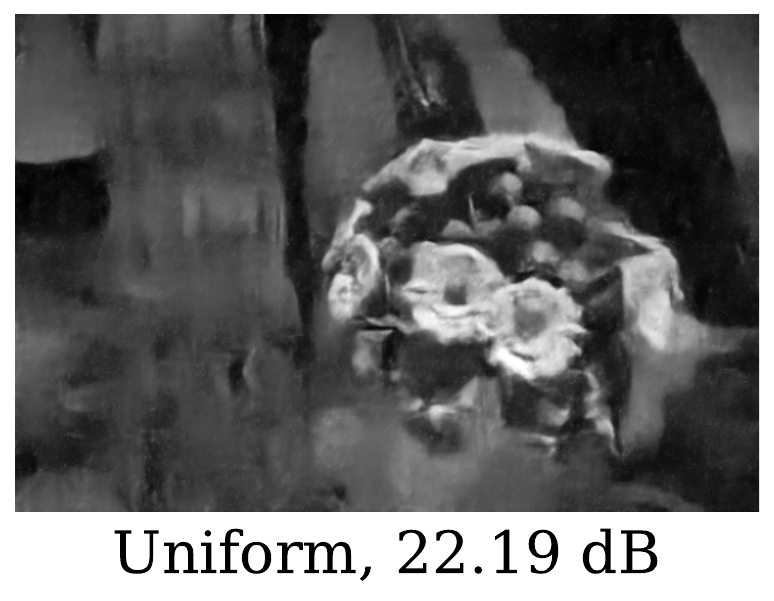}
\includegraphics[width=.24\textwidth]{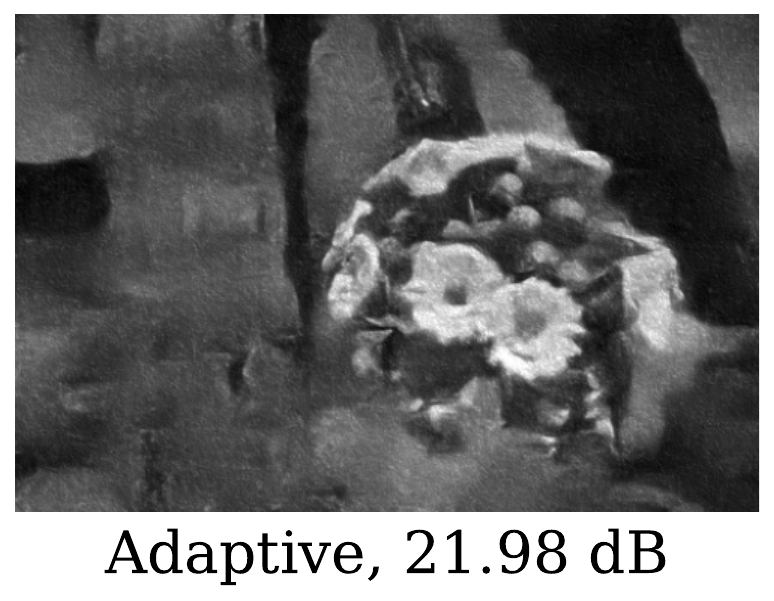}
\caption{ \textbf{Qualitative comparisons.} Comparison between the performance of the ideal, uniform-trained, and adaptive-trained denoisers on a sample image corrupted with a low amount of noise and corrupted with a high amount of noise. Our adaptive blind training strategy performs only marginally worse than an ideal, non-blind baseline when applied to ``easy'' problem specifications, and significantly better than the uniform baseline, while also being only marginally worse than an ideal baseline and uniform baseline under ``hard'' problem specifications.}
\label{fig:qualitative_cactus}
\end{figure*}

\begin{figure*}[t]
\centering

\includegraphics[width=.24\textwidth]{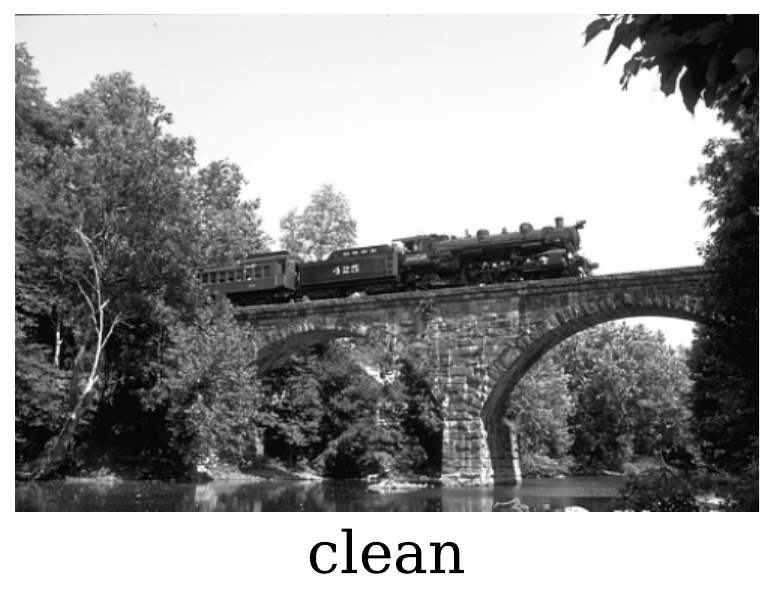}

\includegraphics[width=.26\textwidth]{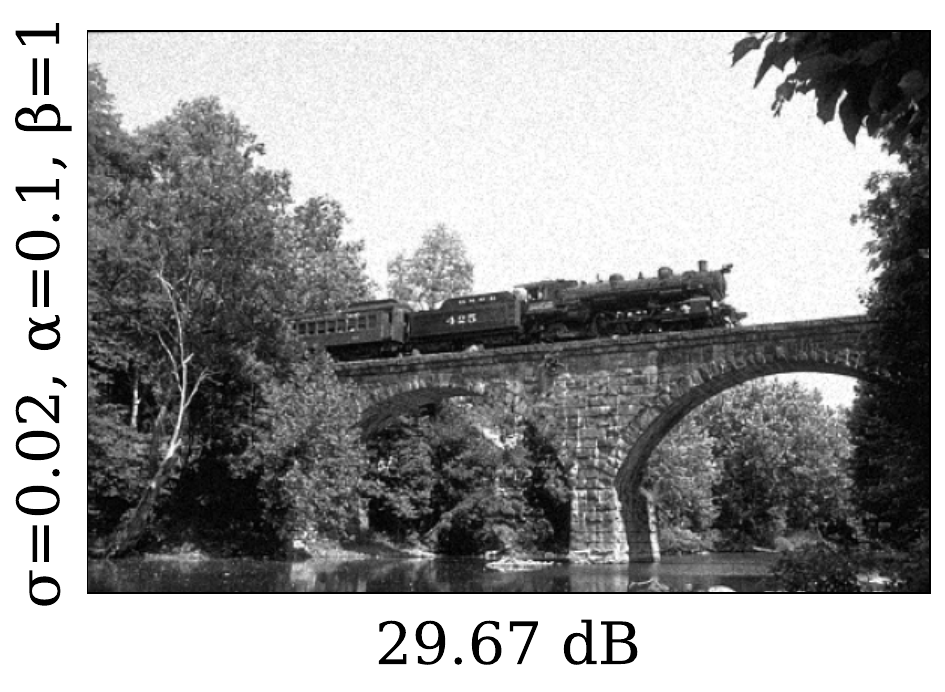}
\includegraphics[width=.24\textwidth]{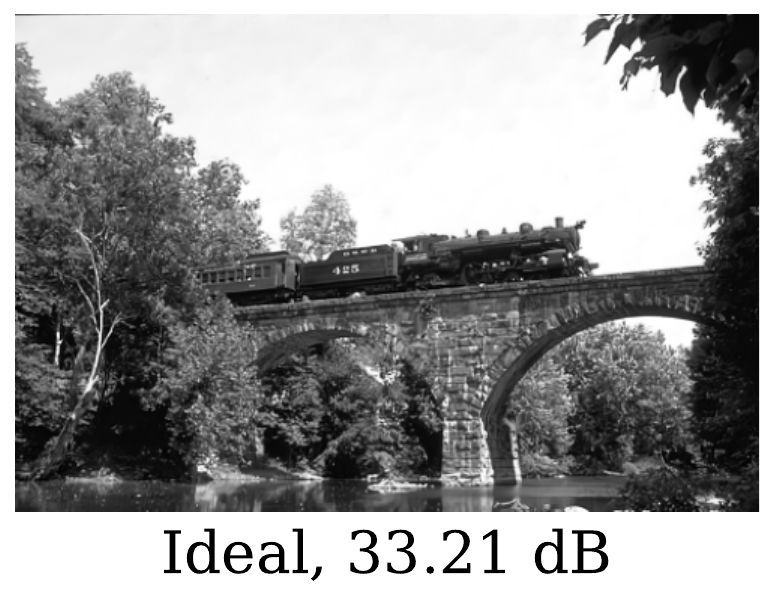} 
\includegraphics[width=.24\textwidth]{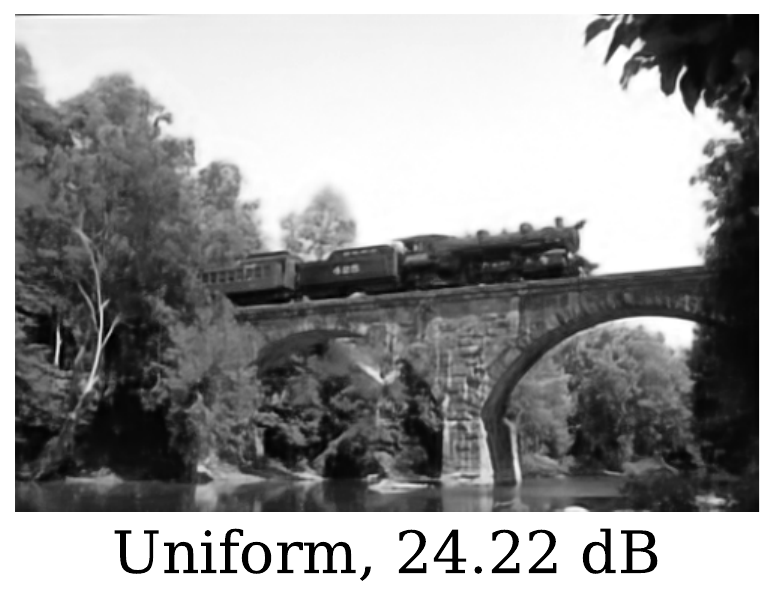}
\includegraphics[width=.24\textwidth]{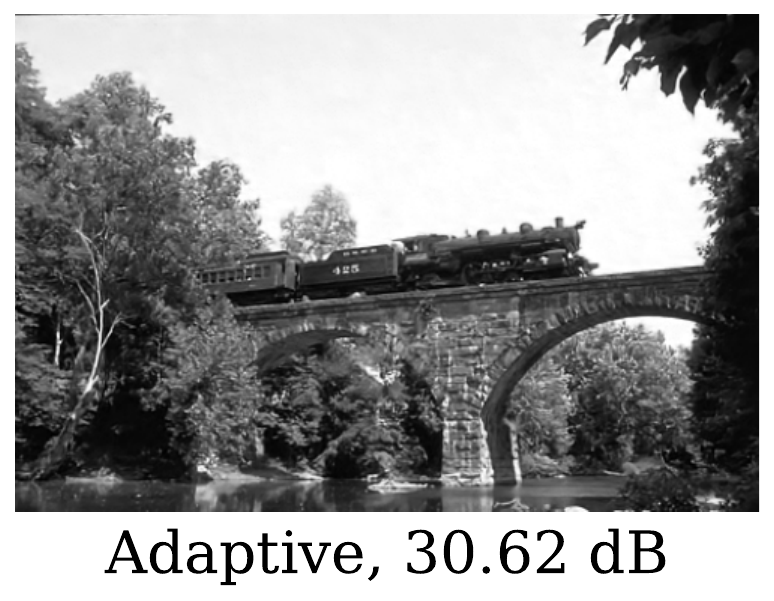}

\includegraphics[width=.26\textwidth]{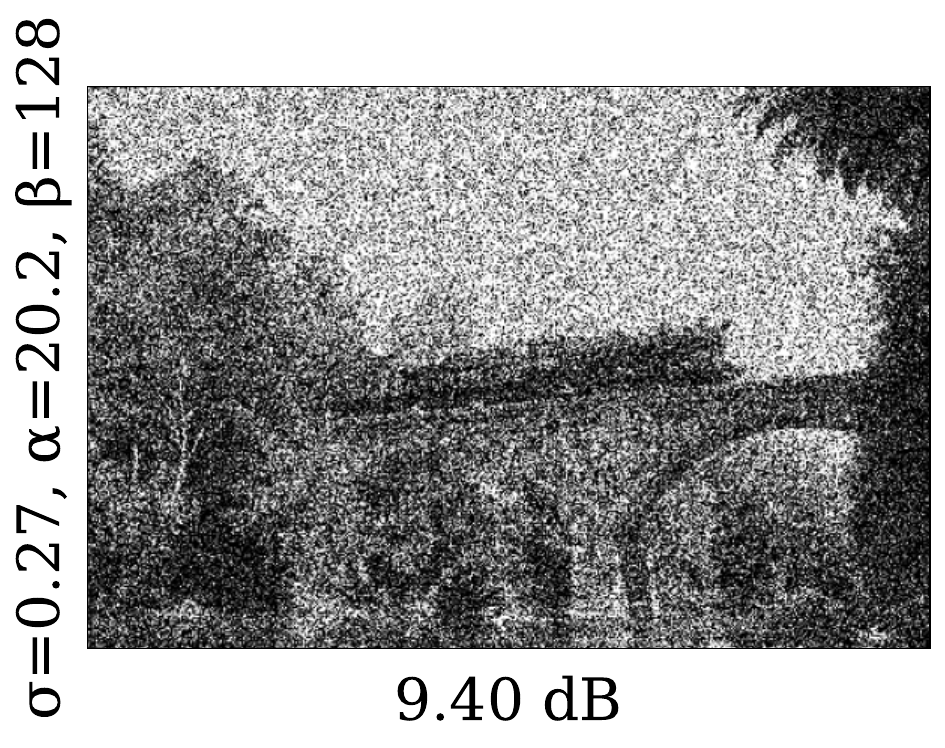}
\includegraphics[width=.24\textwidth]{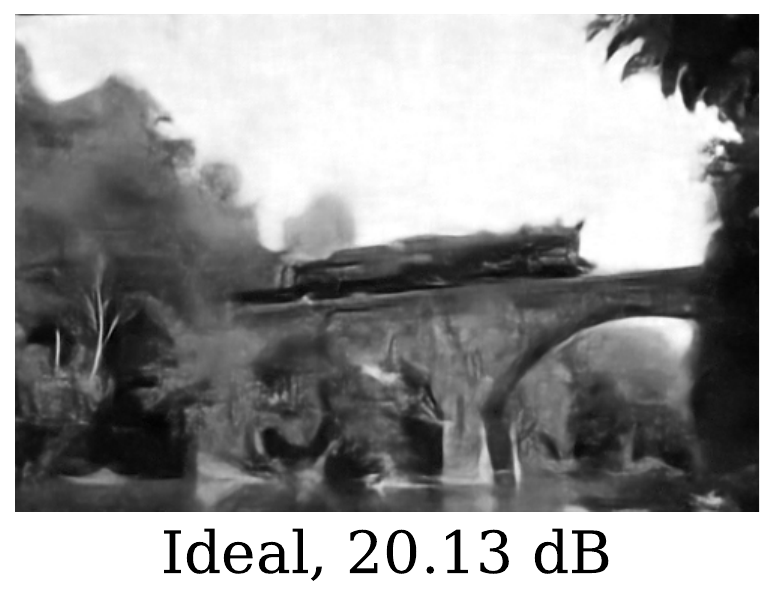} 
\includegraphics[width=.24\textwidth]{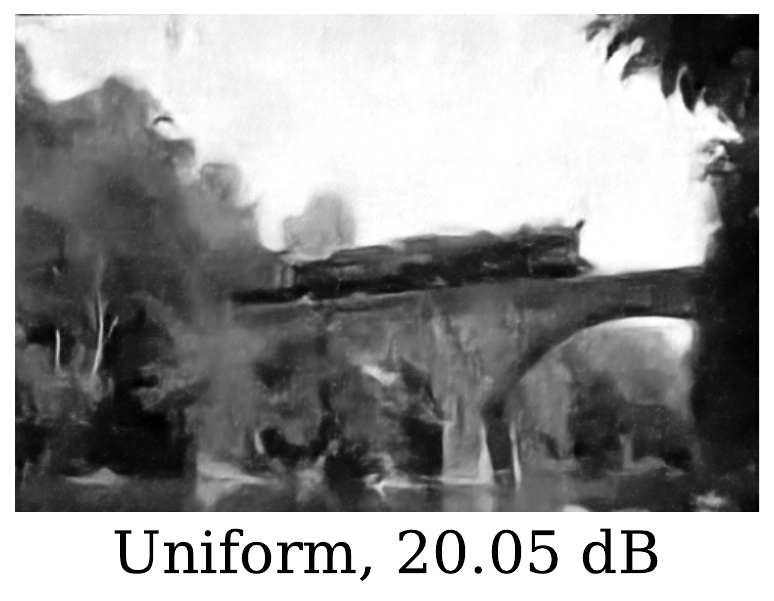}
\includegraphics[width=.24\textwidth]{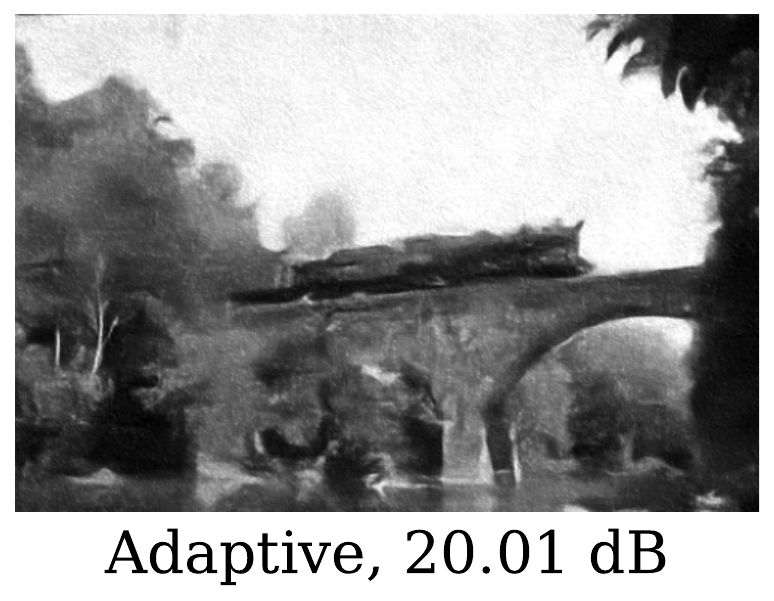}
\caption{ \textbf{Qualitative comparisons.} Comparison between the performance of the ideal, uniform-trained, and adaptive-trained denoisers on a sample image corrupted with a low amount of noise and corrupted with a high amount of noise. Our adaptive blind training strategy performs only marginally worse than an ideal, non-blind baseline when applied to ``easy'' problem specifications, and significantly better than the uniform baseline, while also being only marginally worse than an ideal baseline and uniform baseline under ``hard'' problem specifications.}
\label{fig:qualitative_train}
\end{figure*}

\begin{figure*}
\centering

\includegraphics[width=.31\textwidth]{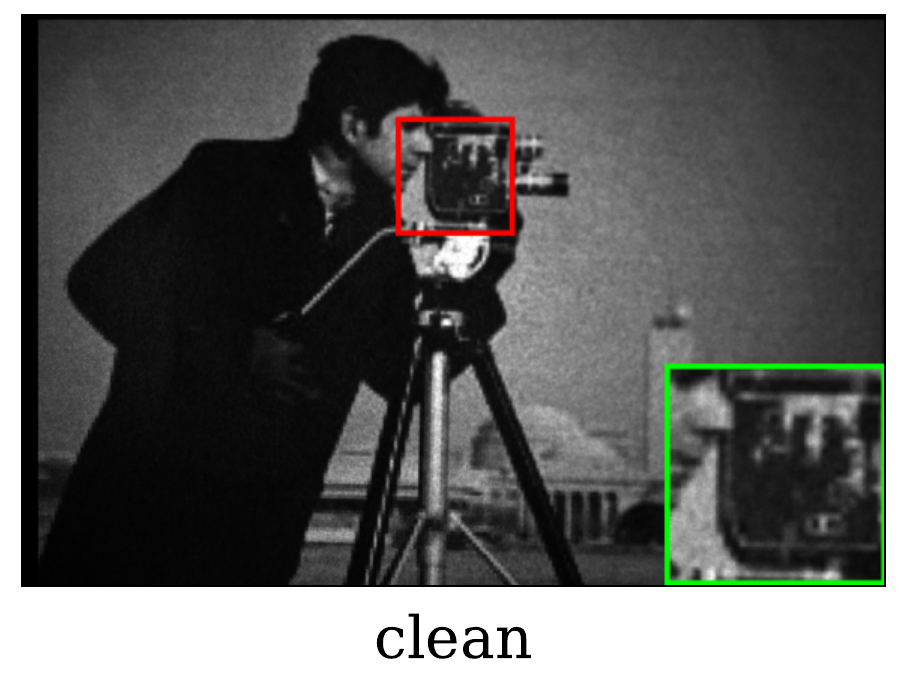}

\includegraphics[width=.34\textwidth]{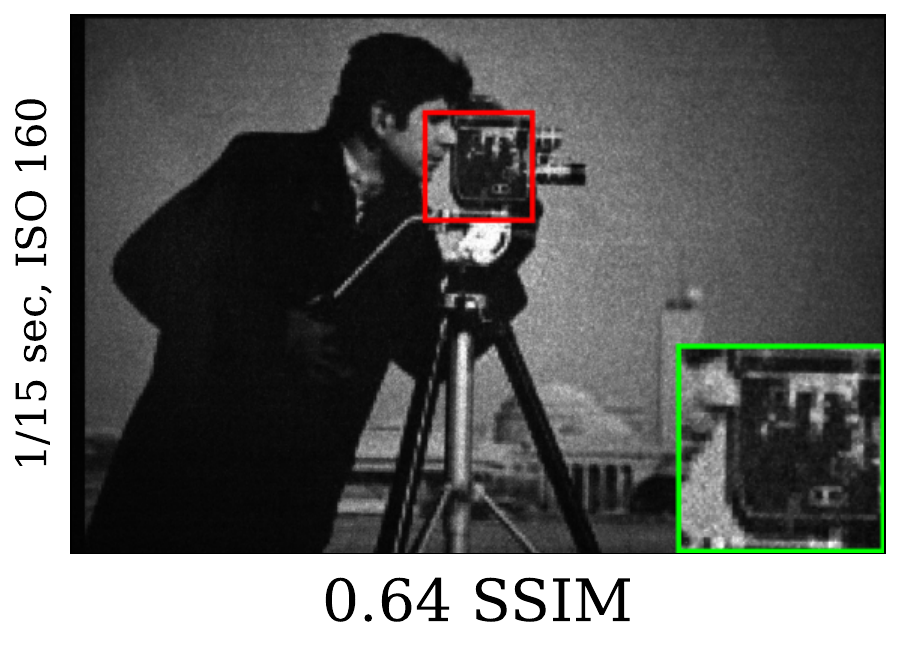}
\includegraphics[width=.31\textwidth]{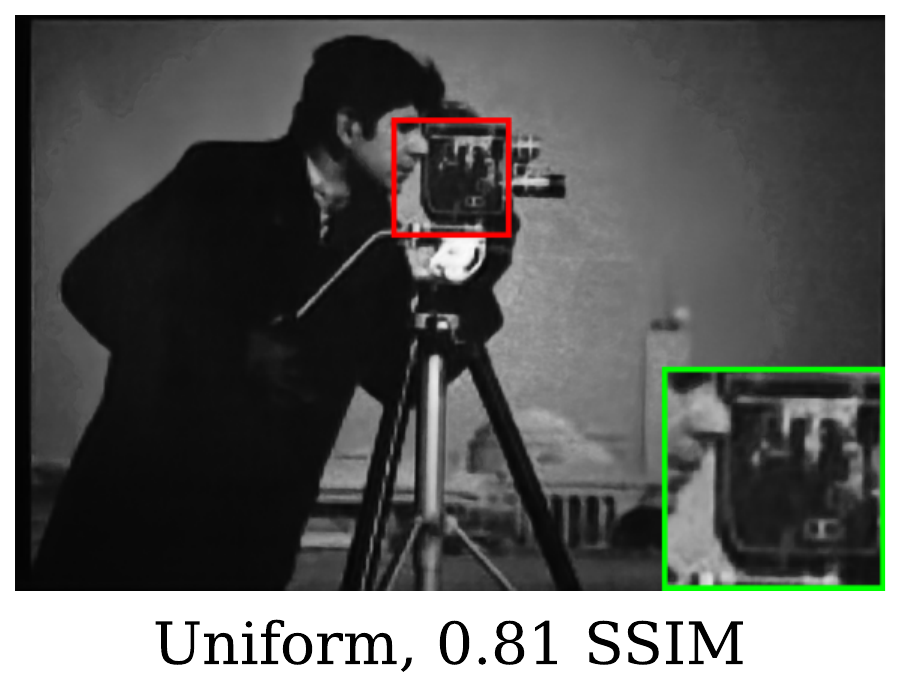} 
\includegraphics[width=.31\textwidth]{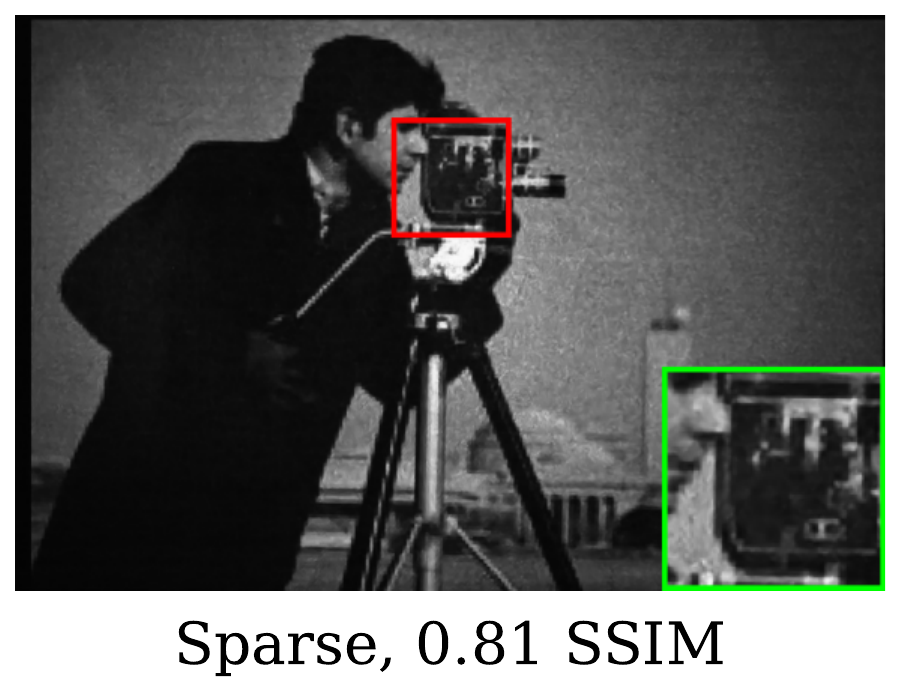}

\includegraphics[width=.34\textwidth]{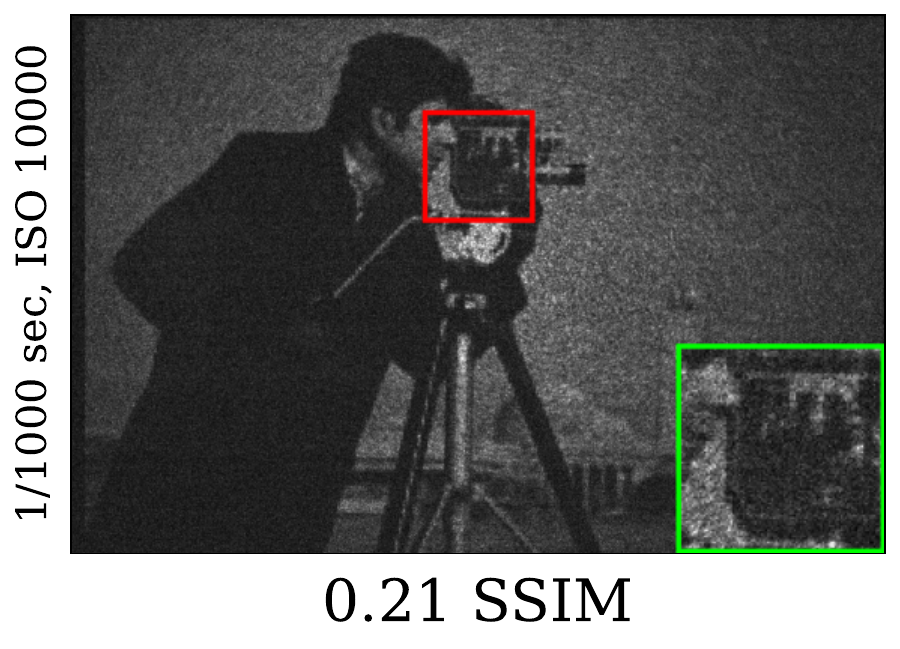}
\includegraphics[width=.31\textwidth]{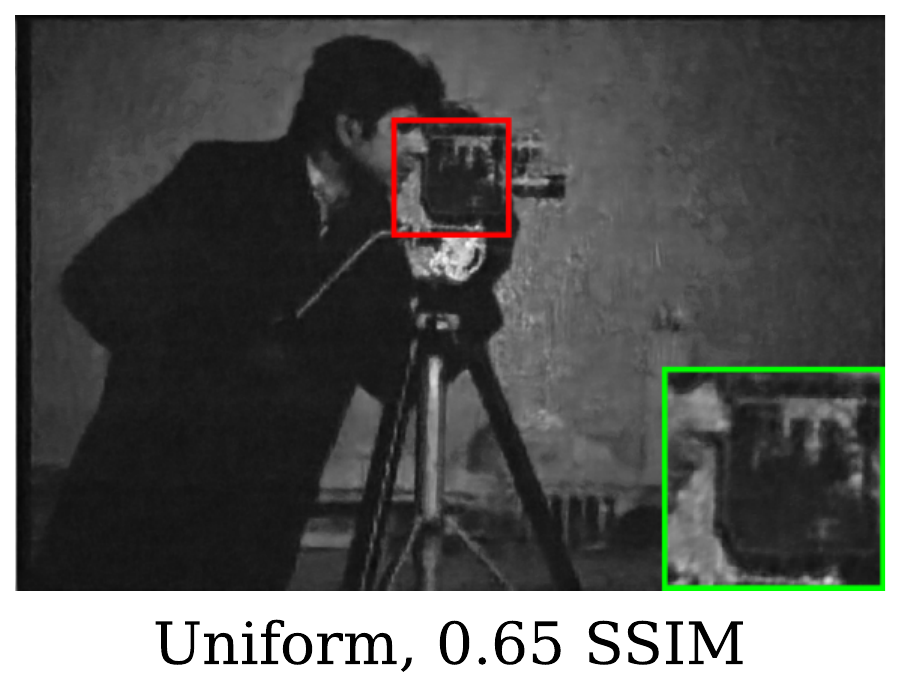} 
\includegraphics[width=.31\textwidth]{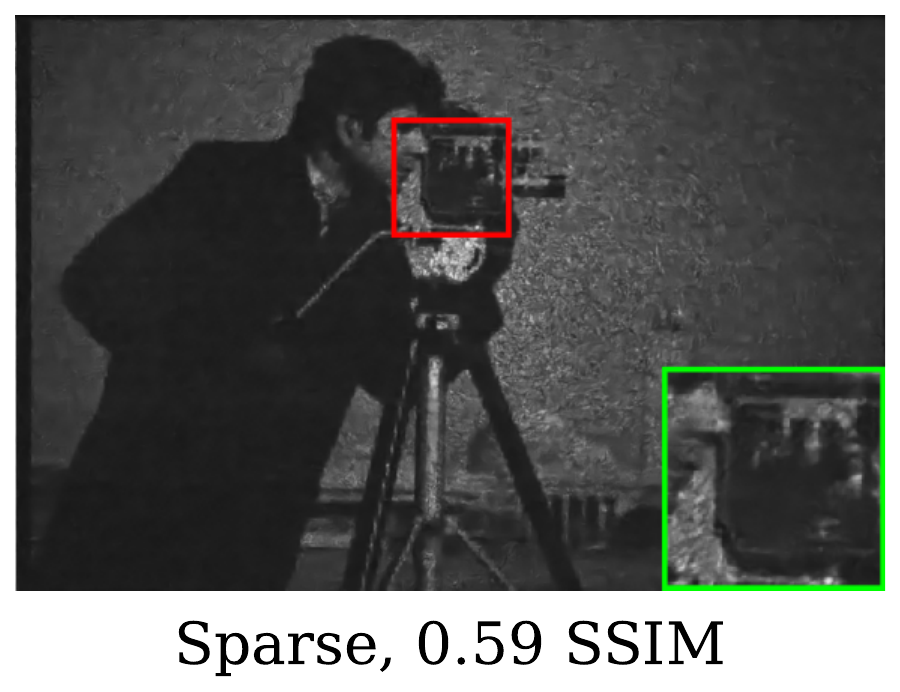}
\caption{ \textbf{Qualitative comparisons, experimental data.} Comparison between the performance of the ideal, uniform-trained, and
adaptive distribution, sparse sampling-trained denoisers on a sample image corrupted with a low amount of noise and
corrupted with a high amount of noise, as determined by the parameters of our experimental setup.
From the closeups it is apparent that the uniform trained denoiser has a tendency to oversmooth its inputs compared to the adaptive distribution trained denoiser. This leads to higher performance for the uniform trained at higher noise levels but lower perforamnce at lower noise levels. }
\label{fig:cameraman}
\end{figure*}

\begin{figure*}
\centering

\includegraphics[width=.31\textwidth]{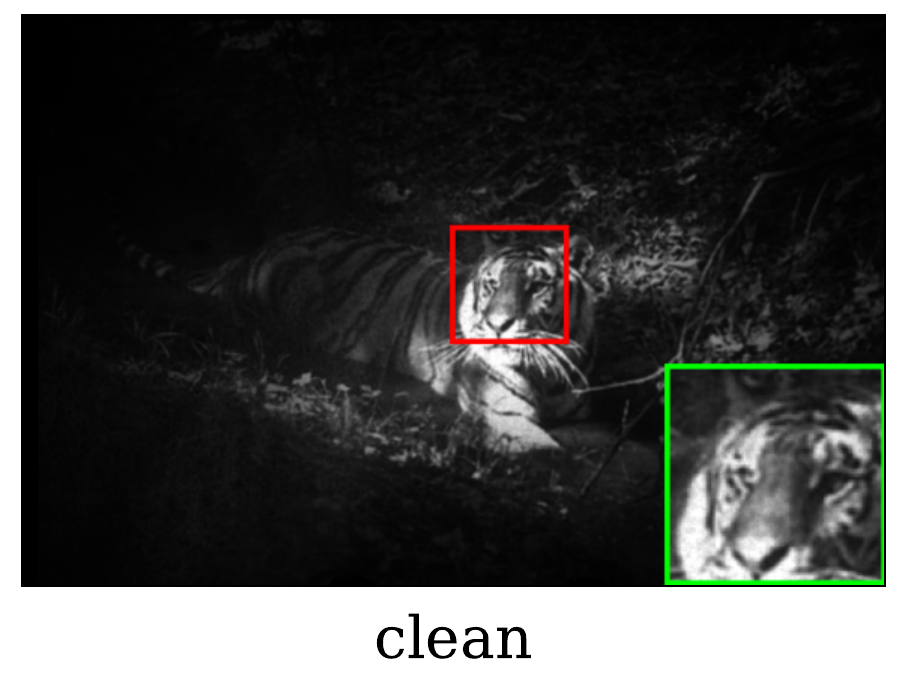}

\includegraphics[width=.34\textwidth]{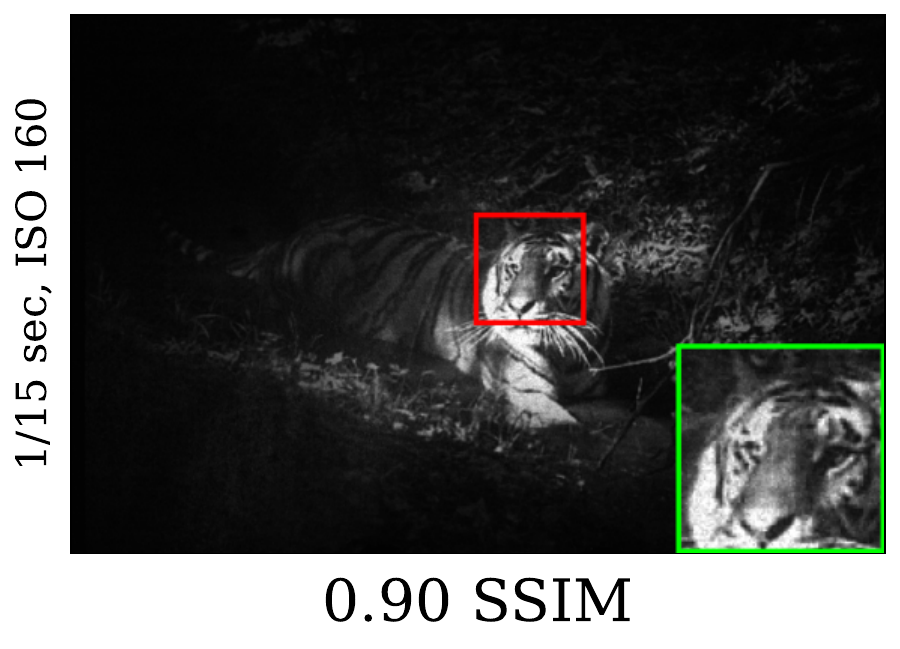}
\includegraphics[width=.31\textwidth]{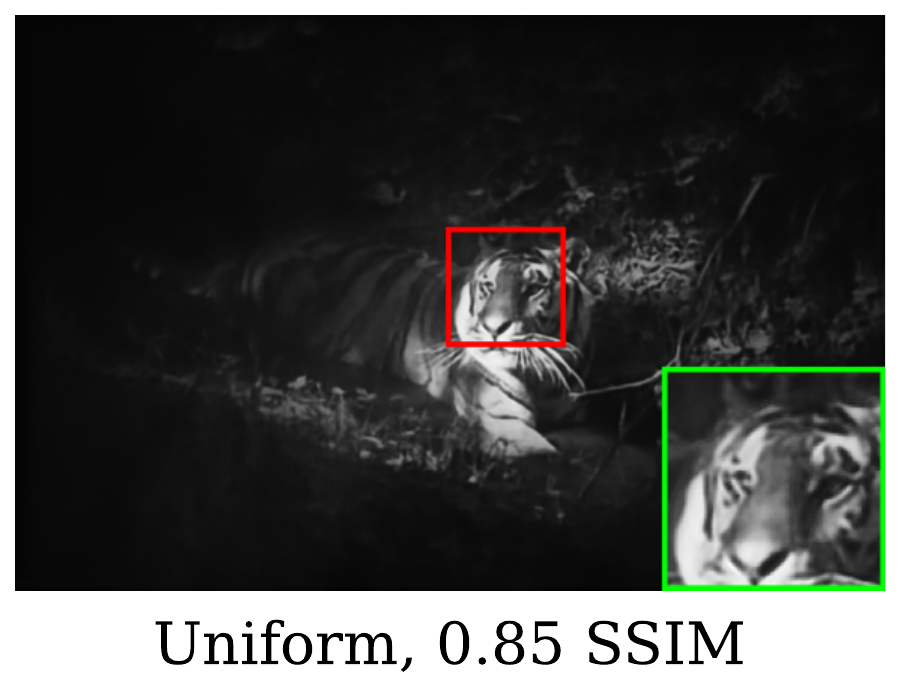} 
\includegraphics[width=.31\textwidth]{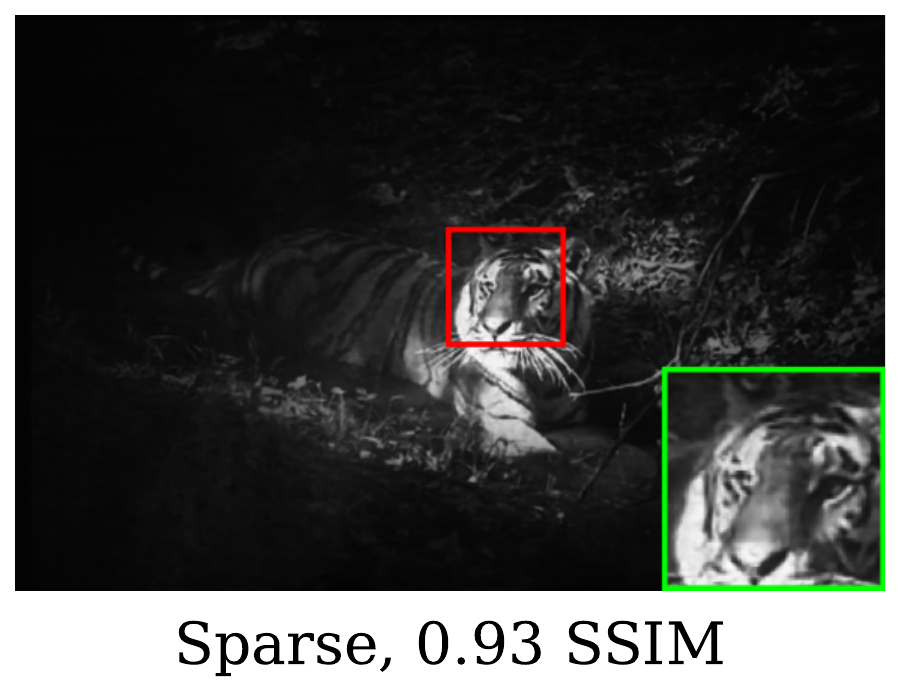}

\includegraphics[width=.34\textwidth]{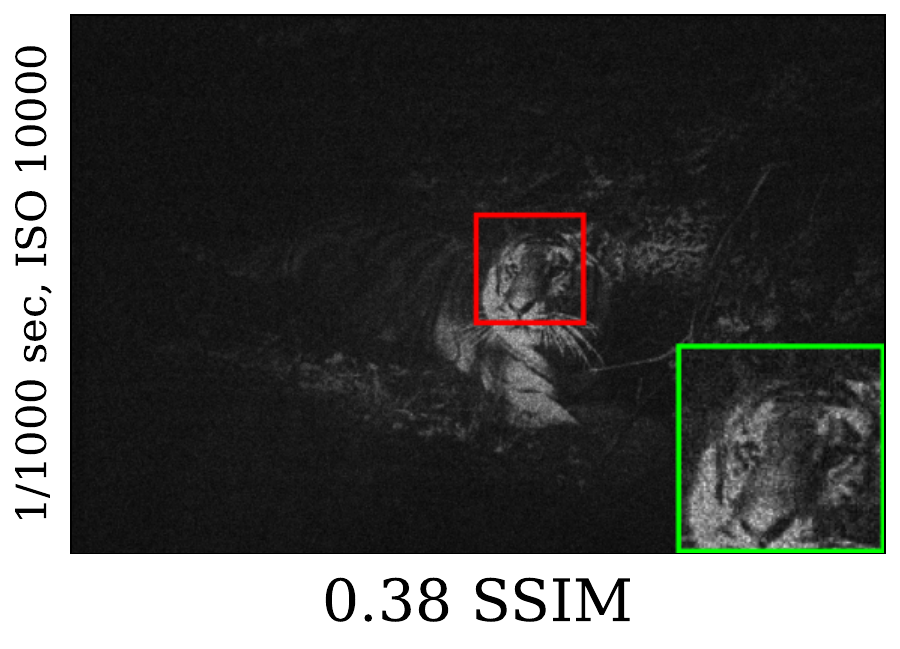}
\includegraphics[width=.31\textwidth]{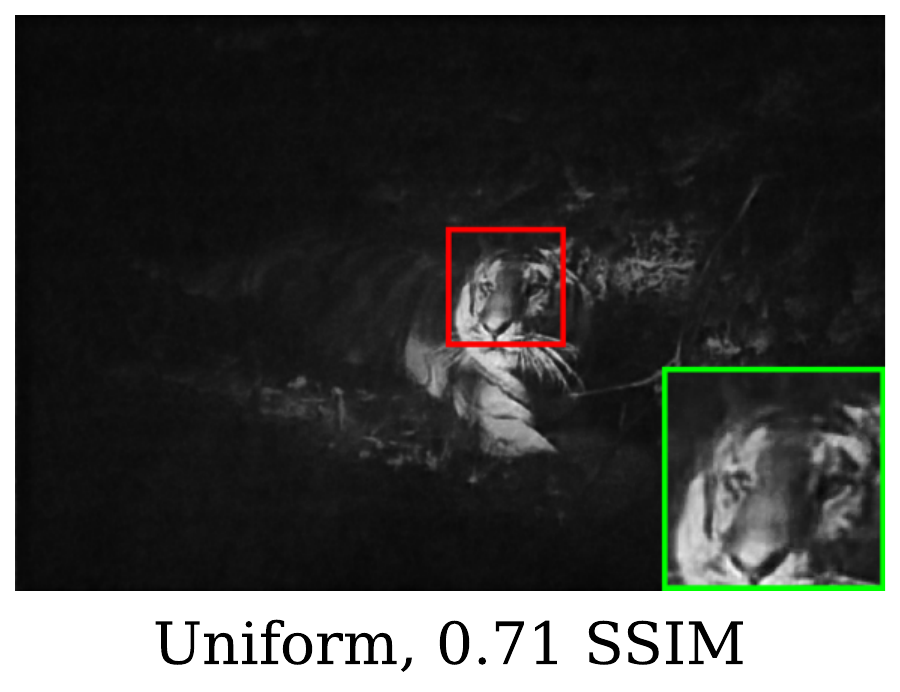} 
\includegraphics[width=.31\textwidth]{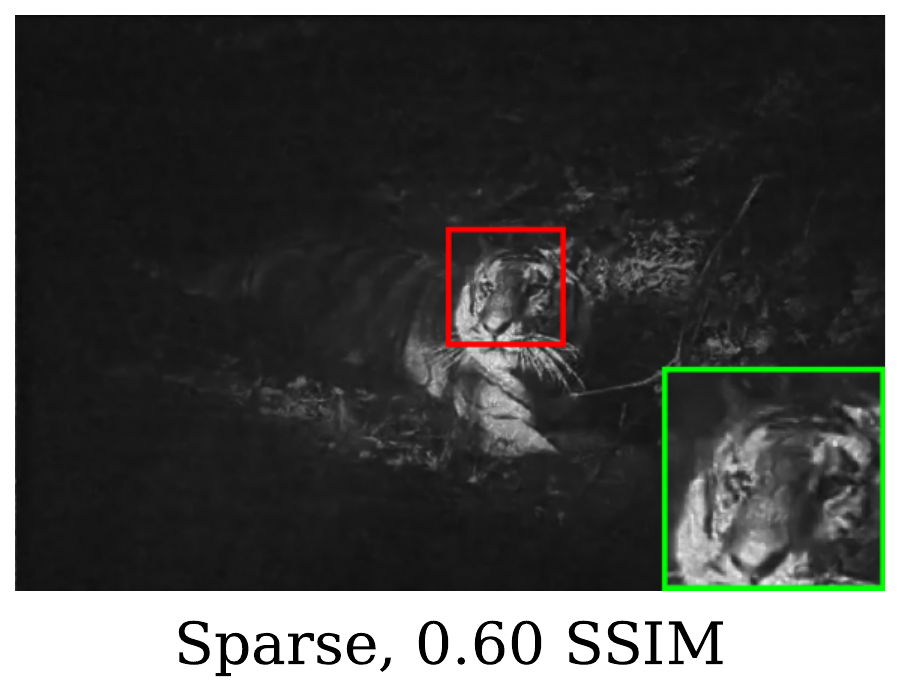}
\caption{ \textbf{Qualitative comparisons, experimental data.} Comparison between the performance of the ideal, uniform-trained, and
adaptive distribution, sparse sampling-trained denoisers on a sample image corrupted with a low amount of noise and
corrupted with a high amount of noise, as determined by the parameters of our experimental setup.
From the closeups it is apparent that the uniform trained denoiser has a tendency to oversmooth its inputs compared to the adaptive distribution trained denoiser. This leads to higher performance for the uniform trained at higher noise levels but lower perforamnce at lower noise levels. }
\label{fig:tiger}
\end{figure*}

\begin{figure*}
\centering

\includegraphics[width=.31\textwidth]{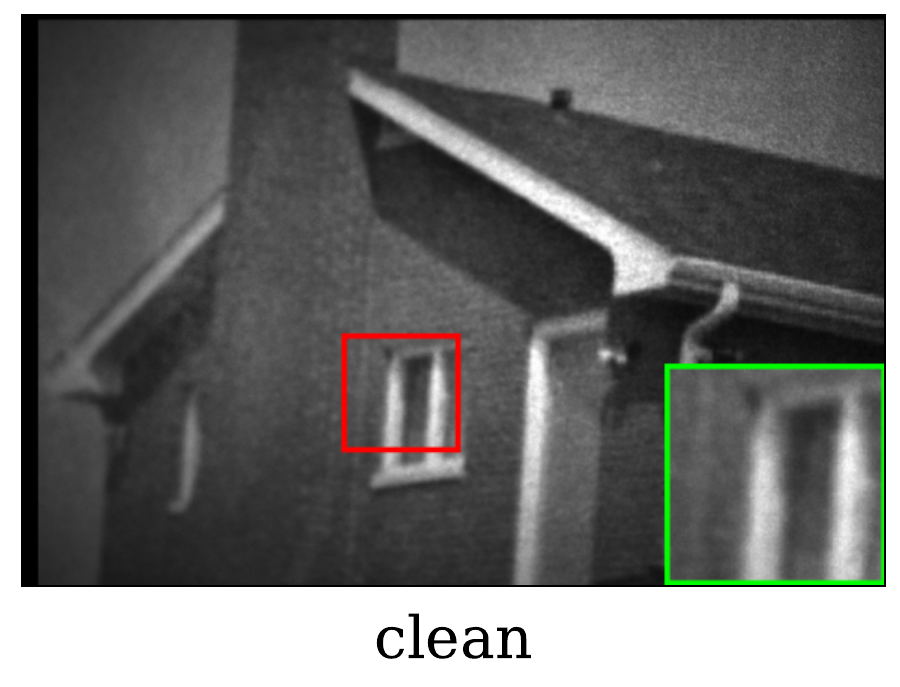}

\includegraphics[width=.34\textwidth]{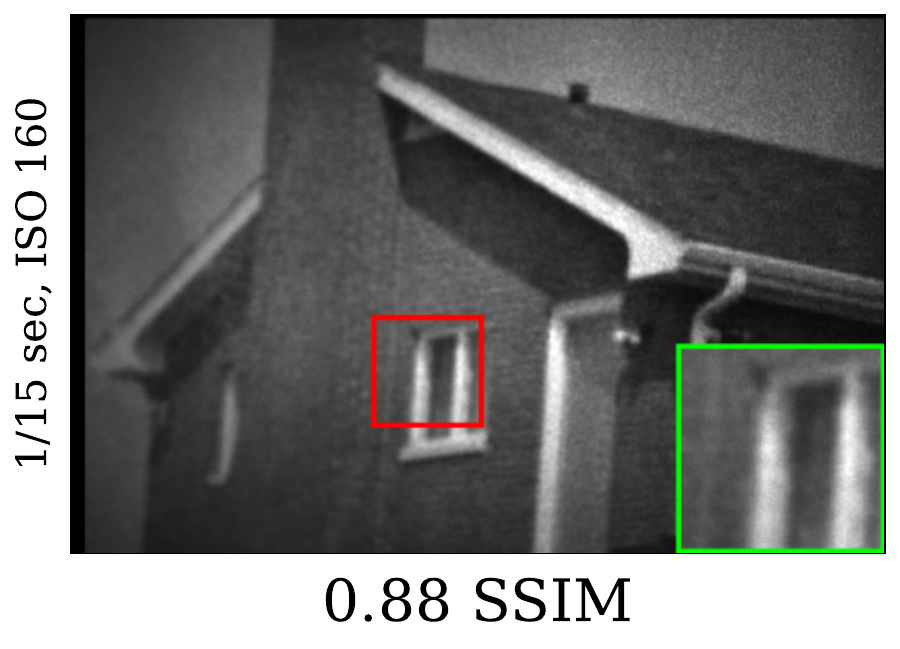}
\includegraphics[width=.31\textwidth]{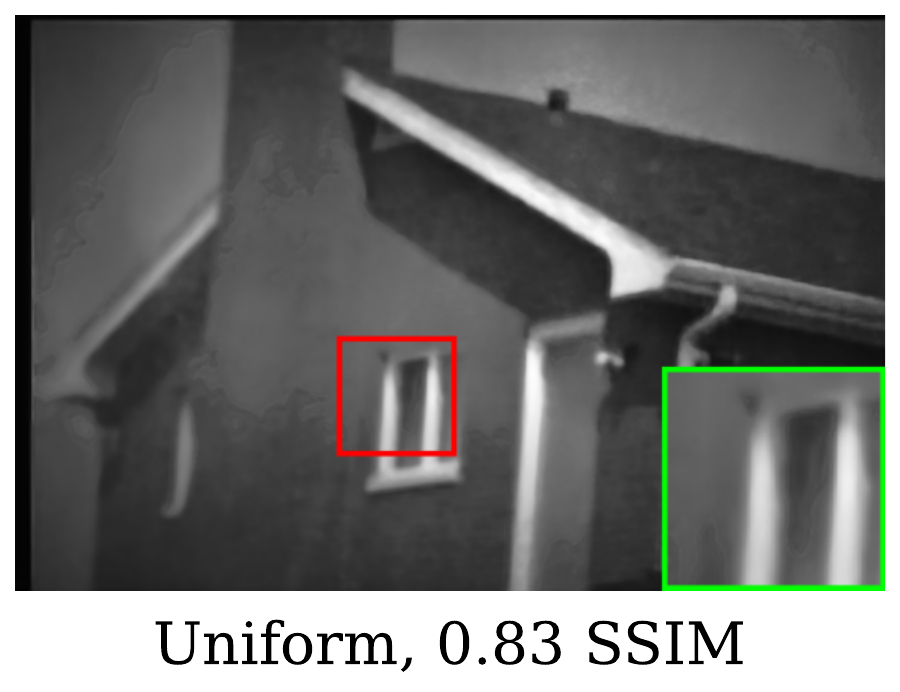} 
\includegraphics[width=.31\textwidth]{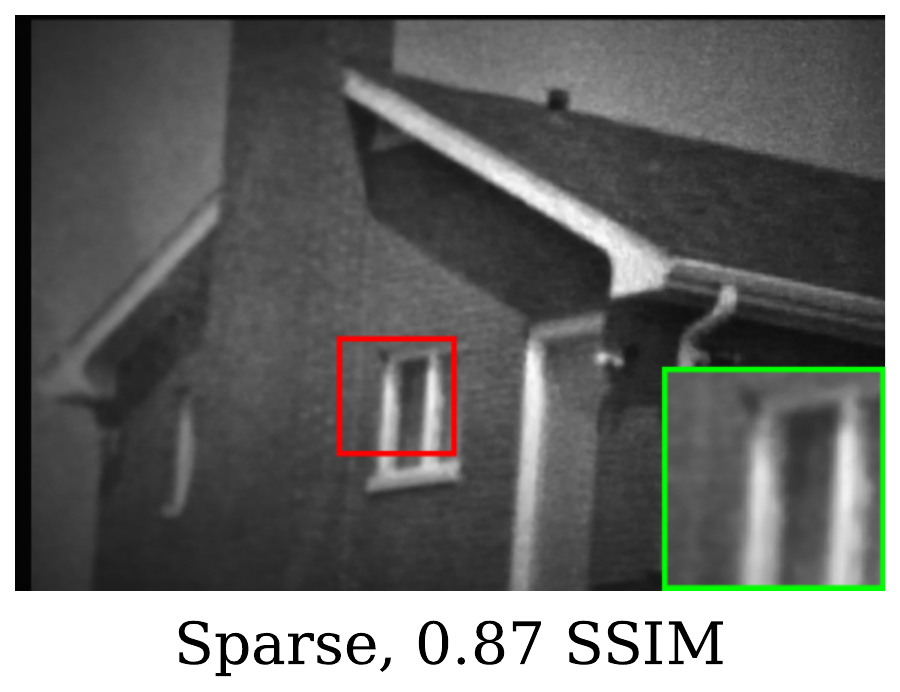}

\includegraphics[width=.34\textwidth]{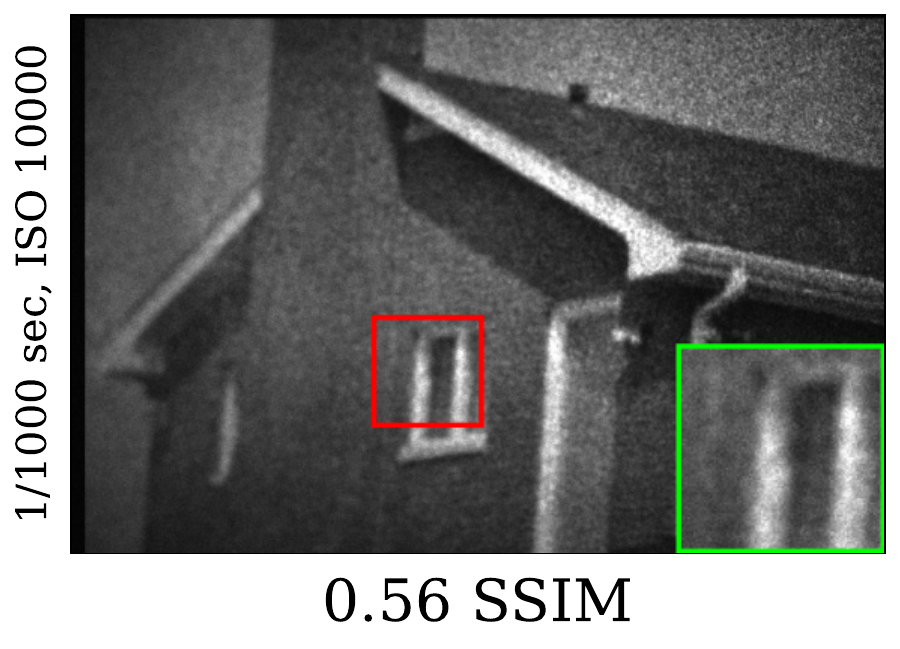}
\includegraphics[width=.31\textwidth]{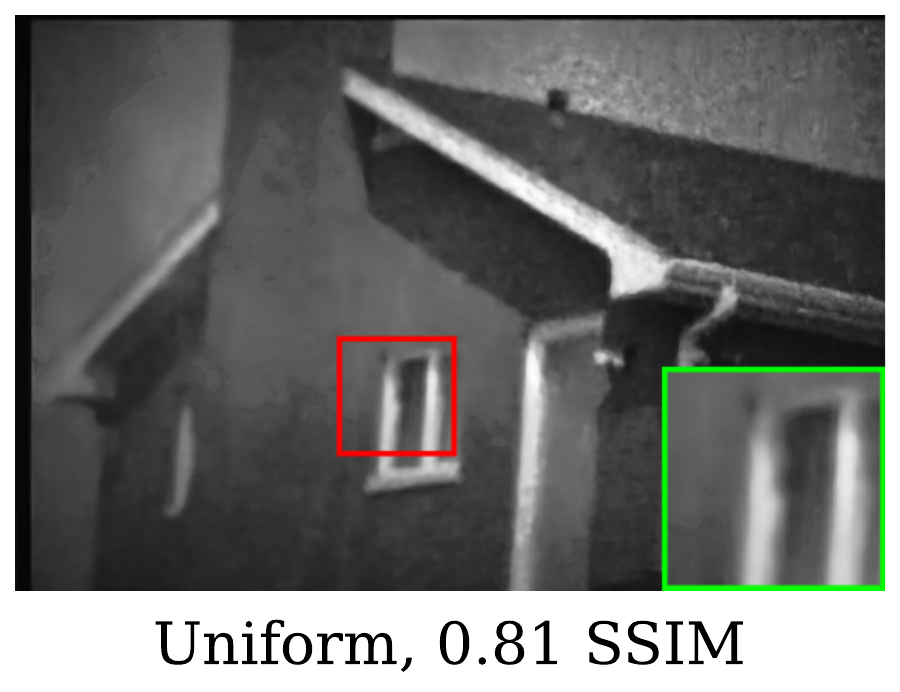} 
\includegraphics[width=.31\textwidth]{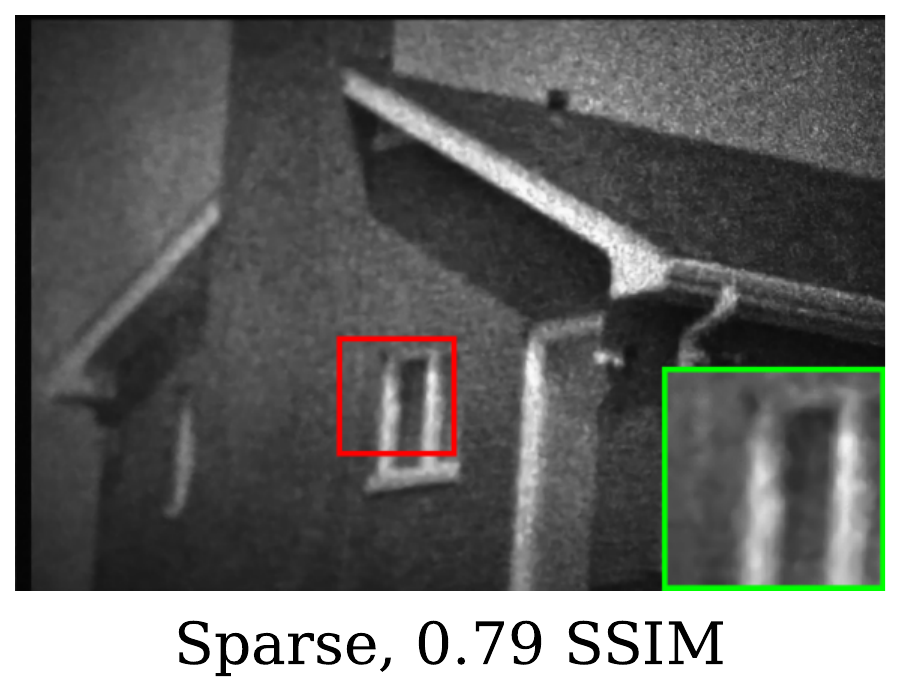}
\caption{ \textbf{Qualitative comparisons, experimental data.} Comparison between the performance of the ideal, uniform-trained, and
adaptive distribution, sparse sampling-trained denoisers on a sample image corrupted with a low amount of noise and
corrupted with a high amount of noise, as determined by the parameters of our experimental setup.
From the closeups it is apparent that the uniform trained denoiser has a tendency to oversmooth its inputs compared to the adaptive distribution trained denoiser. This leads to higher performance for the uniform trained at higher noise levels but lower perforamnce at lower noise levels. }
\label{fig:house}
\end{figure*}

\begin{figure*}
\centering

\includegraphics[width=.31\textwidth]{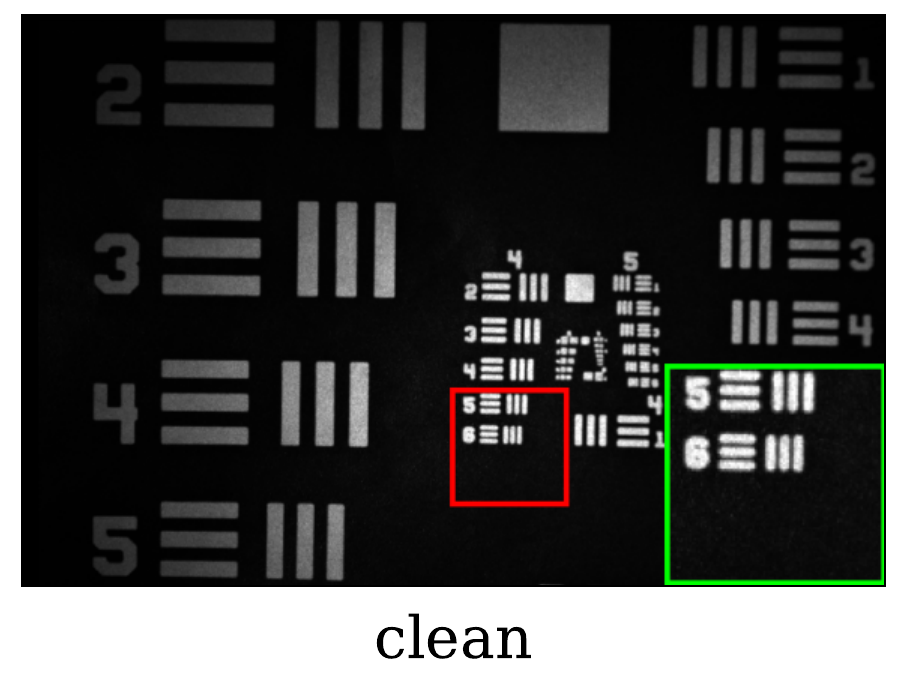}

\includegraphics[width=.34\textwidth]{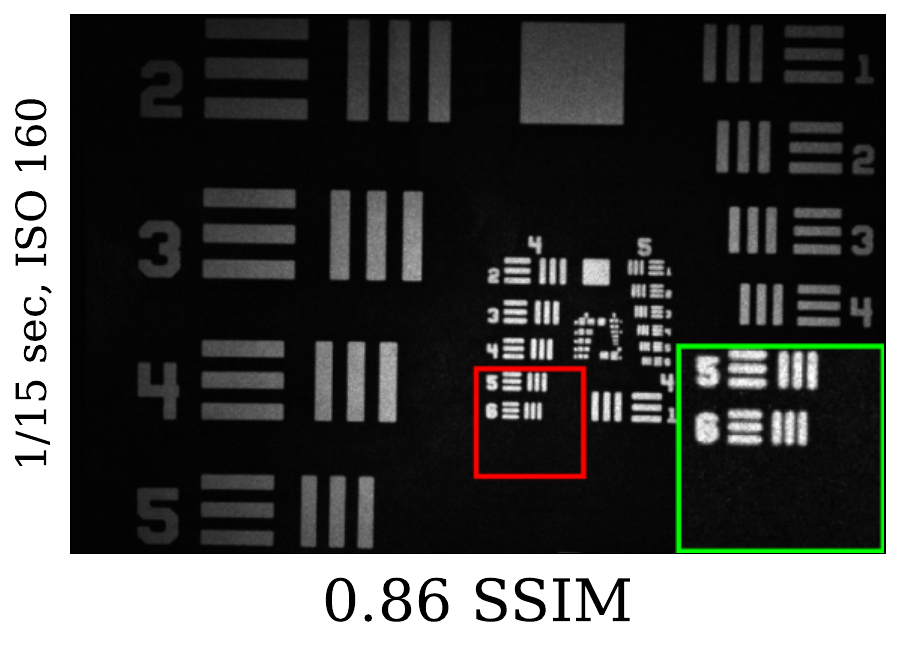}
\includegraphics[width=.31\textwidth]{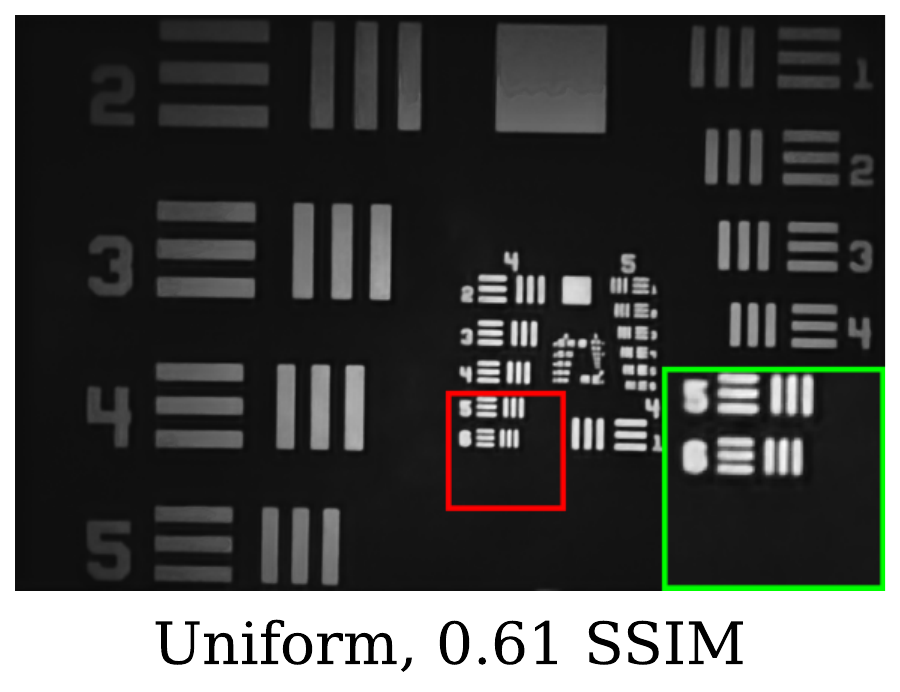} 
\includegraphics[width=.31\textwidth]{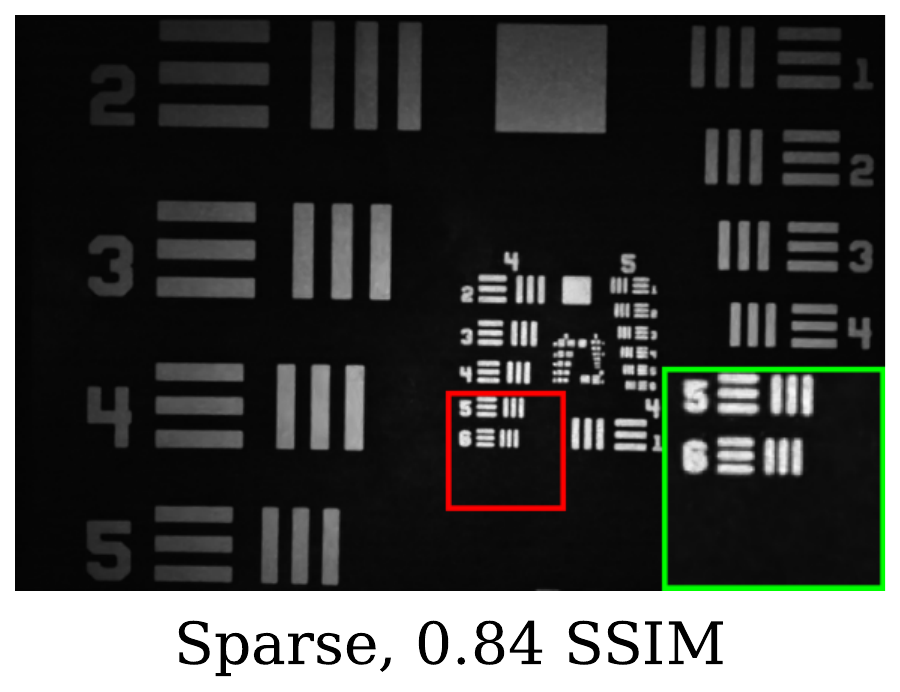}

\includegraphics[width=.34\textwidth]{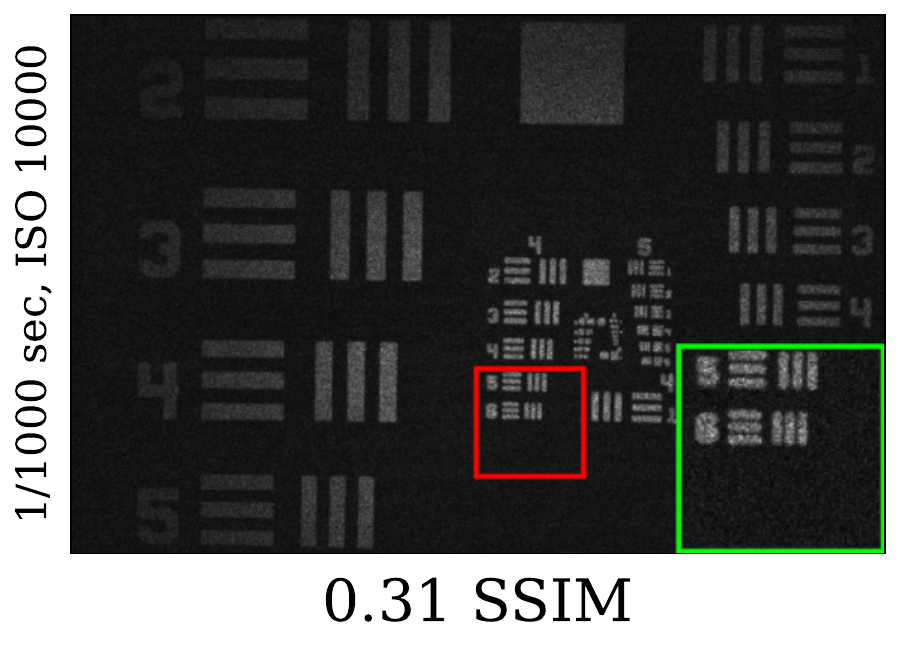}
\includegraphics[width=.31\textwidth]{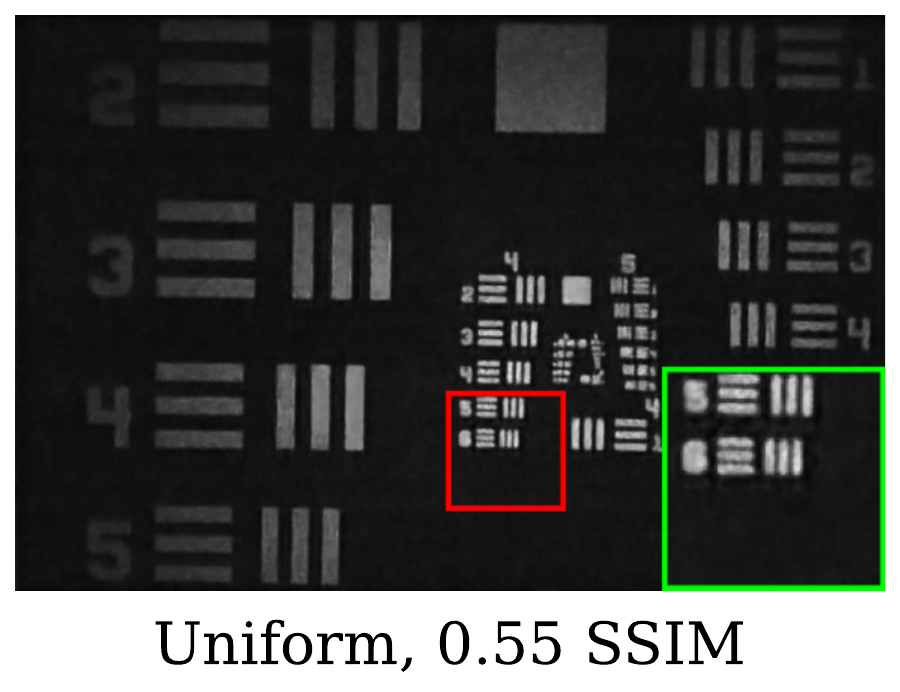} 
\includegraphics[width=.31\textwidth]{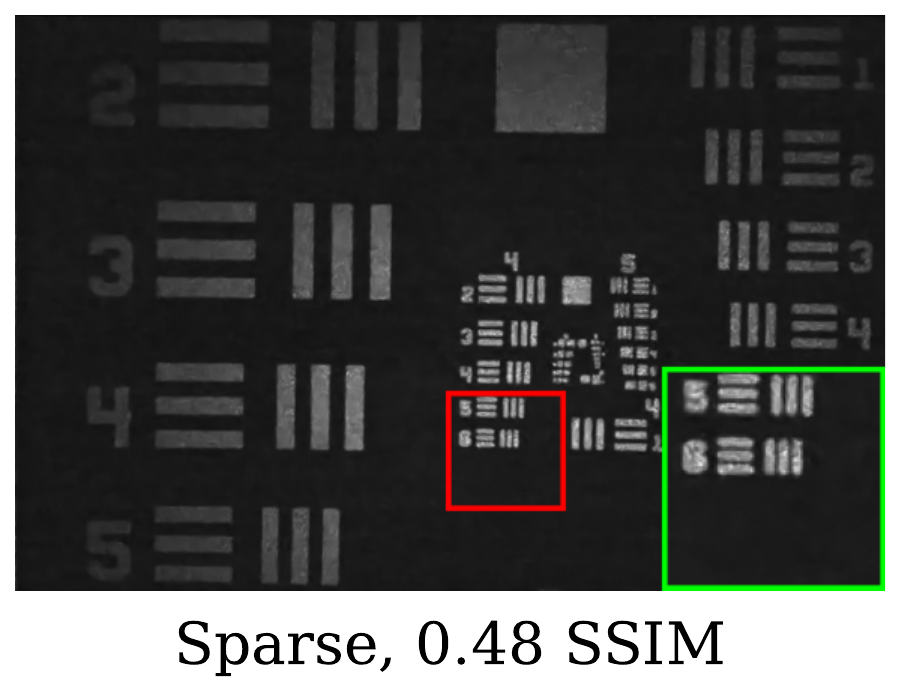}
\caption{ \textbf{Qualitative comparisons, experimental data.} Comparison between the performance of the ideal, uniform-trained, and
adaptive distribution, sparse sampling-trained denoisers on a sample image corrupted with a low amount of noise and
corrupted with a high amount of noise, as determined by the parameters of our experimental setup.
From the closeups it is apparent that the uniform trained denoiser has a tendency to oversmooth its inputs compared to the adaptive distribution trained denoiser. This leads to higher performance for the uniform trained at higher noise levels but lower performance at lower noise levels. }
\label{fig:res_target}
\end{figure*}

Additional qualitative results for the synthetic data are presented in Figure~\ref{fig:qualitative_cactus} and~\ref{fig:qualitative_train}.
Additional qualitative results for the experimental data are presented in Figures~\ref{fig:cameraman},  \ref{fig:tiger}, \ref{fig:house},\ref{fig:res_target}.

\end{document}